\documentclass[acmlarge,nonacm]{acmart}

\usepackage{algorithm}
\usepackage{algorithmic}

\usepackage{enumitem}
\usepackage{makecell}
\usepackage{multirow}
\usepackage{caption}
\usepackage{subcaption}
\usepackage{colortbl}
\usepackage[]{xcolor}
\usepackage{xcolor}
\usepackage{siunitx}
\begin{document}

\title{
RETRACE: Resilience-Guided Trait-Conditioned Craving Estimation from Wearable Physiology in Opioid Use Disorder}

\author{Yi Xiao}
\orcid{0000-0002-5261-5440}
\affiliation{%
  \institution{Arizona State University}
  \city{Tempe}
  \country{USA}}
\email{yxiao124@asu.edu}

\author{Harshit Sharma}
\orcid{0000-0002-7016-6220}
\affiliation{%
  \institution{Arizona State University}
  \city{Tempe}
  \country{USA}}
\email{hsharm62@asu.edu}

\author{Dessa Bergen-Cico}
\orcid{0000-0002-8852-732X}
\email{dkbergen@syr.edu}
\affiliation{%
  \institution{Syracuse University}
  \city{Syracuse}
  \state{New York}
  \country{USA}
  }
\author{Asif Salekin}
\orcid{0000-0002-0807-8967}
\affiliation{%
  \institution{Arizona State University}
  \city{Tempe}
  \country{USA}}
\email{Asif.Salekin@asu.edu}

\begin{abstract}
Detecting opioid craving from wearable physiological signals is critical yet difficult, with the potential to support proactive interventions for individuals with opioid use disorder (OUD). This challenge is especially pronounced under subject-independent evaluation because craving is subjective, heterogeneous, and often physiologically entangled with stress. Our empirical analysis shows that stress elicits strong and reproducible autonomic responses, while craving-related signals are weaker, sparse, and largely embedded within stress-related physiology. We further show that psychological resilience, which shapes stress regulation and craving vulnerability, is not reliably observable from short-term wearable windows, but can be captured through reusable subject-level proxies, including post-stress heart-rate recovery and autobiographical memory recall.

Motivated by these findings, we introduce \textbf{\emph{RETRACE}}, a resilience-guided trait-conditioned framework for subject-independent craving estimation from wearable physiology. \textbf{\emph{RETRACE}} reframes craving detection as trait-conditioned physiological interpretation: rather than assuming the same physiological pattern has the same meaning across individuals, it uses resilience-related subject context to guide inference. Technically, \textbf{\emph{RETRACE}} introduces a novel dual-encoder design that separates generalizable stress physiology from subject-specific craving interpretation. It combines a frozen stress-pretrained encoder with a resilience-conditioned craving encoder, using feature-level gating and representation-level fusion to enable lightweight personalization without target-user craving labels or per-user retraining.

We evaluate \textbf{\emph{RETRACE}} on a novel multimodal OUD dataset containing wearable physiology, stress and craving annotations, post-stress recovery dynamics, and autobiographical narratives. Under strict leave-one-subject-out evaluation, \textbf{\emph{RETRACE}} outperforms classical machine learning, domain generalization, and multimodal conditioning baselines, achieving up to 7\% absolute improvement over the strongest baseline. We further evaluate whether these gains generalize beyond a single benchmark by testing component contributions, comparing physiological and semantic resilience guidance, interpreting resilience-conditioned feature use, and evaluating stability across visits, wrist placements, runtime constraints, and an in-the-wild feasibility setting. These results show that robust wearable craving detection requires not only stronger physiological representations, but also a principled way to interpret ambiguous physiological evidence under human heterogeneity.

\end{abstract}

\maketitle
\section{INTRODUCTION}
Craving is a central driver of relapse in substance use disorders (SUDs) and a major barrier to sustained recovery \cite{sayette2016role,draghmeh2025review}. Among SUDs, 
opioid use disorder (OUD), in particular, remains a severe public health crisis marked by frequent relapse, high mortality, and substantial societal burden \cite{strang2020opioid,wang2025global,hoffman2019opioid}. Timely detection of heightened craving may support early awareness, personalized coping, and recovery support before craving escalates into relapse risk \cite{vafaie2022association,enkema2020disrupting}.
%opioid use disorder (OUD) remains one of the most severe public health crises, associated with frequent relapse, high mortality, and substantial societal burden \cite{strang2020opioid,wang2025global,hoffman2019opioid}. Timely identification of heightened craving states is therefore critical for supporting early awareness, personalized coping strategies, and clinically informed recovery support before craving escalates into relapse risk \cite{vafaie2022association,enkema2020disrupting}. 
Continuous, non-invasive monitoring using wearable physiological sensors offers a promising pathway for scalable craving detection outside clinical settings \cite{perski2022technology,garland2023zoom}.
%Craving is a central driver of relapse in substance use disorders (SUDs) and a critical barrier to sustained recovery \cite{sayette2016role,draghmeh2025review}. Among SUDs, opioid use disorder (OUD) remains one of the most widespread and devastating public health crises worldwide, associated with high mortality, recurrent relapse, and substantial societal burden \cite{strang2020opioid,wang2025global,hoffman2019opioid}. Timely identification of heightened craving states is therefore essential for delivering just-in-time interventions that can prevent relapse and support long-term rehabilitation. Continuous, non-invasive craving detection using wearable physiological sensors has the potential to significantly support individuals with OUD by enabling monitoring and personalized intervention beyond clinical settings \cite{perski2022technology,garland2023zoom}.

Despite this promise, physiological craving detection remains fundamentally challenging, particularly under subject-independent evaluation where models must generalize to unseen individuals \cite{jeyadevan2024time}. Unlike many physiological sensing tasks with relatively consistent autonomic signatures, craving-related physiology is often subtle, heterogeneous, and \emph{embedded within broader stress-related autonomic activity} \cite{carreiro2020wearable,carreiro2024evaluation}. Consequently, similar physiological patterns may carry different psychological meanings across individuals.
%A central difficulty is that stress and craving are closely intertwined but not physiologically equivalent. Stress is a well-established trigger of substance craving and often elicits strong, widespread autonomic responses, which has made stress detection a comparatively robust wearable-sensing task \cite{sinha1999stress,sinha2007role,preston2011stress,ollander2016comparison,menghini2019stressing,haleem2021biosensors}. Craving, in contrast, is a subjective motivational state that may or may not emerge under stress, and its physiological signatures are often subtle, heterogeneous, and embedded within broader stress-related autonomic activity \cite{ray2018neurobiology,carreiro2020wearable,carreiro2024evaluation}. 
Most existing approaches adopt a state-level inference paradigm, directly mapping short windows of physiological signals to craving labels using population-trained models \cite{carreiro2020wearable,chapman2022impact,gullapalli2019body}. This assumes a stable mapping between physiology and craving across individuals. Under substantial inter-individual variability, however, this assumption becomes fragile, making short-window physiological modeling alone insufficient for capturing the subject-dependent meaning of craving-related signals.

A key factor underlying this variability is psychological resilience that shapes how individuals regulate and respond to stress \cite{yang2020resilience,windle2011resilience,ungar2020resilience}. Rather than manifesting as an instantaneous physiological signal, resilience is a relatively stable trait-level construct that governs how physiological responses unfold and are interpreted across individuals. Prior work suggests that \emph{resilience is associated with inter-individual differences in craving sensitivity} \cite{crowell2013psychosocial,schetter2011resilience,compas2017coping,rathinam2021resilience}. This suggests that craving detection requires more than learning invariant short-window physiological representations; it also requires modeling how stable subject-level factors shape the interpretation of physiological evidence. We therefore conceptualize craving detection as a trait-conditioned inference problem, in which individual-level context guides how physiological signals are mapped to reported craving.

To operationalize this perspective, we design a wearable craving detection framework that conditions physiological inference on subject-level resilience. Rather than assuming that each physiological window has the same meaning across individuals, the framework uses reusable subject-specific context to guide how wearable signals are interpreted. A key challenge is that resilience is not directly observable from short-term physiological windows, as shown in Section~\ref{sec:rq1}, and direct measurement may be difficult to scale \cite{tai2023conceptualizing,wolke2025systematic,den2022conceptualizing} in everyday sensing settings. We therefore explore practical resilience proxies that can be collected once and reused during inference. Specifically, we consider two complementary forms of trait-level guidance: heart-rate recovery dynamics, quantified using HR-AUC after stress exposure \cite{den2022conceptualizing,yang2020resilience}, and semantic representations derived from autobiographical memory recall \cite{garcia2013integrative}. HR-AUC provides a physiology-grounded measure of post-stress recovery under controlled conditions, while autobiographical narratives provide a lower-burden source of subject-level context that does not require controlled stress induction.

Building on these insights, we propose \textbf{\emph{RETRACE}}—a \textbf{re}silience-guided \textbf{tra}it-conditioned \textbf{c}raving \textbf{e}stimation from wearable physiology in opioid use disorder. \textbf{\emph{RETRACE}} combines a stress-pretrained encoder with a resilience-related subject-level representation. 
The stress encoder learns transferable autonomic structure from reliable stress labels that partially overlap with craving physiology, providing a stable backbone for interpreting weaker and more heterogeneous craving signals (Section~\ref{Sec:Physio-craving-limit}).
Importantly, \textbf{\emph{RETRACE}} does not treat resilience as a direct physiological marker of craving. Instead, it uses resilience as contextual guidance for inference, controlling which physiological patterns are emphasized for each individual and how strongly generalizable stress physiology vs. subject-specific autonomic patterns contribute to craving prediction. This enables principled, low-capacity personalization under strict subject-independent evaluation.

%Need to motivate: requiring per-user craving labels or user-specific model retraining is hard because getting craving annotation is difficult. Stress inductions are validated and explicit, auto biography is even simpler. In contrast, elicitation of craving is difficult and uncertain, making personalization difficult. 
%population level, vs. subject dependant physiological markers.....

\subsection{Contributions}
This work makes the following contributions:

\begin{itemize}

\item \textbf{Trait-conditioned inference for physiological craving detection.}  
We introduce \textbf{\emph{RETRACE}}, to our knowledge, the first resilience-conditioned, stress-pretrained framework for subject-independent wearable craving detection in OUD. \textbf{\emph{RETRACE}} moves beyond a one-size-fits-all mapping from wearable physiology to craving by using trait-level context to guide how signals are interpreted for each individual. This enables subject-aware craving inference with low-capacity personalization, without requiring the target user’s craving labels or user-specific model retraining.
%We introduce \textbf{\emph{RETRACE}}, to our knowledge, the first resilience-conditioned, stress-pretrained framework for subject-independent wearable craving detection in OUD. Instead of assuming that consistent physiological patterns indicate craving for all individuals, RETRACE uses trait-level context to guide how wearable signals are interpreted. This enables subject-aware, low-capacity personalization without requiring the target user’s craving labels or user-specific model retraining.
%We introduce \textbf{\emph{RETRACE}}, to our knowledge, the first resilience-conditioned, stress-pretrained framework for subject-independent craving detection in OUD. It separates shared physiological structure from subject-specific interpretation. This formulation reframes craving detection from learning invariant mappings to modeling subject-dependent interpretation, enabling personalized inference under subject-independent evaluation without per-user retraining.

%We introduce \textbf{\emph{RETRACE}}, to our knowledge, the first resilience-conditioned, stress-pretrained framework for subject-independent craving detection in OUD. This reframes wearable craving detection from learning invariant physiological signal mappings to modeling trait-conditioned physiological interpretations, enabling personalized inference under subject-independent evaluation without per-user retraining.

%\item \textbf{Practical and complementary proxies for trait-level guidance.}  
\item \textbf{Reusable resilience-related guidance for wearable craving sensing.}
To support this trait-conditioned inference, we operationalize two forms of subject-level resilience guidance or proxies that can be collected once and reused during wearable craving inference: post-stress heart-rate recovery dynamics, quantified by HR-AUC after stress exposure, and semantic representations derived from autobiographical memory recall. Together, these proxies capture a practical design trade-off: HR-AUC provides a physiology-grounded measure under controlled conditions, while autobiographical recall offers a lower-burden source of subject-level context that may better support scalability.

\item \textbf{A multimodal dataset for studying stress, craving, and resilience in OUD.}
We introduce, to our knowledge, the first dataset that jointly captures wearable physiological signals, craving labels, stress responses, and resilience-related attributes from the same participants, including both physiological recovery measures and autobiographical recall-based proxies (Section~\ref{sec:userstudy}). This dataset enables direct empirical investigation of how stress, craving, and resilience interact, moving physiological craving research beyond state-level modeling. To support future work, we will publicly release the wearable physiological signals, derived physiological resilience values, and encoded autobiographical memory representations. Due to IRB restrictions raw autobiographical transcripts will not be publicly released.

%\item \textbf{Comprehensive validation of the \textbf{\emph{RETRACE}} framework.}

\item \textbf{Comprehensive validation and deployment-oriented analysis of \textbf{\emph{RETRACE}}.}
We provide a systematic analysis of the relationship between stress, craving, resilience, and physiological signals (Section~\ref{Emperical-motivation}), revealing fundamental challenges for subject-independent craving detection and motivating the need to incorporate resilience-aware inference. We evaluate \textbf{\emph{RETRACE}} under strict subject-independent settings (Section~\ref{Eval}) against classical machine learning, domain generalization, and multimodal conditioning baselines, demonstrating that neither invariant representation learning nor treating resilience guidance as auxiliary input is sufficient for robust craving estimation (Section~\ref{sec:eval_baseline}). We further conduct ablation studies to quantify the roles of the stress-pretrained encoder, resilience-conditioned craving encoder, feature-level gating, representation-level fusion, and training losses (Section~\ref{sec:eval_ablation}). We compare physiological and semantic resilience guidance to examine their trade-offs between physiological fidelity and practical scalability, and analyze positive and negative autobiographical recall to understand how different forms of subject-level context contribute to inference (Section~\ref{sec:eval_baseline} and ~\ref{sec:eval_ablation}). Through interpretability analyses, we examine how resilience guidance changes physiological feature use and supports subject-aware interpretation (Section~\ref{sec:interpretation}). We assess deployment feasibility through cross-visit and cross-wrist robustness tests, temporal reusability of guidance embeddings (Section~\ref{sec:eval_sensitivity}), runtime analysis for efficient edge deployment (Section~\ref{runtime-val}), and a 14-day EMA-based real-world feasibility study (Section~\ref{sec:inthewild}). Finally, we discuss deployment considerations (Section~\ref{sec:deployment}), including efficiency, usability, and trade-offs between physiological and semantic resilience guidance, along with broader implications and limitations (Section~\ref{Discussion}).

%We evaluate \textbf{\emph{RETRACE}} under strict subject-independent settings (Section~\ref{Eval}) against classical machine learning, domain generalization, and multimodal baselines, and conduct ablation, task-level, and interpretability analyses to examine how resilience-related guidance shapes physiological interpretation in \textbf{\emph{RETRACE}}. We further validate the robustness of the guidance embeddings across time, demonstrating their reusability, and conduct a feasibility evaluation in EMA-based real-world scenarios (Section~\ref{sec:inthewild}), including runtime analysis for efficient edge deployment. We additionally examine deployment considerations for practical use (Section~\ref{sec:deployment}), including efficiency, usability, and trade-offs between physiological and semantic resilience guidance, and discuss broader implications and limitations (Section~\ref{Discussion}).

\end{itemize}

\section{RELATED WORKS}
\label{sec:related_work}

\subsection{Physiological Craving Detection and Its Limitations}
\label{Sec:Physio-craving-limit}
%Wearable physiological sensing has become a central approach for inferring affective states, most notably stress, using signals such as electrodermal activity (EDA), heart rate and heart rate variability (HR/HRV), skin temperature, and motion \cite{sharma2022psychophysiological,mcewen2020revisiting,ollander2016comparison,menghini2019stressing}. Over the past decade, stress detection has matured into a well-established research area, supported by standardized experimental paradigms, dense annotations, and reproducible physiological signatures. Most approaches follow a \emph{state-level inference} paradigm \cite{schmidt2018wearable,vos2023generalizable}, in which short temporal windows of physiological signals are mapped to momentary labels using population-level models.
Wearable physiological sensing is widely used to infer affective states, especially stress, from signals such as electrodermal activity (EDA), heart rate and heart rate variability (HR/HRV), skin temperature, and motion \cite{sharma2022psychophysiological,mcewen2020revisiting,ollander2016comparison,menghini2019stressing}. Stress detection has become a mature sensing task, supported by standardized protocols, dense annotations, and reproducible physiological signatures. Most methods follow a \emph{state-level inference} paradigm, mapping short physiological windows to momentary labels with population-level models \cite{schmidt2018wearable,vos2023generalizable}.

Wearable craving detection, in contrast, remains largely feasibility-oriented. Prior studies have shown that self-reported craving can be inferred from biosignals using classical machine learning models such as k-nearest neighbors and support vector machines \cite{carreiro2020wearable,carreiro2024evaluation}. However, these studies often rely on within-cohort or event-level evaluation, limiting evidence for subject-independent generalization. Related work has also focused on detecting objective opioid use events from wearables \cite{chapman2022impact,gullapalli2021opitrack}, which exhibit more stable pharmacological patterns and longer temporal dynamics than subjective craving. A recent scoping review further identifies the lack of standardized datasets, subject-independent protocols, and principled modeling frameworks as major barriers in this domain \cite{jeyadevan2024time}.
%In contrast, wearable-based craving detection remains largely feasibility-oriented. Prior work has demonstrated that self-reported craving can be inferred from wearable biosignals using classical machine learning models such as k-nearest neighbors and support vector machines \cite{carreiro2020wearable,carreiro2024evaluation}. 
%However, these studies typically rely on within-cohort or event-level evaluation, limiting their applicability to subject-independent settings. Other efforts have focused on detecting objective opioid use events using wearable signals \cite{chapman2022impact,gullapalli2021opitrack}, which exhibit more stable pharmacological patterns and longer temporal dynamics, rather than subjective craving from short-term physiological observations. A recent scoping review highlights the lack of standardized datasets, subject-independent evaluation protocols, and principled modeling frameworks as key barriers in this domain \cite{jeyadevan2024time}.

Stress is a well-established trigger of craving, and stress exposure can increase subjective craving across settings \cite{mereish2024vicarious,sinha2024stress}. Because directly eliciting opioid craving through drug-related cues can raise ethical and clinical concerns, especially for vulnerable populations, many studies use stress induction as a controlled physiological probe while measuring craving independently through self-report \cite{sinha2000psychological,fox2008enhanced}. However, craving-related physiology is often subtle, heterogeneous, and embedded within broader stress-related autonomic activity \cite{weinstein1997imagery}. This creates inconsistent signal-label relationships across individuals and challenges standard state-level inference, which assumes a shared mapping between physiology and labels. We therefore use standard physiological classifiers as baselines and motivate models that explicitly account for inter-individual differences in how physiological signals relate to craving.

\subsection{Modeling Inter-Individual Variability in Physiological Sensing}
\label{sec:DGMM}
A central challenge in physiological sensing is the substantial inter-individual variability in how physiological signals relate to internal psychological states. To address this challenge, prior work has explored approaches to improve generalization across subjects, tasks, and environments.

\textbf{Domain generalization.}
Domain generalization (DG) and invariant learning methods aim to learn representations that remain robust across subjects, tasks, or environments by enforcing invariance across training domains \cite{arjovsky2019invariant,krueger2021out,sagawa2019distributionally}. These approaches have been applied to physiological sensing tasks such as emotion recognition and workload estimation, where models seek subject-invariant features that generalize to unseen users \cite{apicella2024toward,qin2022domain,xiao2025human}. However, DG methods typically assume that a shared mapping exists between physiological signals and labels across individuals. In craving detection, this assumption may not hold because similar autonomic responses can carry different meanings depending on individual differences in stress regulation. Our evaluation includes representative DG methods: IRM \cite{arjovsky2019invariant}, GroupDRO\cite{sagawa2019distributionally}, and V-REx\cite{krueger2021out}, as baselines for subject-independent craving detection.

%Domain generalization (DG) and invariant learning methods aim to learn representations that are robust to distribution shifts across individuals by enforcing invariance across training domains \cite{arjovsky2019invariant,krueger2021out,sagawa2019distributionally}. 
%These approaches have been applied in physiological sensing domains such as emotion recognition and workload estimation, where models are trained to extract subject-invariant features that generalize to unseen users \cite{apicella2024toward,qin2022domain,xiao2025human}. 
%While such methods improve robustness under certain forms of heterogeneity, they fundamentally assume that a shared mapping exists between physiological signals and target labels across individuals. 
%However, in settings such as craving detection—where physiological responses may carry different semantic meanings across individuals due to differences in stress regulation—this assumption can break down. 
%Consequently, invariant representations alone may be insufficient for capturing subject-specific interpretations of physiological signals.
%\textbf{Implication for this work.}
%We therefore include representative DG methods (e.g., IRM, GroupDRO, and V-REx) as baselines to evaluate whether invariance-based approaches can address subject-independent craving detection.

\textbf{Multimodal and auxiliary-information approaches.}
Another line of work addresses inter-individual variability by incorporating additional modalities or auxiliary information, such as behavioral signals, audio-visual cues, or personality-related features \cite{seikavandi2025mumtaffect,singh2024visiophysioenet,wu2025comprehensive,wang2023smoking}. These methods often fuse information using cross-attention \cite{vaswani2017attention}, mixture-of-experts \cite{shazeer2017outrageously}, or feature-wise modulation methods such as FiLM \cite{perez2018film}. While effective for improving prediction, they generally treat auxiliary information as additional input rather than as context that governs how physiological evidence should be interpreted. Moreover, continuous audio or visual sensing can also raise privacy, burden, and deployment concerns in sensitive populations \cite{zhang2025survey,motti2015users}. 
We therefore compare against representative multimodal fusion models, including Cross-Attention, Mixture-of-Experts, and FiLM, to test whether using resilience-related guidance as an auxiliary input is sufficient compared with explicitly conditioning physiological interpretation.

\subsection{Resilience as a Missing Factor for Interpreting Physiological Craving Signals}

Psychological resilience shapes how individuals cope with stress and adversity, particularly in substance use disorder (SUD) populations. Higher resilience is associated with lower perceived stress, better mental health outcomes, reduced relapse risk, and stronger recovery-related resources \cite{yang2020resilience,yamashita2021resilience,rathinam2021resilience}. It has also been linked to craving intensity, suggesting a role in craving vulnerability \cite{rahmati2021investigating}. Beyond a self-reported trait or outcome, resilience is increasingly viewed as a regulatory process that unfolds over time, reflecting how individuals recover from stress and how autonomic arousal resolves after challenge \cite{windle2011resilience,ungar2020resilience,den2022conceptualizing,goldstein2013differential,allen1989patterns,rogers2024addiction}. 
Because stress is a well-established trigger of craving \cite{sinha1999stress,preston2011stress,grusser2006relationship,breese2005stress}, differences in stress regulation may change how physiological responses relate to craving: the same autonomic pattern may reflect effective recovery in one individual but elevated craving vulnerability in another.

This creates a key challenge for wearable craving assessment. Short physiological windows capture momentary autonomic activity, whereas resilience reflects regulation and recovery over time \cite{den2022conceptualizing}. Standard wearable models map these short windows to labels using population-level relationships \cite{zhao2023affective,king2019micro}, assuming that similar physiological patterns have similar meanings across individuals. This assumption becomes fragile when stress and craving physiology overlap and resilience changes how physiological evidence should be interpreted.

Existing approaches to inter-individual variability, including domain generalization and multimodal fusion methods (Section~\ref{sec:DGMM}), improve robustness by learning shared representations or adding auxiliary information as predictive input. However, they do not explicitly model how resilience guides the interpretation of physiological signals. This motivates our approach: rather than treating resilience as another predictive feature, we use resilience-related guidance as a conditioning signal for physiological interpretation. We operationalize this guidance through physiological recovery dynamics and autobiographical language representations, enabling subject-aware craving inference without subject-specific retraining.

\section{User Study}
\label{sec:userstudy}

Existing wearable-sensing datasets do not jointly capture physiological dynamics, stress and craving experiences, and subject-level factors such as resilience within a single controlled setting. To address this gap, we design a laboratory-based user study that collects synchronized multimodal data for craving detection under subject-independent evaluation. 
The resulting dataset includes: (1) continuous wearable physiological signals capturing multimodal autonomic responses; (2) task-level stress labels and self-reported craving annotations, enabling craving to be studied as a subjective state within stress-related conditions; and (3) subject-level autobiographical language used to derive resilience-related representations. 
%In this design, validated stress-induction tasks are used as controlled physiological probes, as direct elicitation of craving raises ethical and practical concerns in individuals with opioid use disorder.
\iffalse
The resulting dataset consists of:
\begin{itemize}
    \item \textbf{Continuous wearable physiological signals}, capturing multimodal autonomic responses.
    \item \textbf{Task-level stress labels and self-reported craving annotations}, enabling the study of craving as a subjective state arising under stress-related conditions.
    \item \textbf{Subject-level autobiographical language}, used to derive representations of resilience.
\end{itemize}
\fi 
This unified design enables, for the first time, a controlled analysis of how physiological dynamics, subjective craving, and resilience-related individual differences interact within the same experimental setting. The details and rationale of the study design are outlined below.

\subsection{Participants and Recruitment}
\label{subsec:participants}

The study was approved by the Institutional Review Board (IRB) of \textbf{X University}, and all participants provided informed consent prior to participation. Participants were recruited from the local community and university population (control group) as well as through clinical and community partners (OUD group), and participated on a voluntary basis. Participants with OUD received a gift card valued at \textbf{XX USD} as compensation.

The dataset comprises 78 sessions, including 28 sessions from 24 individuals with OUD and 50 sessions from 50 control participants. Among OUD participants, data were collected either before or after medication for opioid use disorder (MOUD), referred to as \textit{pre-dose} and \textit{post-dose} sessions. While most participants contributed a single session, 4 individuals were recorded in both pre-dose and post-dose conditions, resulting in a total of 28 sessions from 24 individuals. To ensure strict subject-independent evaluation, these 4 individuals with repeated sessions were used exclusively for training and excluded from evaluation. The remaining participants each contributed a single session. This sample size is consistent with prior work in wearable sensing and HCI, where controlled user studies often require complex experimental protocols and multimodal data collection \cite{meegahapola2026stress,caine2016local,ezer2024somaesthetic,zhao2023affective}. These challenges are further amplified in vulnerable clinical populations such as individuals with substance use disorder, where recruitment and sustained participation are inherently more difficult \cite{gullapalli2019body,gullapalli2021opitrack}. Participant demographics are summarized in Table~\ref{tab:demographics}.

\begin{table}[t]
\centering

\resizebox{\linewidth}{!}{
\begin{tabular}{l l p{4.2cm} p{6.5cm}}
\toprule
\textbf{Group} & \textbf{Age} & \textbf{Gender} & \textbf{Race/Ethnicity} \\
\midrule
Control (n=50) & 
30.80 $\pm$ 9.60 [22--65] &
\makecell[l]{
Female: 35 (70.0\%) \\
Male: 15 (30.0\%)
} &
\makecell[l]{
White: 30 (60.0\%) \\
Asian: 16 (32.0\%) \\
Black or African American: 3 (6.0\%) \\
Other: 1 (2.0\%)
} \\
\midrule
OUD (n=24) & 
35.67 $\pm$ 4.15 [29--47] &
\makecell[l]{
Female: 18 (75.0\%) \\
Male: 5 (20.8\%) \\
Non-binary: 1 (4.2\%)
} &
\makecell[l]{
White: 14 (58.3\%) \\
Black or African American: 7 (29.2\%) \\
Other: 2 (8.3\%) \\
American Indian or Alaska Native: 1 (4.2\%)
} \\
\bottomrule
\end{tabular}

}
\caption{Participant demographics. Values are reported as mean $\pm$ SD [range] or count (\%).}
\vspace{-5mm}
\label{tab:demographics}
\end{table}

\subsection{Apparatus and Experimental Setup}

%Figure~\ref{fig:setup} illustrates the experimental setup. All study sessions were conducted in a private room, either at a university research facility or within a hospital setting, with only the participant and a trained experimenter present. This controlled and private environment was designed to ensure participants’ comfort and confidentiality, allowing them to express their emotions and subjective experiences related to stress and craving without concern for observation or interruption. Participants stood 8-12\,ft from a display screen presenting visual stimuli while wearing an Empatica~E4 wristband on both wrists. 
%The E4 continuously recorded photoplethysmography (PPG), electrodermal activity (EDA), 3-axis accelerometer data (ACC), and skin temperature (TEMP), enabling synchronized multimodal measurement of autonomic responses, which provide observable physiological correlates of stress- and affect-related arousal ~\cite{kreibig2010autonomic,stritzelberger2025validity}.

%Figure~\ref{fig:setup} illustrates the experimental setup. 
All study sessions were conducted in a private room, either at a university research facility or within a hospital setting, with only the participant and a trained experimenter present. This controlled and private environment was designed to ensure participants’ comfort and confidentiality, allowing them to express their emotions and subjective experiences related to stress and craving without concern for observation or interruption. Participants stood 8-12\,ft from a display screen presenting visual stimuli while wearing an Empatica~E4 wristband on both wrists. 
The E4 continuously recorded
%Figure~\ref{fig:setup} illustrates the experimental setup. 
%Physiological data were collected using the Empatica E4 wristband, which continuously records 
photoplethysmography (PPG), electrodermal activity (EDA), 3-axis accelerometer data (ACC), and skin temperature (TEMP). 
%These signals provide synchronized multimodal measurements of autonomic activity associated with stress and affective responses \cite{kreibig2010autonomic,stritzelberger2025validity}. All sessions were conducted in a controlled indoor environment with a trained experimenter present to ensure consistency across participants.

\iffalse
\begin{figure}[ht]
    \centering
    \includegraphics[width=0.80\linewidth]{setup.png}
    \caption{
Experimental setup for the user study. 
Sessions were conducted in a private indoor environment with only the participant and a trained experimenter present. 
Participants stood 8--12\,ft from a display presenting experimental stimuli while wearing Empatica E4 wristbands on both wrists.
}
    \label{fig:setup}
\end{figure}
\fi 

\begin{figure}[ht]
    \centering
    \includegraphics[width=0.8\linewidth]{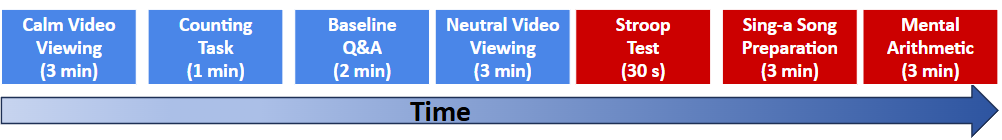}
 
    \caption{Task sequence and timeline used in the user study.
    Blue blocks represent non-stress baseline tasks, and red blocks
    represent stress-inducing tasks.}
    \vspace{-5 mm}
    \label{fig:flow}
\end{figure}

\subsection{Study Procedure}\label{datacollectionprotocol}

\subsubsection{Study Protocol Justification:} Our protocol is motivated by prior SUD laboratory studies \cite{sinha2000psychological, fox2008enhanced} showing that stress exposure can serve as a controlled cue for eliciting craving-related subjective and physiological responses \cite{sinha2009modeling}. Although in-the-wild craving data are important for ecological validity, relying on them can introduce substantial label noise. Craving episodes are often brief, self-reports may be delayed or missed, contextual triggers are uncontrolled, and physiological signals can be confounded. As a result, the temporal alignment between wearable physiology and reported craving can be imprecise, making it difficult to obtain reliable labels for model development. Reliable annotation is especially important in this work because, to our knowledge, this is the first study to examine whether trait-level subject information can modulate the interpretation of wearable physiological patterns for improved craving inference. To study this relationship carefully, we require a setting where physiological responses and craving reports can be collected with stronger temporal control.

Because prior studies \cite{anderson2018ethical} showed that direct opioid craving induction through a period of abstaining from opioid use may raise ethical, clinical, and standardization concerns, we use validated stress-induction tasks as reproducible physiological probes. 
At the same time, we do not treat stress as a proxy label for craving: stress is defined by task structure, while craving is measured independently through self-report after each task block. This design allows us to study craving-related physiology in a controlled and ethically appropriate setting while reducing label noise and preserving a clear distinction between stress exposure and subjective craving.

Notably, given the importance of ecological validity, we further conduct an in-the-wild feasibility study, described in Section~\ref{sec:inthewild}.

%\textcolor{red}{Following prior addiction research that experimentally induces craving using stress- and cue-based paradigms 
%\cite{sinha2000psychological, fox2008enhanced}, we design a laboratory protocol in which participants are exposed to structured experimental conditions consisting of non-stress baseline tasks followed by stress-inducing tasks.In these paradigms, stress is used as a controlled experimental probe to increase the likelihood of craving responses. Prior work has shown that exposure to stress reliably elevates subjective craving\cite{sinha2009modeling}. Importantly, craving is not deterministically elicited by stress.  Instead, it emerges as a subjective and variable response that depends on individual differences and contextual factors. Accordingly, we do not treat task identity as a proxy label for craving. Rather, craving is measured independently via self-report, allowing it to arise under both stress and non-stress conditions. 
%This design decouples experimental conditions from target labels, and introduces an inherently ambiguous mapping between physiological signals and craving. Such ambiguity motivates the need for modeling approaches that account for subject-dependent interpretation of physiological responses.}

\subsubsection{Experimental Design and Task Protocol}\label{datacollectionprotocol}
Following prior work on laboratory-based affect and stress elicitation \cite{xiao2024reading,zhao2023affective,almazrouei2023method}, the study protocol, as outlined in Figure~\ref{fig:flow}, consisted of multiple phases or experimental blocks with interleaved intervals, beginning with non-stress baseline tasks and followed by stress-inducing tasks targeting cognitive, emotional, and social-evaluative stress systems. This structure allowed us to establish individual physiological baselines while exposing all participants to comparable stressors, keeping task-induced stress separate from the self-reported craving labels used in later analyses.

\subsubsection*{Non-Stress Tasks}\label{non-stress-tasks}

Participants first completed a set of non-stress tasks designed to establish a relaxed autonomic baseline prior to stress induction:

\begin{enumerate}
    \item \textbf{Calm video viewing (3 minutes).}
    Participants watched a calming video depicting jellyfish floating in the ocean, which has been shown to elicit a relaxed physiological state \cite{xiao2024reading,almazrouei2023method}.

    \item \textbf{Counting task (1 minute).} Participants counted aloud from 0 to 59 at a self-selected, natural pace, a task commonly used to maintain engagement while preserving a non-stressful condition \cite{sharma2022psychophysiological,tumanova2019autonomic}.

    \item \textbf{Baseline question-and-answer task (2 minutes).} Participants responded to a sequence of neutral, everyday questions (e.g., their name, whether they had eaten breakfast, or their favorite restaurant) in a relaxed setting.

     \item \textbf{Neutral video viewing (3 minutes).}
    Participants watched a National Geographic documentary on pyramids, serving as a neutral, low-arousal baseline condition.

\end{enumerate}

These tasks served as control conditions for comparison with physiological and behavioral responses observed during subsequent stress-inducing tasks. Calm video viewing is intended to actively induce a relaxed state, whereas neutral video viewing provides a low-arousal baseline without explicitly promoting relaxation.

\subsubsection*{Stress-Inducing Tasks}

Participants then completed a set of validated stress-induction tasks adapted from prior psychology and behavioral research \cite{schaefer2010assessing,tulen1989characterization, kirschbaum1993trier,connolly2018negative}:

\begin{enumerate}
    \item \textbf{Stroop task \cite{golden1978stroop,pehlivanouglu2005computer,karthikeyan2014analysis} (3 minutes).} Participants viewed color words whose ink color differed from their semantic meaning and were instructed to verbally identify the ink color. The stimulus presentation rate gradually increased over the task duration, inducing cognitive stress through time pressure and interference.

    \item \textbf{Sing-a-Song Stress Test (30-second preparation).} Participants were unexpectedly informed that they would be asked to sing aloud while being observed and were given 30 seconds to prepare \cite{xiao2024reading}. Consistent with prior work, only the preparation phase was analyzed, as it reliably elicits social-evaluative stress \cite{almazrouei2023method}.

    \item \textbf{Mental arithmetic task (3 minutes).} Participants performed serial subtraction by counting backward from 1000 in increments of 17. Upon making an error, they were instructed to restart from 1000, following the protocol of the Trier Social Stress Test \cite{kirschbaum1993trier}.
\end{enumerate}

\iffalse
Participants first completed a sequence of non-stress baseline tasks designed to elicit different forms of relaxed autonomic states, including passive viewing of calm video scenes with background music (ocean footage with relaxing music), a simple counting task (counting from 0 to 59), a neutral question-and-answer segment (weather and daily routines), and neutral video viewing (National Geographic content).

Participants then completed a set of stress-inducing tasks targeting complementary stress dimensions, including the Stroop task (cognitive stress) \cite{golden1978stroop,pehlivanouglu2005computer,karthikeyan2014analysis}, sing-a-song preparation (social-evaluative stress) \cite{brouwer2014new}, and mental arithmetic (performance and time-pressure stress) \cite{kirschbaum1993trier}. The overall task sequence and timing are illustrated in Figure~\ref{fig:flow}.
\fi

\subsubsection{Stress and Craving Annotations}
\label{sec:annotation}

Stress labels were defined by task structure (baseline vs.\ stress blocks), leveraging established stress-induction protocols that elicit reliable and reproducible physiological responses \cite{ollander2016comparison,menghini2019stressing,zhao2023affective}.
Craving, in contrast, was assessed independently through self-report because it is a subjective motivational state that cannot be determined solely from task identity \cite{maclean2019stress,sinha2000psychological,fox2008enhanced,sinha2009modeling}. Although stress-inducing tasks may create conditions in which craving can arise, craving is not assumed to occur during every stress block and may also occur during non-stress periods, as shown in Section~\ref{data-charact}. 
%was assessed independently via self-report, as it is a subjective motivational state that cannot be deterministically specified by task structure \cite{maclean2019stress,sinha2000psychological, fox2008enhanced,sinha2009modeling}. Although stress-inducing tasks create conditions under which craving may arise, craving can also occur during non-stress periods and is therefore not assumed based on task identity (as evidenced in Section \ref{data-charact}.
Following each task block, participants with OUD were asked to rate their perceived opioid \emph{craving} on a 0–3 Likert scale (0 = not at all, 3 = extremely). For binary classification, craving labels were derived by thresholding self-reports, with ratings of 0 mapped to non-craving and ratings of 1–3 mapped to craving. Control participants were not asked about opioid craving.

Because craving can persist over short time intervals rather than occurring only at a single instant \cite{baillet2024craving,chirokoff2023craving,grunevski2024predictive}, each post-task craving report was treated as reflecting the participant's craving state over the corresponding task interval. This allowed us to assign craving labels to physiological windows within that interval while keeping craving defined as a participant-reported subjective state.

\subsection{Autobiographical Memory and Context-neutral Speech Collection}
\label{sec:Autobiographical}

As part of the study procedure, participants provided short spoken narratives to derive subject-level autobiographical language representations. At the beginning of the session, participants first completed a context-neutral speech task, responding to simple prompts (e.g., their name, daily routines, or general questions) for two minutes. This segment captures general linguistic characteristics without autobiographical or emotional content.

Participants then described one recent positive and one recent negative personal experience, each for two minutes, in a neutral, non-stressful setting independent of the stress-induction tasks. All speech segments were audio-recorded and transcribed. On average, context-neutral responses contained approximately 160 words, while autobiographical narratives contained approximately 210 words each (approximately 420 words per participant).

\section{ Dataset and Feature Construction}
\label{sec:signal_processing}

Physiological signals were processed using a unified pipeline to ensure consistent analysis across participants. We describe preprocessing, windowing, and feature extraction, followed by higher-level feature organization and dataset characterization used in subsequent analyses.

\subsection{Physiological Signal Processing}
\label{sec:feature}

Physiological signals collected from the Empatica~E4 wristband \cite{mccarthy2016validation}, including EDA, PPG, 3-axis ACC, and skin TEMP, were temporally synchronized using device timestamps. EDA and PPG signals were processed using the NeuroKit toolkit~\cite{makowski2021neurokit2}, including denoising and decomposition into tonic and phasic components for EDA, and artifact removal with heart rate and inter-beat interval extraction for PPG. ACC and TEMP signals were aligned to ensure cross-modal consistency.
Physiological recordings were segmented into overlapping windows of 30~seconds with a step size of 15~seconds \cite{dehghani2019quantitative,sharma2022psychophysiological}. This window length balances temporal sensitivity with signal stability and is widely used in short-term autonomic state modeling. Each window is treated as an independent observation. For task-level analyses in Section \ref{sec:rq1}, window-level features are aggregated within each task interval.

\textbf{Feature Extraction.}
For each window, we extracted multimodal physiological features characterizing autonomic activity from EDA, PPG-derived cardiovascular signals, acceleration (ACC), and skin temperature (TEMP). EDA features include SCR-based measures (onsets, peaks, amplitudes, and recovery dynamics) together with tonic/phasic summary statistics, while cardiovascular features capture heart rate and HRV in time and frequency domains. Across modalities, additional FLIRT-style descriptors~\cite{foll2021flirt} were computed, including summary statistics (e.g., mean, median, standard deviation, variance, interquartile range, minimum, maximum, and percentile summaries), shape descriptors (e.g., RMS, mean absolute value, peak amplitude, waveform length, crest factor, and zero crossings), and  spectral descriptors (e.g., total spectral power, spectral centroid, peak frequency, bandwidth, spectral entropy, and band-limited power features). Low-quality features were removed based on missingness and low variability, yielding a final 303-dimensional window-level representation used for all models. Additional details are provided in Appendix~\ref{appendix:feature}.

\begin{figure}[ht]
    \centering

    \begin{minipage}{0.59\linewidth}
        \centering
        \includegraphics[width=\linewidth]{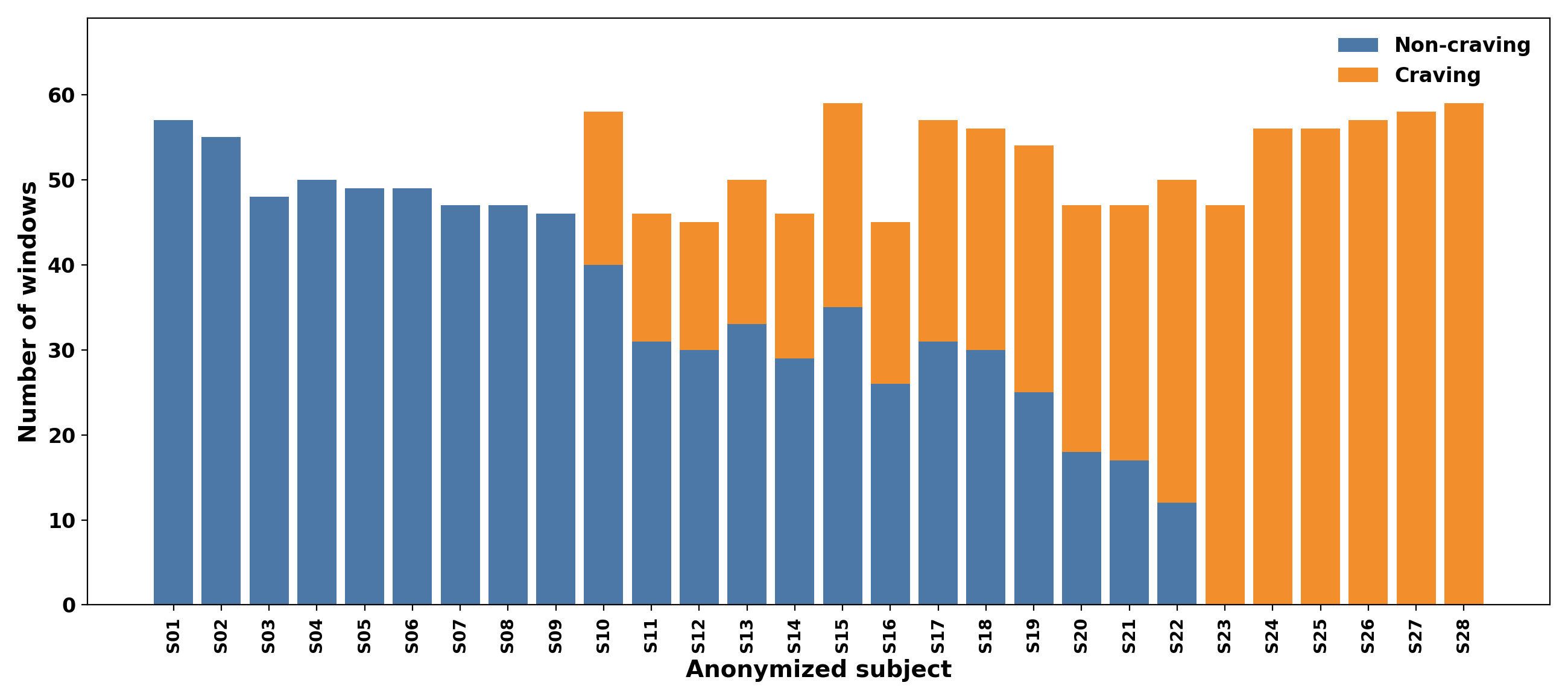}
        \caption{Subject-level distribution of craving and non-craving windows among OUD participants. Each bar corresponds to one anonymized subject and is ordered by the proportion of craving windows.}
        \label{fig:subject_craving_dist}
    \end{minipage}
    \hfill
    \begin{minipage}{0.40\linewidth}
        \centering
        \includegraphics[width=\linewidth]{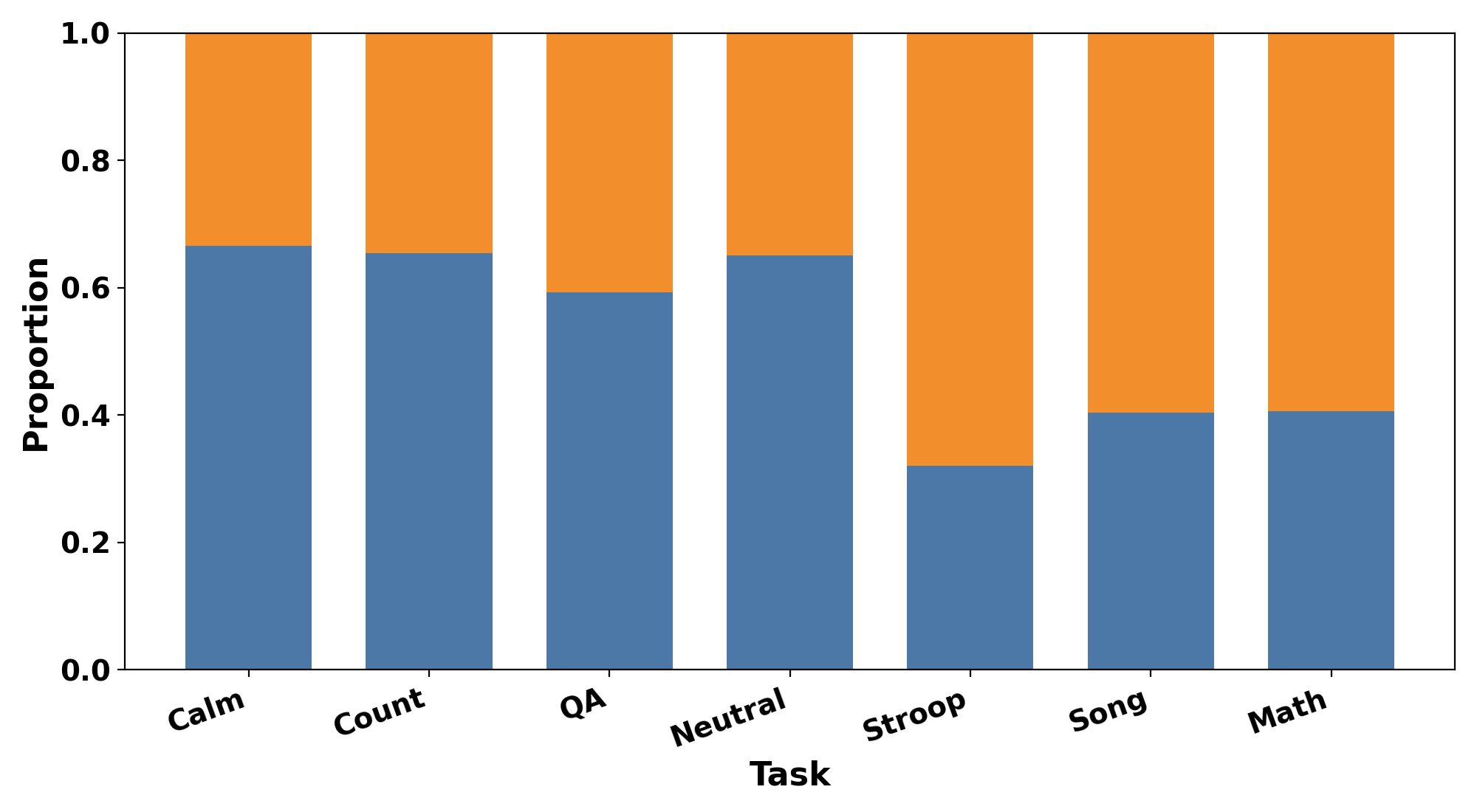}
        \caption{Distribution of craving and non-craving window ratio among each tasks.}
        \label{fig:task_craving_dist}
    \end{minipage}
    
\end{figure}

\subsection{Dataset Characteristics}\label{data-charact}

This section characterizes the distribution of physiological windows and craving labels in the collected dataset. This analysis is important because craving is a subjective state: even under the same task protocol, participants may differ in whether craving occurs, how often it occurs, and during which task blocks it is reported.

Figure~\ref{fig:subject_craving_dist} summarizes subject-level variation in physiological window availability and craving labels. While the number of windows is broadly comparable across participants, the distribution of craving labels varies substantially. This highlights strong inter-subject variability in craving experience, even under controlled laboratory conditions.
Figure~\ref{fig:task_craving_dist} shows the distribution of craving and non-craving labels across tasks. Although stress-inducing tasks are associated with a higher prevalence of craving, craving does not occur deterministically under stress and is also observed during non-stress conditions. This confirms that stress and craving should not be treated as equivalent labels: stress is defined by task structure, whereas craving remains a participant-reported subjective state.

Together, these patterns show that craving labels are heterogeneous across both participants and tasks. The dataset, therefore, presents two key challenges for wearable craving detection: uneven craving prevalence across individuals and non-uniform relationships between stress exposure and reported craving. These challenges motivate subject-independent evaluation and modeling approaches that can account for individual differences in how physiological signals relate to craving.

\section{Empirical Motivation: Physiological Craving Signals and Resilience Proxies}
\label{Emperical-motivation}

Before introducing the \textbf{\emph{RETRACE}} framework, this section provides the empirical and literature-backed rationale for our modeling design. We first characterize how stress, craving, and resilience are reflected in short-term wearable physiology. The analysis shows that stress produces robust autonomic responses, craving-related signals are weaker and more heterogeneous, and resilience is not directly observable from short-term physiology. We then describe HR-AUC as a physiology-grounded resilience proxy and evaluate whether autobiographical recall can provide a lower-burden resilience-related representation. Together, these analyses and discussions motivate \textbf{\emph{RETRACE}}’s design: using stress-pretrained physiological structure as a backbone while conditioning craving inference on reusable subject-level guidance from resilience-related proxies.

%\subsection{RQ1: Do Stress, Craving, and Resilience Exhibit Distinct Physiological Signatures in Short-Term Wearable Data?}

%Building on the dataset described above, we examine whether stress, craving, and psychological resilience can be identified from short-term wearable physiological signals collected during experimental tasks. This question is central to craving detection, where inference must rely on moment-to-moment physiological observations. While stress is known to produce robust and detectable physiological responses, it remains unclear whether craving and psychological resilience exhibit similarly distinguishable signatures at this temporal scale.

\subsection{{Characterizing Stress, Craving, and Resilience in Short-Term Wearable Physiology}}
\label{sec:rq1}
This section analyzes whether stress, craving, and resilience exhibit distinguishable physiological patterns in short-term wearable physiological signals. This question is central to wearable craving detection because deployed models typically operate on short physiological windows and must determine whether those windows contain craving-relevant information that can generalize across individuals. Stress is expected to produce robust autonomic responses and therefore serves as a useful reference condition. 
Resilience, however, is a relatively stable trait-level factor rather than a momentary state, so we examine whether it appears in instantaneous physiology or instead requires reusable subject-level descriptors.

\subsubsection{Analysis overview.}
Following prior work on task-level physiological analysis from wearable sensors~\cite{zhao2023affective,king2019micro,xiao2025human}, we aggregate window-level features to the task-block level because stress and craving annotations were collected after each task rather than at individual window timestamps. Task-level features were baseline-corrected using each participant's neutral \emph{Calm video viewing} task (Section \ref{datacollectionprotocol}) as a reference~\cite{hui2025adjusted,iqbal2022stress,chatzaki2025overview}.

We then estimated task-level physiological associations using population-average generalized estimating equations (GEE), clustering by participant to account for repeated observations within individuals~\cite{liang1986longitudinal}. Stress and craving are defined based on the annotation protocol described in Section~\ref{sec:annotation}. Resilience is operationalized as a subject-level variable based on post-stress heart-rate recovery, quantified by HR-AUC, and discretized into high- and low-resilience groups using a median split for the present analysis. The rationale for this proxy and the grouping procedure are detailed in Section~\ref{sec:resilience_def} and Appendix~\ref{app:resilience_group}. For each construct (stress, craving, and resilience), we fit separate GEE models with the construct as the main predictor, adjusting for subject group and the number of aggregated windows per task block. The estimated coefficients ($\beta$) represent population-average associations between the construct of interest and physiological features. To account for multiple comparisons across features, we apply the Benjamini--Hochberg false discovery rate (FDR) correction \cite{waite2006controlling}. All reported significance results are based on FDR-corrected p-values.
We analyze physiological modulation at two complementary levels. 

\begin{figure}[ht]
    \centering
    \includegraphics[width=\linewidth]{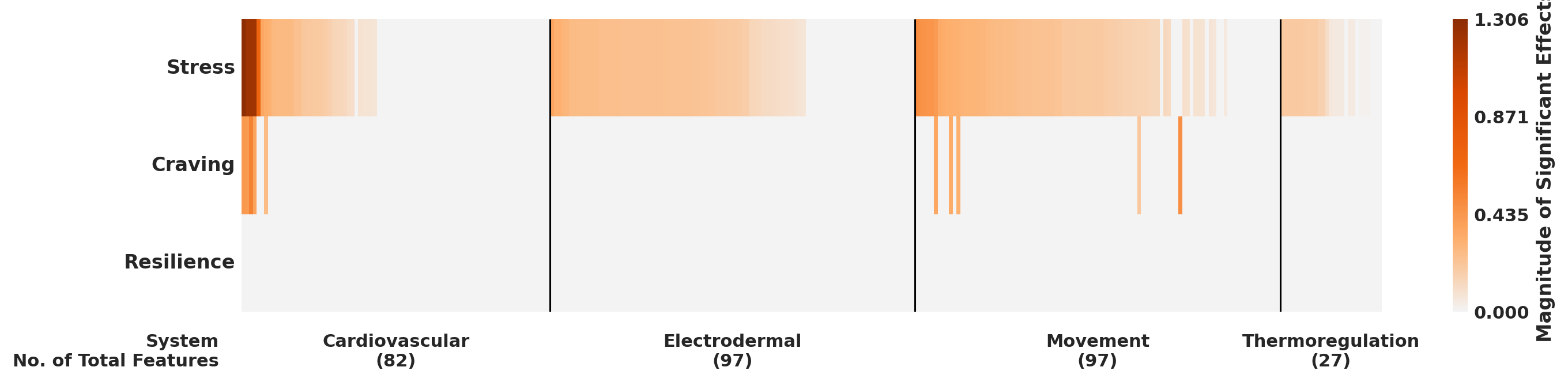}
   \caption{ Feature-level physiological effects across modalities. Each column corresponds to an individual physiological feature, grouped by system and ordered within each group by the magnitude of stress-related effects. Color intensity represents the magnitude of significant effects (|$\beta$|) estimated by GEE models, while non-significant features are masked. Rows correspond to different constructs (stress, craving, and resilience). }
    \vspace{-3mm}
    \label{fig:rq1_feature_heatmap}
    
\end{figure}

\emph{At the system level,} we capture coarse-grained modality-level effects by grouping features within four physiological systems—\textbf{\emph{Cardiovascular, Electrodermal, Movement, and Thermoregulation}}—and computing PCA-based representations within each system, followed by GEE analysis on these representations (see Appendix~\ref{feature:system} for grouping details and rationale).
\emph{At the feature level,} we apply the same GEE framework directly to individual features (total $303$) to capture fine-grained, feature-specific modulation.
System-level results are summarized in Table~\ref{tab:rq1_system_summary}, while Figure~\ref{fig:rq1_feature_heatmap} visualizes feature-level effects across all features. In the figure, color intensity represents the magnitude of significant effects (|$\beta$|), and non-significant features are masked.

\subsubsection{Stress-related physiological modulation.}
Stress elicited robust and widespread physiological responses.

At the feature level, stress was associated with increased variability and activation across multiple modalities. As shown in Figure~\ref{fig:rq1_feature_heatmap}, these effects are widespread, with a large number of features exhibiting significant modulation.

At the system level, these effects are consistent across all physiological modalities. As shown in Table~\ref{tab:rq1_system_summary}, stress produces strong and statistically significant shifts across cardiovascular, electrodermal, movement, and thermoregulatory systems (all $p<10^{-3}$).

Overall, these results confirm that the protocol elicited strong task-evoked autonomic modulation and that stress provides a reliable physiological reference condition.

%Stress elicited robust and widespread physiological responses. At the feature level, stress was associated with increased variability and activation across multiple modalities. Cardiovascular responses showed greater dispersion of heart rate, including increased heart-rate interquartile range ($\beta=0.29$, $p=4.2\times10^{-12}$) and heart-rate standard deviation ($\beta=0.30$, $p=2.3\times10^{-11}$). Electrodermal activity increased through larger peak-to-peak amplitude of phasic EDA signals ($\beta=0.40$, $p=6.7\times10^{-12}$), and movement responses intensified through higher peak-to-peak acceleration magnitude ($\beta=0.31$, $p=1.1\times10^{-8}$).

%These effects were also visible at the system level. As shown in Table~\ref{tab:rq1_system_summary}, stress produced large and highly significant shifts in electrodermal ($\beta=3.81$, $p<10^{-9}$), movement ($\beta=2.86$, $p<10^{-10}$), and thermoregulatory systems ($\beta=1.49$, $p<.001$). Cardiovascular effects were significant in mean-based composites but weaker in the PCA representation ($\beta=0.60$, $p=.18$), suggesting greater inter-individual variability in average heart-rate responses. Overall, these results confirm that the protocol elicited strong task-evoked autonomic modulation and that stress provides a reliable physiological reference condition.

\subsubsection{Craving-related physiological modulation.}
Compared with stress, craving-related physiological effects were weaker and more selective. At the feature level, craving-related modulation is sparse. As shown in Figure~\ref{fig:rq1_feature_heatmap}, only a small subset of features exhibit significant effects. After FDR correction \cite{waite2006controlling}, only 10 out of 303 features are significant, compared with 200 under stress. Notably, 9 out of these 10 craving-related features are also significant under stress, indicating a strong overlap between craving- and stress-related physiological responses, while a small number of features exhibit craving-specific effects. 

At the system level, these effects are limited and less consistent across modalities. As shown in Table~\ref{tab:rq1_system_summary}, only the thermoregulatory system shows a significant effect, while other systems do not reach statistical significance. Interestingly, although no individual thermoregulatory features are significant at the feature level, the system-level analysis reveals a significant effect. This discrepancy arises because aggregation captures multiple weak but consistent effects across features. While no single feature is strong enough to survive FDR correction, their combined variation forms a detectable pattern at the system level, indicating that craving-related thermoregulatory modulation is distributed and low-amplitude rather than driven by a single dominant feature.

Overall, these results indicate that craving has measurable physiological correlates, but these effects are weak, sparse, and largely embedded within broader stress-related activity, making craving difficult to distinguish from non-craving using short-term physiological signals alone.

%Craving-related physiological modulation. Compared with stress, craving-related physiological effects wereweaker and more selective. At the system level, craving showed a significant effect only in the thermoregulatory PCA representation ($\beta$ = 1.09, SE= 0.53, p = .038), while electrodermal modulation approached significance ($\beta$ = 1.44, p = .093). Cardiovascular and movement systems showed smaller effects with larger uncertainty and did not reach significance. The feature-level results showed the same pattern. After Benjamini–Hochberg False Discovery Rate correction (FDR) correction, only 10 features showed significant craving-related modulation, compared with 197 significant features under stress. Moreover, all 10 craving-related features were also significant under stress. This indicates that craving has measurable physiological correlates, but these correlates are sparse and largely embedded within broader stress-related autonomic activity. Thus, craving does not appear as a strong, clearly separable physiological state in short-term wearable data, making it difficult to distinguish craving from non-craving using physiology alone.
\begin{table}[t]
\centering

\resizebox{\linewidth}{!}{
\begin{tabular}{lccc}
\toprule
\textbf{Physiological System} 
& \textbf{Stress} 
& \textbf{Craving} 
& \textbf{Resilience} \\
\midrule
Cardiovascular 
& $\beta{=}0.946$, SE{=}0.284, $p{<}10^{-3}$ 
& $\beta{=}0.380$, SE{=}0.685, $p{=}0.579$ 
& $\beta{=}{-}0.917$, SE{=}1.173, $p{=}0.434$ \\

Electrodermal 
& $\beta{=}3.317$, SE{=}0.673, $p{<}10^{-6}$ 
& $\beta{=}2.977$, SE{=}1.816, $p{=}0.101$ 
& $\beta{=}3.525$, SE{=}2.164, $p{=}0.103$ \\

Movement 
& $\beta{=}2.798$, SE{=}0.471, $p{<}10^{-9}$ 
& $\beta{=}1.478$, SE{=}1.462, $p{=}0.312$ 
& $\beta{=}1.504$, SE{=}1.497, $p{=}0.315$ \\

Thermoregulation 
& $\beta{=}1.314$, SE{=}0.289, $p{<}10^{-5}$ 
& $\beta{=}1.071$, SE{=}0.484, $p{=}0.0269$ 
& $\beta{=}{-}0.962$, SE{=}0.650, $p{=}0.139$ \\
\bottomrule
\end{tabular}}
\caption{System-level physiological modulation. Reported coefficients ($\beta$) summarize population-average task effects estimated by the GEE models. Positive $\beta$ values indicate higher physiological activity under the task condition, while negative values indicate reduced activity. Larger absolute $\beta$ values, when observed consistently across related features or physiological modalities, reflect stronger and more coherent task-evoked physiological modulation. Standard errors (SE) quantify uncertainty in these estimates, and $p$-values indicate whether the observed effects differ reliably from zero after accounting for within-subject dependence. } 
 \vspace{-10mm}
\label{tab:rq1_system_summary}
\end{table}

\subsubsection{Resilience-related physiological modulation.}
In contrast to stress and craving, resilience did not show a robust short-term physiological signature.

At the feature level, resilience-related effects do not exhibit a consistent pattern across features, as reflected in Figure~\ref{fig:rq1_feature_heatmap}. At the system level, resilience effects show substantial variability across modalities, with relatively large but highly uncertain estimates, and none reaching statistical significance (Table~\ref{tab:rq1_system_summary}).

These results suggest that psychological resilience does not manifest as a stable task-evoked pattern that can be reliably detected from short-term wearable windows. Rather than treating resilience as an instantaneous physiological state, this motivates modeling it as a subject-level factor. This finding supports the use of reusable resilience-related descriptors, such as post-stress HR-AUC and autobiographical recall, to guide craving inference.

%Resilience-related physiological modulation. In contrast to stress and craving, resilience did not show a robust short-term physiological signature. At the system level, resilience effects were uniformly small across all modalities (all |p| < 0.12), with non-significant p-values across cardiovascular, electrodermal, movement, and thermoregulatory systems (Table 2). This pattern was consistent across both mean-based and PCA-based system representations. These results suggest that psychological resilience does not manifest as a stable task-evoked pattern that can be reliably detected from short-term wearable windows. Rather than treating resilience as an instantaneous physiological state, this motivates modeling it as a subject-level factor. This finding supports the use of reusable resilience-related descriptors, such as post-stress HR-AUC and autobiographical recall, to guide craving inference.

\paragraph{\textbf{Summary and design implications}}
This section's analysis shows that craving has measurable population-level physiological correlates, supporting the feasibility of subject-independent wearable craving detection. However, these correlates are weak, selective, and largely embedded within broader stress-related activity, with limited distinct or craving-specific components, which reduces the effectiveness of a single population-level mapping from physiology to craving. In contrast, resilience is not directly detectable from short-term wearable windows. These findings motivate \textbf{\emph{RETRACE}}’s design: using stress-pretrained representations as a stable physiological backbone and using resilience-related subject-level guidance to interpret ambiguous craving-related signals across individuals. As outlined in Section \ref{Model-archi}, the stress-pretrained encoder provides a shared autonomic reference, while craving is learned separately through resilience-conditioned inference task that identifies craving-related deviations without treating stress itself as craving.

%Summary and design implications. Overall, this section’s analysis shows that craving has measurable but weak and stress-embedded physiological correlates, while resilience is not directly detectable from shortterm wearable windows. These findings motivate RETRACE’s design: using stress-pretrained representations as a stable physiological backbone and using resilience-related subject-level guidance to interpret ambiguous craving-related signals across individuals. As outlined in Section 6.2, the stress-pretrained encoder provides a shared autonomic reference, while craving is learned separately through resilience-conditioned inference task that identifies craving-related deviations without treating stress itself as craving

\subsection{Post-Stress Heart-Rate Recovery as a Resilience Proxy}
\label{sec:resilience_def}

\begin{figure}[htbp]
  \centering
  \begin{minipage}[t]{0.4\textwidth}
    \centering
    \includegraphics[width=\textwidth]{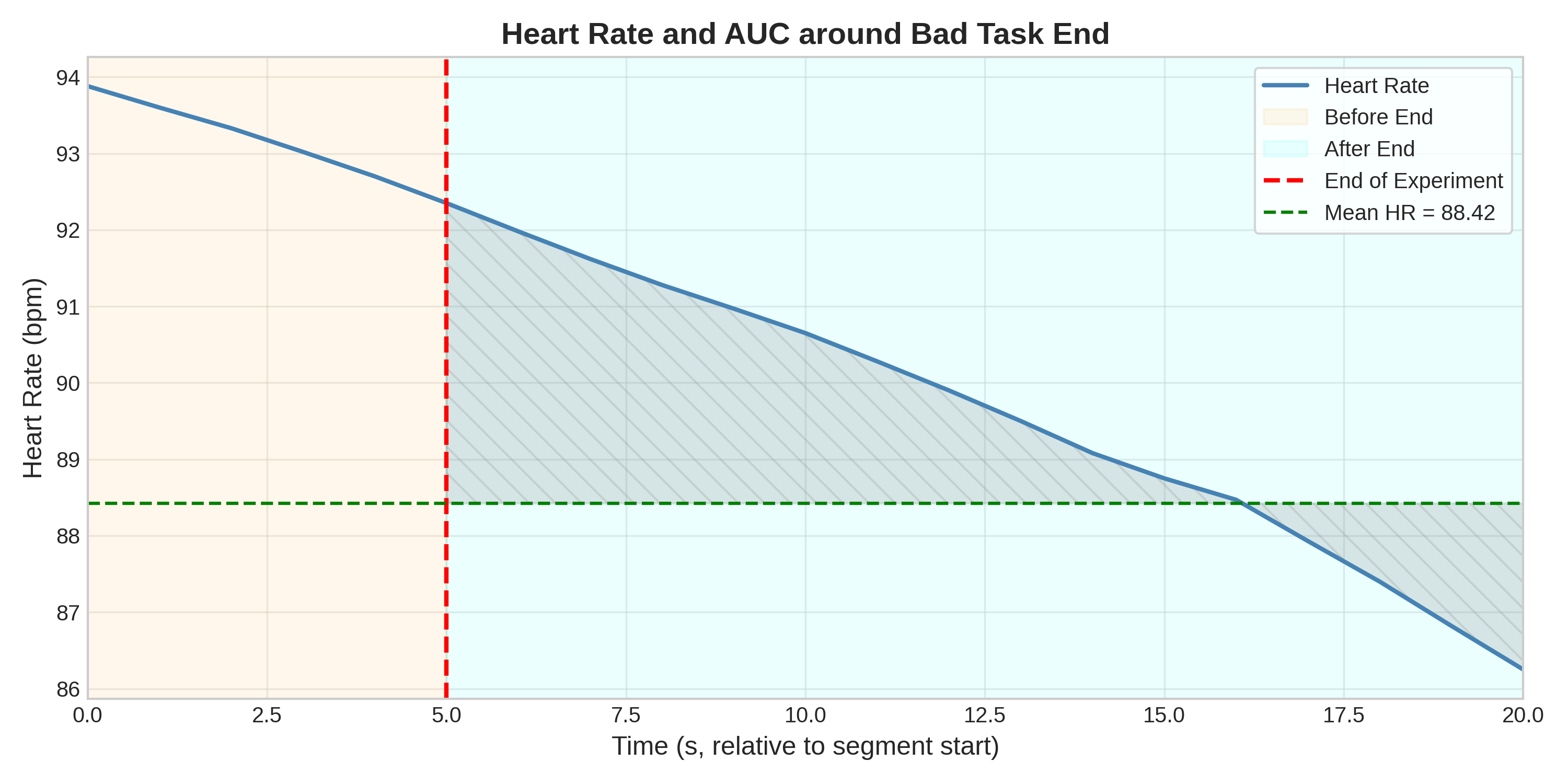}
    \caption{Heart rate recovery AUC of participant 1. Smaller AUC indicates faster recovery (higher resilience).}
    \label{fig:AUC1}
  \end{minipage}
  \hfill
  \begin{minipage}[t]{0.4\textwidth}
    \centering
    \includegraphics[width=\textwidth]{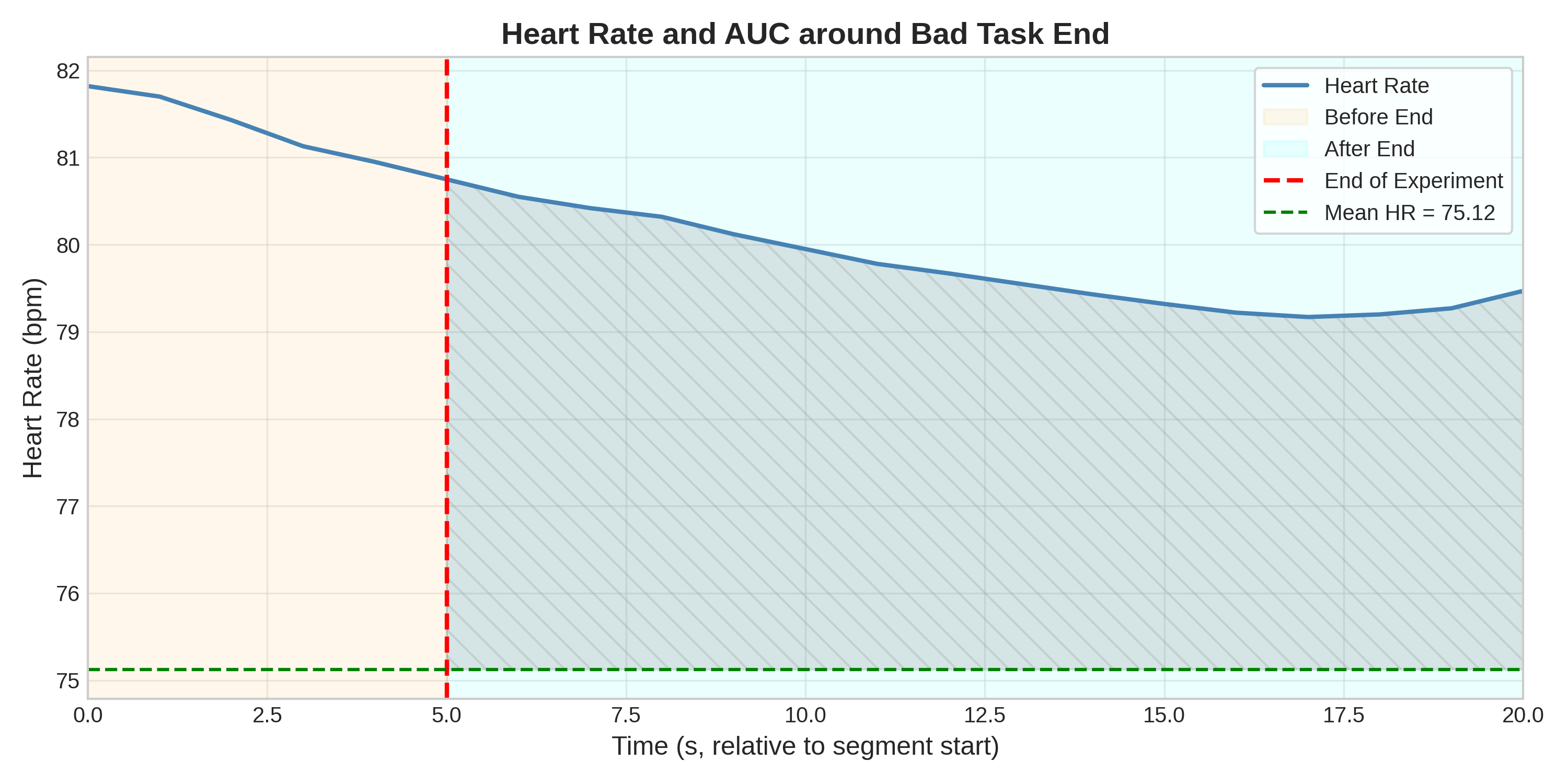}
    \caption{Heart rate recovery AUC of participant 2. Larger AUC indicates slower recovery (lower resilience).}
    \label{fig:AUC2}
  \end{minipage}
\end{figure}

The preceding analysis confirms that resilience is not reliably detectable from short-term instantaneous wearable physiological signal windows. Following literature~\cite{den2022conceptualizing}, we therefore use post-stress heart-rate recovery as a subject-level proxy for resilience-related regulatory capacity. Specifically, we compute the area under the curve (AUC) of heart rate during the post-stress recovery period, defined as the interval following the mental arithmetic task in the protocol described in Section~\ref{datacollectionprotocol}. Smaller HR-AUC values indicate faster recovery toward baseline and are interpreted as higher physiological resilience, whereas larger HR-AUC values indicate slower recovery and are interpreted as lower physiological resilience.

Figures~\ref{fig:AUC1} and~\ref{fig:AUC2} illustrate two example recovery trajectories. In these examples, the participant with the smaller HR-AUC shows faster post-stress recovery, while the participant with the larger HR-AUC shows more prolonged physiological activation. HR-AUC is suitable for our analysis because it summarizes integrated autonomic recovery dynamics, including sympathetic withdrawal and parasympathetic reactivation~\cite{den2022conceptualizing}. We use this measure as a stable, interpretable, subject-level resilience proxy for subsequent analysis and resilience-guided craving inference.

\subsection{Autobiographical Recall as a Lower-Burden Resilience Proxy}
\label{sec:rq2_alignment}

The previous section defines HR-AUC as a physiology-grounded proxy for resilience-related recovery. However, HR-AUC requires controlled stress exposure, a clearly defined recovery period, and stable baseline conditions. Although it can be collected once as a pre-deployment measure and reused during wearable-based craving inference for a target user, it still requires exposing the individual to a controlled stressor. This may add burden during practical deployments. We therefore examine whether autobiographical recall can provide a lower-burden representation of resilience-related individual differences.

\subsubsection{{Motivation for Autobiographical Recall}} 
Autobiographical memory recall is relevant to resilience because it captures how individuals appraise, organize, and regulate emotionally meaningful experiences. Resilience is commonly conceptualized as the capacity to adapt to, recover from, and maintain functioning following adversity~\cite{den2022conceptualizing,troy2023psychological}. This capacity is closely linked to emotion regulation, coping, affective flexibility, and stress-recovery processes~\cite{schetter2011resilience,aldao2010emotion,troy2011resilience,compas2017coping}. Because autobiographical narratives reflect how individuals interpret past experiences, construct personal meaning, and regulate emotional responses during recall, they may provide a stable subject-level window into resilience-related individual differences~\cite{conway2000construction,mcadams2001psychology,adler2016incremental}.

We consider both positive and negative autobiographical memories because they capture complementary aspects of resilience-related processing. 
Positive recall may reflect adaptive affective engagement, reward sensitivity, and access to positive self-relevant experiences, which are important for coping and psychological recovery~\cite{tugade2004resilient,fredrickson2001role}. This is particularly relevant in OUD, where chronic opioid exposure can disrupt reward and stress systems, producing accumulated allostatic load, reduced sensitivity to non-drug rewards, and heightened stress reactivity~\cite{koob2007stress,koob2008addiction,koob2013addiction,koob2014addiction}. Under this reward-deficit and stress-surfeit framework, positive autobiographical recall may not simply evoke pleasure; for some individuals, a mismatch between expected and experienced positive affect may instead produce ambivalence, anxiety, or physiological arousal~\cite{koob2014addiction,koob2008addiction,qu2025nucleus,schultz2016dopamine}. Thus, positive recall may encode individual differences in reward responsiveness, stress reactivity, and craving vulnerability.

Negative autobiographical recall, in contrast, directly engages adverse experiences and therefore provides a natural context for examining emotion regulation and stress recovery. Prior work shows that regulation strategies such as cognitive reappraisal shape how negative memories are recalled and experienced, including their detail, emotional tone, and negative bias~\cite{colombo2021moderating,pascuzzi2017emotion}. The ability to suppress or control intrusive memory retrieval is also considered an important component of emotion regulation and can reduce the emotional impact of unwanted recollections~\cite{engen2018memory,katsumi2020suppress}. 
Therefore, negative autobiographical narratives may reveal how individuals organize, reinterpret, and regulate adverse experiences.
A detailed neurophysiological context motivation discussion is further provided in Appendix \ref{sec:rq2_context}.

\subsubsection{Representational Alignment Analysis}
\label{sec:rq2_alignment}
To test whether autobiographical language captures resilience-related information, we evaluate its alignment with HR-AUC, the physiology-grounded recovery proxy defined in Section~\ref{sec:resilience_def}. We compare \emph{Autobiographical Language}, derived from participants' positive and negative memory recall transcripts, with \emph{context-neutral speech}, collected at the beginning of the study as outlined in Section \ref{sec:Autobiographical}. Context-neutral speech serves as a control condition for general linguistic properties such as verbosity and speaking style without autobiographical or emotional content.

High-dimensional language embeddings extracted from these narratives were treated as fixed representations and summarized using Partial Least Squares (PLS) \cite{cha1994partial}. Because text embeddings capture heterogeneous semantic and stylistic information, unsupervised methods such as PCA may emphasize dominant linguistic variation rather than resilience-relevant dimensions~\cite{abdi2010principal}. PLS instead provides a supervised dimensionality reduction approach that identifies latent directions where language representations covary with an external outcome, making it suitable for representation--outcome alignment~\cite{abdi2010partial,krishnan2011partial}. In our analysis, embeddings were projected onto a single PLS component aligned with physiological resilience, operationalized as continuous HR-AUC in Section~\ref{sec:resilience_def}. We assessed the association between this PLS-derived dimension and HR-AUC using Spearman rank correlation, with significance evaluated by permutation testing using 5{,}000 permutations~\cite{dobriban2020permutation}. Technical details of PLS are provided in Appendix~\ref{appendix:pls}.

\begin{table}[htbp]
\centering

\begin{tabular}{l c c c}
\toprule
Language Condition & Spearman $r$ & $p$ (Spearman) & $p$ (Permutation) \\
\midrule
Autobiographical Language & 0.843 & $2.9 \times 10^{-17}$ & $< 0.001$ \\
Context-neutral Speech                     & 0.132 & 0.313               & 0.316 \\
\bottomrule
\end{tabular}
\caption{PLS-based representational alignment between \emph{Autobiographical Language} and HR recovery.
Spearman rank correlations and permutation-based $p$-values are reported for associations between a one-dimensional PLS latent representation derived from Autobiographical Language embeddings and HR area-under-the-curve (AUC) following stress cessation.}
 \vspace{-8mm}
\label{tab:rq2_alignment}
\end{table}

As shown in Table~\ref{tab:rq2_alignment}, autobiographical language exhibits a strong and statistically reliable association with HR-AUC ($r=0.843$, $p<0.001$), whereas context-neutral speech shows weak and non-significant alignment ($r=0.132$, $p=0.316$). This contrast suggests that resilience-related recovery information is selectively reflected in autobiographical recall rather than in general speech patterns alone. 
\emph{These findings support autobiographical recall as a lower-burden, reusable proxy for resilience-related subject context. }

\paragraph{\textbf{Overall,}} the findings in Section~\ref{Emperical-motivation} directly motivate our modeling approach and establish HR-AUC and autobiographical recall as resilience-related subject-level proxies. %Building on this empirical grounding, \textbf{\emph{RETRACE}} treats stress physiology as a transferable autonomic backbone and incorporates resilience-related subject context, derived from either HR-AUC or autobiographical recall transcripts, to disambiguate craving-related inter-individual deviations from stress-related physiological activity.

\section{\textbf{\emph{RETRACE}}—a \textbf{re}silience-conditioned, \textbf{s}tress-\textbf{p}retrained \textbf{i}nference framework}
\label{sec:approach}

\subsection{Problem Formulation}

We present a resilience-guided framework for subject-independent craving detection from wearable physiological signals. 
Given a physiological window $\mathbf{x} \in \mathbb{R}^d$, the goal is to predict a binary craving label $y \in \{0,1\}$ derived from self-reported ratings. 
In addition to window-level observations, each subject is associated with a fixed embedding $\mathbf{e}$ that captures stable individual differences through a resilience proxy.

Our analysis in Section~\ref{sec:rq1} suggests that wearable physiology contains measurable subject-independent information related to craving, supporting the feasibility of physiological craving detection. At the same time, craving-related subject-independent signals are weaker and more selective than stress responses, and they are often embedded within broader stress-related autonomic activity. This indicates that craving detection is feasible, but achieving more effective and reliable inference requires models that can interpret subtle, subject-dependent physiological evidence rather than rely on a single population-level mapping.

We therefore formulate craving detection as a subject-conditioned inference problem. The physiological window $\mathbf{x}$ provides momentary autonomic evidence, while the subject-level embedding $\mathbf{e}$ provides a stable context for interpreting that evidence. Under this formulation, similar physiological patterns can contribute differently to craving prediction across individuals. This enables subject-independent inference that accounts for inter-individual variability without requiring per-user craving labels or user-specific model retraining.

%A key challenge in this setting arises from two factors. 
%First, similar physiological patterns may carry different psychological meanings across individuals, leading to substantial inter-subject variability. 
%Second, analysis in Section~\ref{sec:rq1} shows that while stress exhibits strong and consistent physiological signatures, craving-related effects are substantially weaker, sparser, and less consistent across individuals. 
%Importantly, although some features overlap with those observed under stress, craving-related signals also exhibit distinct patterns.

%As a result, the mapping from physiological signals to self-reported craving is inherently ambiguous and subject-dependent, posing a fundamental challenge for subject-independent inference.

\subsection{Architecture Overview}\label{Model-archi}

\begin{figure}[t]
    \centering
    \includegraphics[width=0.85\linewidth]{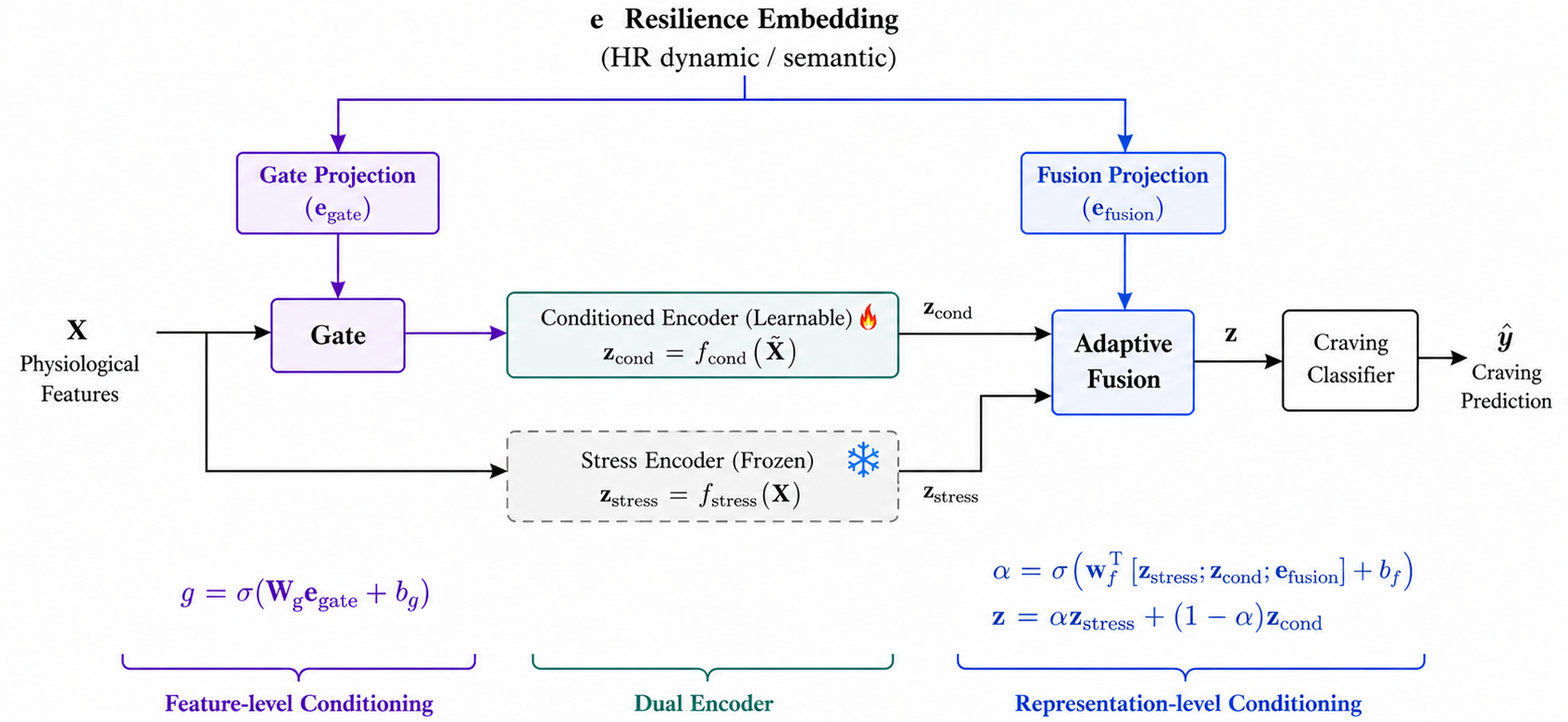}
    \caption{
    Overview of the proposed resilience-guided architecture. 
    A frozen stress encoder captures stable physiological structure, while a subject-conditioned encoder models adaptive variations. 
    A resilience embedding modulates the input conditioning and representation fusion. 
    }
    \vspace{-3mm} 
    \label{fig:architecture}
\end{figure}

\textbf{\emph{RETRACE}} is designed to separate two complementary components of wearable craving inference: (1) generalizable autonomic structure and (2) craving-related patterns that vary from person to person. As illustrated in Figure~\ref{fig:architecture}, the model contains two encoder pathways.
A frozen \textbf{stress-pretrained encoder} provides a stable physiological reference, while a \textbf{conditioned craving encoder} learns adaptive craving-related representations.

%First, a frozen \textbf{stress-pretrained encoder} captures generalizable autonomic structure. Because stress responses are stronger and more consistent than craving responses, this encoder provides a stable physiological reference. Importantly, it is not used to treat stress as craving; rather, motivated by Section~\ref{sec:rq1}, it anchors the model in physiologically meaningful autonomic patterns against which weaker craving-related deviations can be interpreted. Second, a \textbf{conditioned craving encoder} learns adaptive craving-related representations. This encoder operates on resilience-conditioned physiological inputs, is trained with the craving objective, and is designed to learn information distinct from the stress-pretrained representation. It therefore focuses on subtle and heterogeneous craving-related patterns that may not be fully explained by general stress-related physiology.

The \textbf{\emph{resilience embedding}} $\mathbf{e}$ modulates the inference process at two levels. At the input level, it guides which physiological features should be emphasized for a given individual. At the representation level, it controls how much the final prediction should rely on the stress-pretrained representation versus the conditioned craving representation. Together, these components allow \textbf{\emph{RETRACE}} to perform low-capacity personalization through $\mathbf{e}$ while preserving a shared model structure.

%\textbf{\emph{RETRACE}} separates shared physiological structure from adaptive, subject-specific interpretation. As illustrated in Figure~\ref{fig:architecture}, the model consists of two complementary pathways. A frozen stress encoder captures stable physiological patterns that are consistent across individuals, while a learnable encoder models adaptive variations associated with craving-related responses.

%The resilience embedding $\mathbf{e}$ acts as a control signal that modulates both pathways. It influences how physiological inputs are processed and how representations from the two pathways are combined, enabling subject-specific interpretation without altering the underlying physiological feature space.

%This decomposition allows the model to combine a stable physiological reference with subject-specific adaptation: the stress pathway captures shared physiological structure across individuals, while the conditioned pathway models individual-specific variations that are not consistently expressed across subjects.

\subsection{Dual-Encoder Design}\label{encoder-approach}

The dual-encoder design is motivated by the empirical findings in Section~\ref{sec:rq1}: stress produces robust physiological modulation, whereas craving-related signals are weaker, more selective, and largely embedded within stress-related autonomic activity. A single encoder would need to learn both stable autonomic structure and subtle craving-related variation at the same time, which can entangle stress and craving representations and reduce subject-independent generalization. \textbf{\emph{RETRACE}} instead separates these roles into two pathways.

\paragraph{Stress-pretrained encoder.}
The stress encoder $f_{\text{stress}}$ is pretrained on a stress prediction task using cross-entropy loss and then kept fixed during craving training. Given an input physiological window $\mathbf{x}$, it produces a stress-pretrained representation:
$\mathbf{z}_{\text{stress}} = f_{\text{stress}}(\mathbf{x})$. The encoder is trained using data that excludes the target subject used for evaluation, ensuring subject-independent pretraining. Because this encoder is trained using stronger and more reliable stress labels, it captures generalizable autonomic structure across individuals. Importantly, this pathway is not used to treat stress as craving. Rather, it provides a stable physiological reference that anchors the model in meaningful autonomic patterns, allowing weaker craving-related deviations to be interpreted relative to this reference.

\paragraph{Conditioned craving encoder.}
The conditioned encoder $f_{\text{cond}}$ learns craving-related variation from resilience-conditioned inputs. Given a gated input $\tilde{\mathbf{x}}$, it produces: $\mathbf{z}_{\text{cond}} = f_{\text{cond}}(\tilde{\mathbf{x}})$.
This pathway is trained with the craving objective and, through the `orthogonality term objective’ in Section \ref {loss-objectives}, is designed to learn information distinct from $\mathbf{z}_{\text{stress}}$; thus, it focuses on physiological patterns that are subtle, heterogeneous, and potentially subject-dependent. Because input $\tilde{\mathbf{x}}$ is modulated by the resilience-related embedding, the encoder can emphasize different physiological evidence for different individuals. This allows the model to learn craving-related deviations that may not be fully captured by the stress-pretrained representation. In this way, craving is modeled as a separate resilience-conditioned inference task rather than as a direct extension of stress detection.

Together, the two encoders provide a structured representation space: the stress encoder anchors the model in generalizable autonomic physiology, while the conditioned encoder captures adaptive craving-related deviations.

\subsection{Resilience-Guided Conditioning}

%To address the subject-dependent and ambiguous mapping between physiological signals and self-reported craving, we incorporate the resilience embedding $\mathbf{e}$ as a conditioning signal that modulates the inference process. Rather than treating $\mathbf{e}$ as an additional predictive feature, we use it to control how physiological signals are interpreted for craving detection.
To account for subject-dependent differences in how physiological signals relate to craving, \textbf{\emph{RETRACE}} uses the resilience embedding $\mathbf{e}$ as a conditioning signal. Rather than treating $\mathbf{e}$ as an additional predictive feature, the \textbf{\emph{RETRACE}} uses it to guide how physiological autonomic structure is interpreted and integrated.

As shown in Figure~\ref{fig:architecture}, the resilience embedding is projected into two conditioning signals: $\mathbf{e}_{\text{gate}}$ for feature-level conditioning and $\mathbf{e}_{\text{fusion}}$ for representation-level conditioning. These two pathways serve complementary roles. Feature-level conditioning helps the model determine which physiological features are more informative for a given individual.
Representation-level conditioning controls how much the final prediction relies on the generalizable stress-pretrained representation versus the 
conditioned and adaptive craving-related representation.

%As illustrated in Figure~\ref{fig:architecture}, the resilience embedding $\mathbf{e}$ is used to derive two distinct projections that support feature-level and representation-level conditioning. These two pathways correspond to complementary roles in the inference process. First, the model determines which physiological features are more informative for a given individual. Second, it governs how different physiological representations derived through subject -dependant and -independant encoding should contribute to the final decision.

\subsubsection{Feature-Level Conditioning}

At the feature level, we introduce a gating mechanism that adaptively modulates the input representation based on the feature-level resilience embedding projection $\mathbf{e_{gate}}$. 
Specifically, we compute a gating vector: $\mathbf{g} = \sigma\big(W_g \mathbf{e_{gate}} + \mathbf{b}_g\big)$, and apply it to the input as $\tilde{\mathbf{x}} = \mathbf{x} \odot \mathbf{g}$,

where $\sigma(\cdot)$ denotes the sigmoid function and $\odot$ represents element-wise multiplication.
This mechanism assigns subject-specific importance weights to physiological features, enabling selective amplification or suppression of input dimensions. 
As a result, the same physiological signal can be interpreted differently across individuals, addressing variability in how signals relate to craving.

\subsubsection{Representation-Level Conditioning}

At the representation level, we adaptively combine different representations based on both physiological signal representations (Section \ref{encoder-approach}) and the fusion projection of resilience embedding $\mathbf{e_{fusion}}$. 
Given representations $\mathbf{z}_{\text{stress}}$ and $\mathbf{z}_{\text{cond}}$, we compute a fusion weight:

$\alpha = \sigma\big(\mathbf{w}_f^\top [\mathbf{z}_{\text{stress}}, \mathbf{z}_{\text{cond}}, \mathbf{e_{fusion}}] + b_f\big)$, and obtain the final representation as: $\mathbf{z} = \alpha \mathbf{z}_{\text{stress}} + (1 - \alpha)\mathbf{z}_{\text{cond}}$.

This formulation allows the model to dynamically balance between a stable physiological reference of a broader activation and an adaptive representation. 
By conditioning the fusion process on $\mathbf{e}$, the model adjusts its decision behavior across individuals, accounting for individual differences in how physiological signals map to craving.

Together, these two forms of conditioning enable a resilience-guided inference process that adapts both feature utilization and decision behavior, without modifying the underlying physiological representation space.

\subsection{Resilient Embedding Extraction}\label{resilience-embedding-extract}
In this work, $\mathbf{e}$ can be instantiated in multiple ways. We consider two alternatives: (1) \textbf{\emph{HR-dynamics guidance}}, a physiological embedding derived from heart-rate (HR) dynamics, and (2) \textbf{\emph{semantic guidance}}, a representation obtained from autobiographical memory recall.

The \textbf{\emph{HR-dynamics guidance}} is constructed from short (15s) heart-rate segments following calm and stress conditions. For each subject, heart-rate signals are normalized relative to a baseline period (Calm video viewing), and a single post-calm and post-stress segment is extracted after the corresponding task conditions (Section~\ref{sec:resilience_def}). These segments capture complementary baseline-like and stress-response recovery dynamics, enabling characterization of individual differences in recovery. Each segment is summarized using statistical and temporal descriptors (e.g., mean, standard deviation, signed and absolute AUC, slope). The resulting features from the post-calm and post-stress segments are concatenated into a fixed-length embedding (32-dimensional).

The \textbf{\emph{semantic guidance}} is derived from autobiographical memory recall. Participants describe recent positive and negative experiences, which are transcribed and encoded using a pretrained language model into high-dimensional embeddings. The final representation is formed by concatenating the two narratives, yielding a fixed subject-level embedding (3072-dimensional).

%In this work, $\mathbf{e}$ can be instantiated in multiple ways. We consider two alternatives: (1) a physiological state embedding derived from post-task heart-rate (HR) dynamics, and (2) a semantic embedding obtained from autobiographical memory recall. The physiological resilience embedding is derived from post-task heart-rate dynamics, capturing how individuals respond to and recover from stress. For each subject, short heart-rate segments following calm and stress conditions are extracted (Section \ref{sec:resilience_def}) and normalized relative to a baseline period (Section \ref{non-stress-tasks}). Each segment is summarized using statistical and temporal descriptors that characterize both the magnitude and dynamics of heart-rate changes, including measures of central tendency, variability, extrema, and recovery patterns. Features from post-calm and post-stress segments are concatenated to form a fixed-length embedding.

%The semantic resilience embedding is obtained from autobiographical memory recall. Participants provide descriptions of recent positive and negative experiences, which are transcribed and encoded into a fixed-length representation using a large language model. This embedding captures higher-level differences in how individuals interpret and describe personal experiences.

Details on how these two guidance representations are extracted are provided in Appendix~\ref{app:guidance}.

\subsection{Craving Prediction}

The final representation $\mathbf{z}$ is passed to a prediction head to estimate the probability of craving: $\hat{y} = h_{\text{craving}}(\mathbf{z})$.

The model is trained using the binary craving labels derived from self-reported ratings. At inference time, \textbf{\emph{RETRACE}} requires only the physiological window $\mathbf{x}$ and a one-time collected subject-level resilience embedding $\mathbf{e}$. Thus, it supports subject-aware craving inference without requiring craving labels from the target user or user-specific model retraining.

\subsection{Training Objective}
\label{loss-objectives}

\textbf{\emph{RETRACE}} is trained to predict craving using empirical risk minimization with additional regularization terms:
\[
\mathcal{L} = \mathcal{L}_{\text{craving}} 
+ \lambda_{\text{sparse}} \|\mathbf{g}\|_1
+ \lambda_{\text{orth}} \|\mathbf{z}_{\text{stress}}^\top \mathbf{z}_{\text{cond}}\|_2
+ \lambda_{\text{stress}} \, \mathrm{CE}\big(h_{\text{stress}}(\mathbf{z}_{\text{cond}}), y_{\text{stress}}\big).
\]

The primary term, $\mathcal{L}_{\text{craving}}$, is the binary classification loss for predicting self-reported craving labels. The remaining terms serve as regularization to stabilize representation learning and support the proposed architecture.

The sparsity term $\|\mathbf{g}\|_1$ encourages the feature-level conditioning gate to emphasize only a subset of physiological features for each subject, rather than assigning high weights to all input dimensions. This supports selective, subject-aware feature use.

The orthogonality term $\|\mathbf{z}_{\text{stress}}^\top \mathbf{z}_{\text{cond}}\|_2$ encourages the stress-pretrained and conditioned craving representations to capture complementary information. This reduces redundancy between the two encoders and helps separate general autonomic structure from adaptive craving-related variation.

Finally, the auxiliary stress loss applies a lightweight stress prediction head $h_{\text{stress}}(\cdot)$ to the conditioned representation $\mathbf{z}_{\text{cond}}$. This term encourages the conditioned encoder to remain grounded in physiologically meaningful signals while learning craving-related patterns, reducing the risk of relying on spurious features.

The weights $\lambda_{\text{sparse}}$, $\lambda_{\text{orth}}$, and $\lambda_{\text{stress}}$ control the strength of each regularization term. All auxiliary components are used only during training; at inference time, \textbf{\emph{RETRACE}} requires only the physiological window $\mathbf{x}$ and the subject-level resilience embedding $\mathbf{e}$ for craving prediction.

Full architectural specifications for all components, layer configurations, and training hyperparameters for \textbf{\emph{RETRACE}} are provided in Appendix~\ref{appendix:reproduce}.

\section{Evaluation}\label{Eval}

We evaluate the proposed \textbf{\emph{RETRACE}} model on opioid craving detection, comparing against classical machine learning approaches, domain generalization (DG) and multimodal learning approaches as discussed in Section \ref{sec:related_work}.
We further conduct ablation and interpretability analyses to examine the role of subject-level conditioning.

\subsection{Evaluation Protocol and Metrics}

\label{sec:results}
\paragraph{Evaluation Cohort.}
The OUD dataset contains 24 participants (28 sessions). We exclude participants with no within-subject craving variability and those with repeated sessions from evaluation, while retaining them for training. From the remaining eligible participants, 10 OUD subjects are randomly selected as the evaluation cohort.

\paragraph{Protocol and Data Statistics.}
We adopt a leave-one-subject-out (LOSO) protocol with 10 folds \cite{chouhan2026supervised}. In each fold, one subject is held out for testing and the rest are used for training. For stress encoder pretraining, the stress encoder is trained separately within each fold using all control participants and the OUD training subjects in that fold, excluding the held-out test subject. For craving detection, only OUD participants are used. Across all OUD sessions used for craving modeling, the dataset contains 1,421 windows (785 non-craving and 636 craving). Under the LOSO protocol, the test folds collectively contain 494 windows (258 non-craving and 236 craving).
%\paragraph{Evaluation Cohort.}
%We exclude subjects with no within-subject craving variability and participants with multiple sessions from evaluation, while retaining them for training. From the remaining subjects, ten OUD participants are randomly selected as the evaluation cohort.

%\paragraph{Protocol.}
%We adopt a leave-one-subject-out (LOSO) protocol with 10 folds \cite{chouhan2026supervised}. In each fold, one subject is held out for testing and the rest are used for training. For stress detection pre training, models are trained on both control and OUD participants; for craving detection, only OUD participants are used. All models are evaluated under the same experimental protocol. Full architectural specifications for all components, layer configurations, and training hyperparameters for \textbf{\emph{RETRACE}} are provided in Appendix~\ref{appendix:reproduce}.

\paragraph{Metrics.}
We report micro-aggregated accuracy (Acc) and Macro-F1, computed on pooled predictions across test subjects, along with subject-wise mean $\pm$ standard deviation. 
Statistical significance is assessed using paired tests across subjects on accuracy (Acc), with $p$-values and confidence intervals (CI) reported relative to the proposed model with HR dynamics guidance.

\begin{table*}[t]
\centering
\small
\setlength{\tabcolsep}{4pt}
\begin{tabular}{llccccccc}
\toprule
\textbf{Model} & \textbf{Guidance} & \textbf{Acc}  & \textbf{Macro-F1}  & \textbf{Mean $\pm$ Std} & $\boldsymbol{\Delta}$Acc& \textbf{p-value} & \textbf{CI} \\
\midrule
MLP               & –        & 0.581 & 0.578 & $0.579 \pm 0.107$ & 0.139 & 0.010 & [0.063, 0.215] \\
RBF-SVM           & –        & 0.630 & 0.624 & $0.623 \pm 0.205$ & 0.094 & 0.105 & [0.009, 0.195] \\
XGBoost           & –        & 0.638 & 0.632 & $0.629 \pm 0.152$ & 0.088 & 0.160 & [0.003, 0.174] \\
IRM               & –        & 0.619 & 0.616 & $0.619 \pm 0.096$ & 0.099 & 0.051 & [0.015, 0.188] \\
GroupDRO          & –        & 0.609 & 0.607 & $0.636 \pm 0.114$ & 0.108  & 0.050    & [-0.0226, 0.1903] \\
V-REx             & –        & 0.623 & 0.622 & $0.609 \pm 0.115$ & 0.094 & 0.066 & [0.015, 0.173] \\
Cross-Attn        & HR dynamics   & 0.650 & 0.647 & $0.646 \pm 0.188$ & 0.071 & 0.193 & [-0.038, 0.189] \\
Cross-Attn        & Semantic & 0.607 & 0.607 & $0.610 \pm 0.180$ & 0.108 & 0.084 & [0.006, 0.238] \\
MoE               & HR dynamics   & 0.626 & 0.624 & $0.626 \pm 0.095$ & 0.092 & 0.037 & [0.014, 0.167] \\
MoE               & Semantic & 0.621 & 0.613 & $0.621 \pm 0.094$ & 0.096 & 0.037 & [0.024, 0.163] \\

FiLM       & HR dynamics   & 0.630 & 0.625 & $0.631 \pm 0.127$ & 0.086 & 0.155 & [-0.001, 0.181] \\
FiLM     & Semantic & 0.640 & 0.636 & $0.641 \pm 0.157$ & 0.077 & 0.375 & [-0.033, 0.192] \\
\textbf{\emph{RETRACE}} &\textbf{HR dynamics} & \textbf{0.721} & \textbf{0.720} & $\mathbf{0.718 \pm 0.112}$ & –& –& –\\
\textbf{\emph{RETRACE}} &\textbf{Semantic}  & \textbf{0.704} & \textbf{0.694} & $\mathbf{0.699 \pm 0.238}$ & –& –& –\\

\bottomrule
\end{tabular}
\caption{
Baseline comparison on craving detection. 
Metrics are computed on pooled predictions across subjects.  Mean $\pm$ std is computed across subject-wise folds.  $\boldsymbol{\Delta}$Acc denotes the accuracy improvement relative to the proposed model with HR dynamics guidance (\textbf{\emph{RETRACE}} $-$ Baseline). Confidence intervals (CI) are reported for accuracy.
}
\vspace{-8mm} 
\label{tab:baseline_results}
\end{table*}

\subsection{Comparison with Baselines}
\label{sec:eval_baseline}
We evaluate \textbf{\emph{RETRACE}} with both resilience-related guidance representations introduced earlier: a physiology-grounded HR-AUC/HR-dynamics proxy from post-stress recovery and a semantic proxy from autobiographical memory recall. This comparison tests whether \textbf{\emph{RETRACE}}'s subject-level conditioning remains effective across both physiological and autobiographical forms of resilience-related guidance.
Table~\ref{tab:baseline_results} summarizes results across representative baselines for physiological sensing, as outlined in Section~\ref{sec:related_work}. These include classical machine learning models (MLP, SVM, XGBoost)\cite{carreiro2020wearable,carreiro2024evaluation}, domain generalization methods (IRM\cite{arjovsky2019invariant}, GroupDRO\cite{sagawa2019distributionally}, V-REx \cite{krueger2021out}), and multimodal conditioning architectures (FiLM \cite{perez2018film}, Cross-Attention\cite{vaswani2017attention}, MoE\cite{shazeer2017outrageously}).
\paragraph{Overall performance.}
\textbf{\emph{RETRACE}} achieves the best performance across all metrics. 
In particular, the HR dynamics variant reaches an accuracy of 0.721 and Macro-F1 of 0.720, outperforming the strongest baseline (Cross-Attn with HR dynamics, 0.650 accuracy) by over 7\%. 
The semantic variant also consistently exceeds all baselines (0.704 accuracy), demonstrating robustness across different forms of subject-level guidance. 
Notably, these improvements are achieved under a strictly subject-disjoint (LOSO) setting, highlighting the model’s ability to generalize to unseen individuals.
\paragraph{Comparison with classical and domain generalization methods.}
Compared to classical models and domain generalization (DG) approaches, the proposed method consistently achieves higher performance. 
The best classical baseline (XGBoost) reaches 0.638 accuracy, while DG methods such as IRM and V-REx achieve 0.619 and 0.623, respectively, all substantially below our model (0.721). 
These methods rely on learning a shared mapping from physiological signals to labels across subjects, which breaks down under inter-subject heterogeneity. 
In contrast, our model explicitly conditions the interpretation of physiological signals on subject-level information, resulting in improved generalization.
\paragraph{Comparison with multimodal conditioning approaches.}
We further compare against multimodal baselines, including FiLM, Cross-Attention, and Mixture-of-Experts (MoE). 
While these methods incorporate subject-level information, their improvements remain limited (e.g., Cross-Attn with HR dynamics achieves 0.650 accuracy and FiLM reaches 0.640), still significantly below our model (0.721). 
This indicates that incorporating subject-level information alone is insufficient; how such information modulates physiological interpretation is critical. 
Our structured conditioning approach, combining feature-level gating and representation-level fusion, leads to consistent performance gains over these architectures.
\paragraph{Impact of guidance representation.}
The choice of subject-level guidance consistently affects performance across models. 
HR dynamics guidance outperforms semantic embeddings not only in the proposed model (0.721 vs.\ 0.704), but also across multimodal baselines (e.g., Cross-Attn: 0.650 vs.\ 0.607; MoE: 0.626 vs.\ 0.621). 
This suggests that representations aligned with physiological response patterns provide more effective conditioning signals. 
At the same time, the proposed model maintains strong performance under both guidance types, indicating that improvements are not solely driven by embedding choice, but by the conditioning mechanism itself.

\paragraph{Statistical significance.}
The proposed model consistently outperforms all baselines in terms of point estimates ($\Delta$Acc $> 0$). 
While only a subset of comparisons reach statistical significance at the 0.05 level (e.g., MLP: $p=0.010$), many baselines show marginal significance (e.g., IRM: $p=0.051$, V-REx: $p=0.066$), indicating a consistent performance trend across subjects despite variability.

\subsection{Ablation study}
\label{sec:eval_ablation}

\begin{table*}[t]
\centering
\small
\setlength{\tabcolsep}{4pt}
\begin{tabular}{lccc|ccc}
\toprule
\multirow{2}{*}{\textbf{Variant}} 
& \multicolumn{3}{c}{\textbf{HR dynamics Guidance}} 
& \multicolumn{3}{c}{\textbf{Semantic Guidance}} \\
\cmidrule(lr){2-4} \cmidrule(lr){5-7}
& Acc & F1 & Acc Mean $\pm$ Std 
& Acc & F1 & Acc Mean $\pm$ Std \\
\midrule

\multicolumn{7}{l}{\textbf{(1) Component Ablations}} \\
%All zeros guidance   & 0.607 & 0.605 & $0.604 \pm 0.123$ & 0.640 & 0.638 & $0.639 \pm 0.108$ \\
%Learnable guidance embedding & 0.621 & 0.620 & $0.616 \pm 0.146$ & 0.611 & 0.609 & $0.609 \pm 0.112$ \\
Remove fusion gate   & 0.619 & 0.617 & $0.616 \pm 0.129$ & 0.636 & 0.630 & $0.636 \pm 0.157$ \\
Remove input gate    & 0.658 & 0.658 & $0.652 \pm 0.147$ & 0.654 & 0.649 & $0.648 \pm 0.273$ \\
Remove frozen stress encoder & 0.660 & 0.656 & $0.659 \pm 0.170$ & 0.623 & 0.620 & $0.628 \pm 0.141$ \\
Remove learnable encoder & 0.674 & 0.663 & $0.670 \pm 0.217$ & 0.654 & 0.645 & $0.663 \pm 0.211$ \\

\midrule
\multicolumn{7}{l}{\textbf{(2) Loss Ablations}} \\
Remove sparsity loss & 0.664 & 0.663 & $0.662 \pm 0.108$ & 0.666 & 0.667 & $0.676 \pm 0.165$ \\
Remove orthogonality loss & 0.658 & 0.657 & $0.657 \pm 0.155$ & 0.654 & 0.624 & $0.654 \pm 0.179$ \\
Remove stress loss   & 0.650 & 0.647 & $0.642 \pm 0.185$ & 0.666 & 0.660 & $0.666 \pm 0.131$ \\
Remove stress loss and orthogonality loss 
                     & 0.692 & 0.691 & $0.686 \pm 0.149$ & 0.696 & 0.696 & $\mathbf{0.692 \pm 0.152}$ \\

\midrule
\multicolumn{7}{l}{\textbf{(3) Embedding Variants}} \\
Only post stress / bad memory recall & 0.702 & 0.702 & $0.699 \pm 0.140$ & 0.646 & 0.646 & $0.648 \pm 0.133$ \\
Only post calm / good memory recall & 0.658 & 0.657 & $0.651 \pm 0.150$ & 0.674 & 0.660 & $0.675 \pm 0.146$ \\
\textbf{\emph{RETRACE}}: HR-dynamics / semantic & \textbf{0.721} & \textbf{0.720} & $\mathbf{0.718 \pm 0.112}$ & \textbf{0.704} & \textbf{0.694} & $\mathbf{0.699 \pm 0.238}$ \\

\bottomrule
\end{tabular}
\caption{Ablation study results under HR dynamics and semantic guidance.}
\vspace{-8mm} 
\label{tab:ablation}
\end{table*}

We conduct ablation experiments to analyze the contributions of architectural components, auxiliary losses, and guidance representations. Table~\ref{tab:ablation} summarizes results under both HR dynamics and semantic guidance.

\textbf{Component Ablations.}
Removing individual components leads to consistent performance degradation, with the largest drop observed when the \emph{adaptive fusion gate} is removed (e.g., from 0.721 to 0.619 under HR dynamics), indicating that it serves as the primary driver of performance. 
Other components, such as the \emph{input gate} and \emph{encoders}, also contribute (e.g., removing the input gate reduces accuracy from 0.721 to 0.658), but their impact is comparatively smaller. 
Overall, these results suggest that while architectural components are important, their effectiveness depends on the presence of meaningful guidance.

\textbf{Loss Ablations.}
We further examine the effect of auxiliary losses. 
Removing the \emph{auxiliary stress} loss decreases performance from 0.721 to 0.650, while removing the \emph{orthogonality loss} leads to a drop to 0.658, indicating that both contribute to stable learning. 
Interestingly, removing both losses yields better performance than retaining only one (0.692), suggesting that these objectives interact and must be jointly optimized to effectively regularize the representation. 
Overall, the full model achieves the best performance, indicating that the combination of these losses improves representation quality and stability.

\textbf{Embedding Variants.}
We analyze how different forms of resilience representation discussed in Section \ref{resilience-embedding-extract} contribute to performance. Under semantic guidance, using only one autobiographical recall component—positive or negative memory recall—consistently reduces performance. This suggests that the semantic guidance benefits from information distributed across both types of autobiographical narratives.

HR-dynamics guidance shows a different pattern. Using only the post-stress embedding results in a smaller drop in accuracy, from 0.721 to 0.702, whereas using only the calm-related embedding leads to a larger decrease, from 0.721 to 0.658. This is expected because the HR-dynamics proxy is intended to capture post-stress recovery, while the calm segment mainly provides a baseline reference for interpreting recovery. Together, these results suggest that HR-dynamics guidance is primarily driven by post-stress recovery dynamics, whereas semantic guidance captures a more distributed representation of subject-level context.

\textbf{Summary.}
Overall, these results highlight that guidance is the primary factor driving performance improvements, while architectural components and auxiliary losses play supporting roles in stabilizing and refining representation learning. 
The full model achieves the best performance by combining subject-level guidance with structured conditioning and complementary regularization.

\subsection{Model Behavior Across Task Conditions}
The goal of this analysis is to examine whether \textbf{\emph{RETRACE}} detects subjective craving rather than simply using task-induced stress as a shortcut. Because stress-inducing tasks are associated with higher craving prevalence, a model could appear effective by primarily detecting stress. We therefore evaluate performance separately across task conditions and compare aggregate performance between stress and non-stress segments.

As shown in Table~\ref{tab:task_performance_transposed}, performance remains relatively consistent across tasks and does not systematically favor stress-inducing conditions. Aggregated results also show comparable accuracy across stress and non-stress segments: 0.678 vs.\ 0.743 under HR-dynamics guidance, and 0.725 vs.\ 0.693 under semantic guidance. If the model relied mainly on task-induced stress, we would expect substantially higher accuracy during stress tasks and degraded performance during non-stress tasks. This pattern is not observed, suggesting that the model does not simply exploit task or stress-related cues.

This result is consistent with the dataset characteristics in Section~\ref{data-charact}, where craving does not deterministically follow task structure. Although stress-inducing tasks show higher craving prevalence, craving also occurs during non-stress conditions and varies substantially across individuals. Together, these findings support the interpretation that \textbf{\emph{RETRACE}} captures physiological patterns associated with participant-reported craving rather than merely detecting task-induced stress.

%We first examine model performance across different task conditions, as shown in Table~\ref{tab:task_performance_transposed}. Performance remains relatively consistent across tasks, without systematically favoring stress-inducing conditions. Aggregated results further show comparable accuracy across stress and non-stress segments (HR dynamics: 0.678 vs.\ 0.743; semantic: 0.725 vs.\ 0.693). 

%If the model relied on task-induced stress as a shortcut, we would expect substantially higher accuracy under stress and degraded performance otherwise; however, such a pattern is not observed. This suggests that the model does not simply exploit task or stress-related cues.

%This observation is consistent with the dataset characteristics described in Section~\ref{data-charact}. As shown there, craving does not deterministically follow task structure: although stress-inducing tasks are associated with higher craving prevalence, craving also occurs during non-stress conditions and varies substantially across individuals. This confirms that task structure alone cannot reliably determine craving, supporting the interpretation that the model captures patterns associated with subjective craving rather than task-induced stress.
\begin{table}[t]
\centering

\resizebox{0.8\linewidth}{!}{
\begin{tabular}{lccccccc}
\toprule
\textbf{Model} & \textbf{Neutral Video} & \textbf{Arithmetic} & \textbf{QA Baseline} & \textbf{Counting} & \textbf{Calm} & \textbf{Song Preparation} & \textbf{Stroop} \\
\midrule
HR dynamics Acc & 0.707 & 0.714 & 0.923 & 0.714 & 0.768 & 0.600 & 0.658 \\
Semantic Acc & 0.724 & 0.684 & 1.000 & 0.633 & 0.659 & 0.800 & 0.763 \\
\bottomrule
\end{tabular}
}
\caption{Task-level craving detection accuracy across models.}
\vspace{-10mm} 
\label{tab:task_performance_transposed}
\end{table}

\subsection{Effect of Guidance Quality and Specificity}
\label{sec:guidance_ablation}

The goal of this analysis is to test whether \textbf{\emph{RETRACE}} benefits from resilience guidance because it provides meaningful subject-level context, rather than simply because it adds extra inputs or model parameters. To do so, we evaluate variants that systematically remove, weaken, or misalign the guidance signal: (1) \textit{all-zero guidance}, which removes resilience-related information entirely; (2) \textit{shared learnable guidance}, where a single global embedding is learned and shared across all samples to control for added parameterization; and (3) \textit{shuffled guidance}, where resilience embeddings are randomly reassigned at test time to break the correspondence between each physiological signal and its true subject-level context.

Results in Table~\ref{tab:guidance} reveal a consistent pattern. Removing guidance leads to a substantial drop in performance (Acc: 0.607), indicating that physiological signals alone are insufficient under subject-independent evaluation. Introducing a shared learnable embedding provides only marginal improvement (Acc: 0.621), suggesting that gains are not driven by additional parameters or global conditioning. In contrast, shuffled guidance yields intermediate performance (HR: 0.662, Semantic: 0.668), performing better than no guidance but worse than correctly aligned guidance. This indicates that the model benefits most when resilience guidance is correctly matched to the subject.
Overall, these results support the role of resilience guidance as a meaningful subject-level conditioning signal. Correctly aligned resilience information helps \textbf{\emph{RETRACE}} interpret physiological patterns in a subject-dependent way, rather than acting as a generic auxiliary input.

\subsection{Sensitivity Evaluation}
\label{sec:eval_sensitivity}
We evaluate the robustness of the proposed framework under sensor and temporal distribution shifts, focusing on whether subject-level guidance can generalize across deployment conditions. Table~\ref{tab:sens} summarizes the results.

\textbf{Cross-wrist Generalization.}
We examine robustness to sensor placement by training on left-wrist data and evaluating on right-wrist signals. 
Performance decreases under HR dynamics guidance (0.721 $\rightarrow$ 0.679), while semantic guidance remains stable (0.704 $\rightarrow$ 0.714). 
This suggests that semantic guidance provides a more invariant participant-level signal that is less sensitive to sensor-specific variations, whereas HR-based features depend more on consistent measurement conditions.

\textbf{Cross-visit Generalization.}
We evaluate whether resilience representations collected at one-time can be reused across sessions by comparing same-visit and cross-visit embeddings. 
Performance remains stable across visits (e.g., 0.700 $\rightarrow$ 0.700 and 0.789 $\rightarrow$ 0.771), with no consistent degradation. 
This indicates that resilience embeddings capture subject-level characteristics that generalize over time, supporting their use as reusable conditioning signals.

\textbf{Summary.}
Overall, the proposed framework demonstrates robustness to both sensor and temporal shifts. 
Semantic guidance shows stronger invariance to deployment-related changes, while HR dynamics guidance achieves slightly higher peak performance. 
Together, these results suggest that resilience-based conditioning provides a stable and reusable signal for real-world craving inference.

\iffalse
\begin{table}[t]
\centering
\small
\setlength{\tabcolsep}{4pt}
\begin{tabular}{lccc}
\toprule
\textbf{Variant} 
& Acc & F1 & Acc Mean $\pm$ Std \\
\midrule

All zeros guidance   & 0.607 & 0.605 & $0.604 \pm 0.123$ \\
Shared Learnable guidance embedding & 0.621 & 0.620 & $0.616 \pm 0.146$ \\

Shuffle (HR dynamics) & 0.662 & 0.659 & $0.662 \pm 0.128$ \\
Shuffle (Semantic)    & 0.668 & 0.668 & $0.672 \pm 0.168$ \\

\bottomrule
\end{tabular}
\caption{Effect of different guidance strategies under HR dynamics and semantic guidance settings.}
\label{tab:guidance}
\end{table}

\begin{table}[t]
\centering
\small
\setlength{\tabcolsep}{4pt}
\begin{tabular}{lcc|cc}
\toprule
\multirow{2}{*}{\textbf{Group}} 
& \multicolumn{2}{c}{\textbf{HR dynamics}} 
& \multicolumn{2}{c}{\textbf{Semantic}} \\
\cmidrule(lr){2-3} \cmidrule(lr){4-5}

& Acc & F1 
& Acc & F1 \\

\midrule
\multicolumn{4}{l}{\textbf{(1) Cross visit}} \\
Visit 1 / Same-visit embedding   & 0.700 & 0.635 &0.660 &0.656  \\
Visit 1 / cross-visit embedding   & 0.700 & 0.670 & 0.681 & 0.679 \\
Visit 2 / Same-visit embedding & 0.789 & 0.787 & 0.771 &.761  \\
Visit 2 / cross-visit embedding & 0.771 & 0.764 &  0.789& 0.769 \\
\midrule
\multicolumn{4}{l}{\textbf{(2) Cross wrist}} \\
Left wrist(Train\&Test)  & 0.721 & 0.720 & 0.704 & 0.694  \\
Left wrist $\rightarrow$ Right wrist    &  0.6786 & 0.6786 & 0.7143  & 0.7107 \\
\bottomrule
\end{tabular}
\caption{Performance comparison across gender groups under HR dynamic and Semantic guidance.}
\label{tab:sens}
\end{table}
\fi

\begin{table}[t]
\centering

\begin{minipage}{0.47\linewidth}
\centering
\small
\setlength{\tabcolsep}{4pt}
\begin{tabular}{lccc}
\toprule
\textbf{Variant} 
& Acc & F1 & Acc Mean $\pm$ Std \\
\midrule
All zeros guidance   & 0.607 & 0.605 & $0.604 \pm 0.123$ \\
Shared learnable embedding & 0.621 & 0.620 & $0.616 \pm 0.146$ \\
Shuffle (HR dyn.) & 0.662 & 0.659 & $0.662 \pm 0.128$ \\
Shuffle (Semantic) & 0.668 & 0.668 & $0.672 \pm 0.168$ \\
\bottomrule
\end{tabular}
\caption{Effect of different guidance strategies.}
\label{tab:guidance}
\end{minipage}
\hfill
\begin{minipage}{0.47\linewidth}
\centering
\scriptsize
\setlength{\tabcolsep}{3pt}
\begin{tabular}{lcc|cc}
\toprule
\multirow{2}{*}{\textbf{Setting}} 
& \multicolumn{2}{c}{\textbf{HR}} 
& \multicolumn{2}{c}{\textbf{Semantic}} \\
\cmidrule(lr){2-3} \cmidrule(lr){4-5}
& Acc & F1 & Acc & F1 \\
\midrule

\multicolumn{5}{l}{\textbf{Cross visit}} \\
\makecell[l]{Visit1 / Same-visit embedding  }  & 0.700 & 0.635 & 0.660 & 0.656 \\
\makecell[l]{Visit1 / cross-visit embedding} & 0.700 & 0.670 & 0.681 & 0.679 \\
\makecell[l]{Visit2 / Same-visit embedding }  & 0.789 & 0.787 & 0.771 & 0.761 \\
\makecell[l]{Visit2 / cross-visit embedding} & 0.771 & 0.764 & 0.789 & 0.769 \\

\midrule
\multicolumn{5}{l}{\textbf{Cross wrist}} \\
\makecell[l]{Left (train \& test)} & 0.721 & 0.720 & 0.704 & 0.694 \\
\makecell[l]{Left $\rightarrow$ Right} & 0.679 & 0.679 & 0.714 & 0.711 \\

\bottomrule
\end{tabular}
\caption{Cross-condition generalization under HR and semantic guidance.}
\vspace{-8mm}  
\label{tab:sens}
\end{minipage}

\end{table}

\subsection{Empirical Justification of \textbf{\emph{RETRACE}}: Model Interpretation and Feature Attribution}
\label{sec:interpretation}
\begin{figure}[t]
    \centering
    \includegraphics[width=0.9\linewidth]{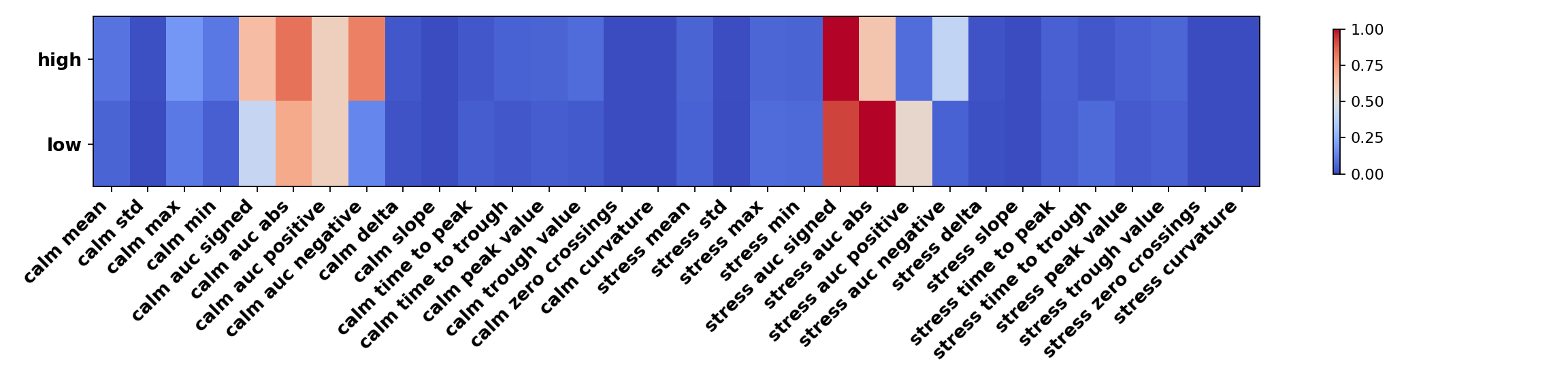}
    \caption{
    Feature attribution under HR dynamics guidance for high- and low-resilience groups. AUC-based features (signed, absolute, positive, and negative) dominate attribution, while other temporal features show low importance.
    }
    \vspace{-3mm} 
    \label{fig:hr_auc_attr}
\end{figure}

To better understand how resilience-guided conditioning shapes physiological inference, we analyze feature attribution and representation weighting during inference. These analyses aim to verify whether the model adapts its interpretation of physiological signals in a subject-specific manner, as intended by the proposed design.
Specifically, we use Integrated Gradients to estimate feature-level attribution and directly extract fusion gate weights from the forward pass to analyze how different representations are weighted during inference.
To examine how this modulation varies across individuals, we aggregate attribution scores and gate weights at the subject level and group participants into high- and low-resilience based on their post-stress HR-AUC recovery, as defined in Appendix~\ref{app:resilience_group}. This grouping allows us to compare how the model behaves under different levels of resilience and to assess whether resilience systematically influences the interpretation of physiological signals.
\paragraph{Feature Attribution under HR dynamics Guidance.}
Figure~\ref{fig:hr_auc_attr} presents feature attribution results under HR dynamics guidance, comparing high- and low-resilience groups. Notably, only a small subset of features exhibits consistently high importance, primarily corresponding to AUC statistics of heart rate changes, including signed AUC, absolute AUC, and their positive and negative components. 
Furthermore, AUC features derived from post-stress segments generally show higher importance than those derived from post-calm segments. This pattern suggests that recovery dynamics following stress play a more prominent role in the model’s inference process. In particular, HR-based AUC features of post-stress recovery emerge as the most influential features, indicating that they capture key aspects of resilience-related variation in physiological responses. This interpretation is consistent with prior work that conceptualizes resilience as a recovery process and operationalizes it using AUC-based measures \cite{den2022conceptualizing}.

\begin{figure}[htbp]
  \centering
  \vspace{-3mm}
  \begin{minipage}[t]{0.45\textwidth}
    \centering
    \includegraphics[width=\textwidth]{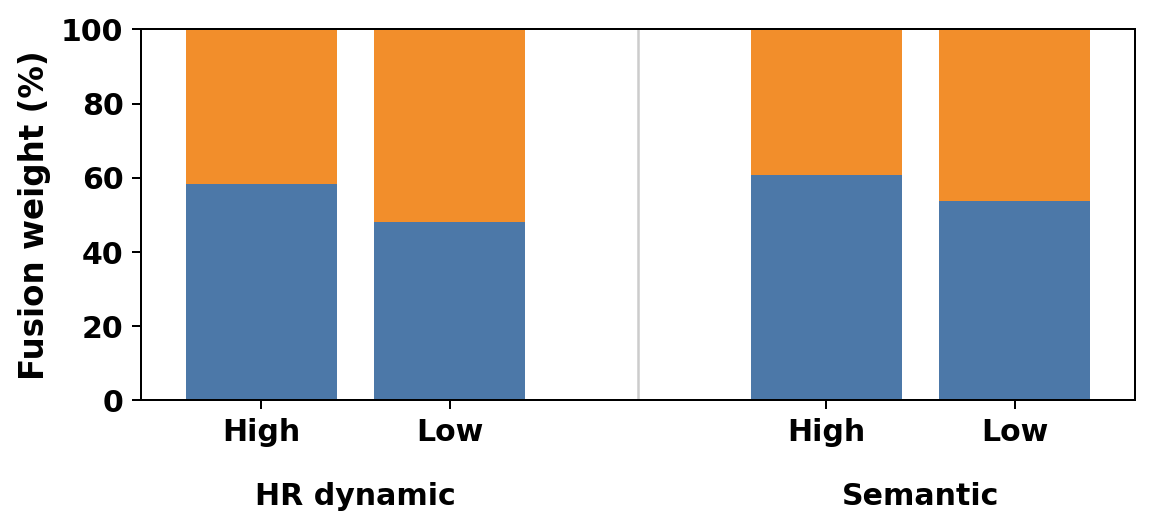}
    \caption{Fusion gate weights under different guidance types. Blue denotes the frozen stress encoder, and orange denotes the learnable encoder.}
     \vspace{-3mm} 
    \label{fig:fusion_gate_combined}
  \end{minipage}
  \hfill
  \begin{minipage}[t]{0.45\textwidth}
    \centering
    \includegraphics[width=\textwidth]{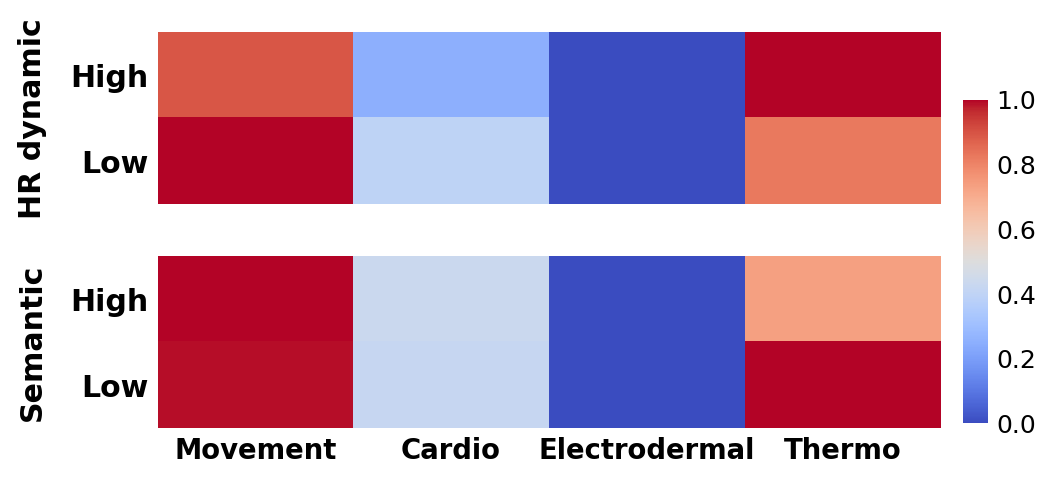}
    \caption{Modality-level attribution under different guidance types.}
     \vspace{-8mm} 
    \label{fig:modality_combined}
  \end{minipage}
\end{figure}

\paragraph{Fusion Gate Behavior.}
Figure~\ref{fig:fusion_gate_combined} shows the fusion gate weights assigned to the learnable physiological encoder and the frozen stress encoder under both HR dynamics and semantic guidance. We observe a consistent pattern across guidance types: low-resilience individuals receive higher weights on the learnable encoder and lower weights on the frozen stress encoder.
This behavior provides direct evidence that the model adapts its representation based on subject-level resilience, as intended by the conditioning mechanism. The frozen encoder captures shared stress-related structure, while the learnable encoder captures craving-specific deviations. For low-resilience individuals, the model relies more heavily on the learnable encoder, indicating greater sensitivity to craving-related physiological patterns. In contrast, for high-resilience individuals, physiological responses are more likely to reflect stress without escalating into craving, resulting in relatively greater reliance on the stress encoder.
Importantly, the similarity of this pattern across both HR dynamics and semantic guidance suggests that different forms of participant-level guidance lead to consistent modulation of physiological interpretation. This indicates that both guidance mechanisms capture aligned representations of subject-specific resilience, despite being derived from distinct sources.

\paragraph{Modality-Level Attribution.}
Figures~\ref{fig:modality_combined} show modality-level attribution under HR dynamics and semantic guidance, respectively.
Across both guidance types, the model assigns higher importance to thermoregulation and movement signals, while electrodermal and cardiovascular features receive comparatively lower weights. This pattern is consistent with our statistical analysis in Section~\ref{sec:rq1}, which shows that thermoregulation features exhibit significant differentiation with respect to craving.

This finding is notable because electrodermal and cardiovascular signals are traditionally associated with stress detection, with electrodermal activity (EDA) widely regarded as a gold-standard biomarker for stress \cite{xiao2024reading}. The reduced attribution of these canonical stress-related modalities suggests that the model is not simply relying on stress-related biomarkers, but instead learns from alternative physiological channels that better capture craving-related dynamics.

\paragraph{Summary.}
Overall, these analyses provide evidence that the model operates in a resilience-conditioned manner. Feature attribution shows that inference is driven by recovery-related dynamics rather than dominant stress responses. Fusion gate behavior demonstrates that subject-level guidance modulates the balance between shared and adaptive representations. Finally, the consistency between HR dynamics and semantic guidance indicates that both representations capture aligned notions of resilience. 

Together, these findings support the central hypothesis that craving detection requires subject-specific interpretation of physiological signals, rather than a fixed mapping across individuals.

\section{In-the-Wild Feasibility Study}\label{sec:inthewild}

To examine whether \textbf{\emph{RETRACE}} can transfer beyond the controlled laboratory setting, we conducted a small in-the-wild feasibility study using real-world craving reports. This analysis is not intended as a definitive validation of real-world performance. Rather, it tests whether a model trained only on laboratory data can produce meaningful craving predictions when applied to naturally occurring daily-living wearable data from a held-out participant.

%To evaluate whether the \textbf{\emph{RETRACE}} framework generalizes beyond controlled laboratory settings, we conducted an in-the-wild evaluation using real-world craving reports. Specifically, we test whether a model trained on lab-collected physiological data can predict craving as it naturally occurs in daily life.
\paragraph{Data Collection.}
We collected in-the-wild data from one participant who also completed the laboratory study. Over a 14-day period, the participant completed ecological momentary assessments (EMA) three times per day, reporting craving levels over the preceding two-hour interval. This resulted in 32 EMA instances (26 non-craving, 6 craving) and 15{,}148 physiological windows. To ensure strict separation between training and evaluation, the model was trained on the Lab dataset, \emph{excluding this participant,} and evaluated only on this participant's in-the-wild data.

\paragraph{Evaluation Protocol.}
\textbf{\emph{RETRACE}} produces predictions at the physiological-window level. To align these predictions with EMA annotations, we aggregated all window-level predictions within the two-hour interval preceding each EMA report. Binary EMA-level predictions were obtained using majority voting across the corresponding physiological windows. This protocol preserves window-level inference while enabling evaluation against participant-reported craving over the EMA recall interval.

\paragraph{Results.} Because the in-the-wild labels are highly imbalanced, with 26 non-craving and 6 craving instances, we report balanced accuracy (BAcc) and AUC rather than standard accuracy alone. As shown in Table~\ref{tab:in_the_wild}, \textbf{\emph{RETRACE}} achieves above-chance performance under both guidance types. Semantic guidance obtains the highest balanced accuracy (0.686), suggesting stronger binary classification performance, while HR-dynamics guidance obtains the highest AUC (0.679), suggesting a more stable ranking of craving likelihood across EMA instances.

Overall, this feasibility study provides initial evidence that resilience-guided craving inference can transfer from laboratory training to naturalistic wearable sensing. Because it includes only one participant and a few craving EMA reports, the findings are preliminary and require validation through larger longitudinal studies across participants, contexts, adherence patterns, and naturally occurring craving episodes.

\section{\textbf{\emph{RETRACE}}'s Deployment Considerations for Practical Use}
\label{sec:deployment}

\textbf{\emph{RETRACE}} is designed for low-burden wearable craving sensing. At inference time, it requires only a short physiological window from a wearable device and a one-time collected subject-level resilience guidance representation. It does not require craving labels from the target user or user-specific model retraining, making it suitable for practical settings where continuous physiology can be collected passively and subject-level guidance can be obtained during onboarding.

\paragraph{Two deployment pathways.}
The two guidance variants offer different trade-offs between efficacy, usability, and deployment burden. The HR-dynamics variant uses post-stress heart-rate recovery as a physiology-grounded resilience proxy \cite{den2022conceptualizing}. It may be most appropriate in supervised settings, such as clinical intake, rehabilitation programs, or structured onboarding, where a brief calm baseline and stress/recovery protocol can be administered. This guidance is closely tied to physiological recovery and may support stable probabilistic ranking of craving likelihood over time, but it requires controlled measurement conditions.

The semantic variant uses autobiographical memory recall as a lower-burden source of subject-level context. A user provides short descriptions of recent positive and negative personal experiences, which are transcribed and converted into a compact embedding for later inference. This approach does not require stress induction or physiological recovery measurement, making it easier to deploy remotely or in unsupervised settings. However, autobiographical narratives are privacy-sensitive \cite{jat2023harnessing}. In practical deployment, raw narratives should not be stored or reused after embedding generation; instead, only compact embeddings should be retained with appropriate encryption, access control, and deletion options. This reduces, but does not fully eliminate, privacy risks because embeddings may still encode sensitive information \cite{song2020information}. Thus, HR-dynamics guidance favors physiological specificity, while semantic guidance favors usability and scalability.

\paragraph{Use in wearable recovery support systems.}
In both deployment paths, \textbf{\emph{RETRACE}} can operate on continuous wearable sensor streams and estimate craving risk without repeatedly prompting users for self-reports. This is important because mHealth systems that rely on sparse ecological momentary assessments may miss brief or acute periods of vulnerability~\cite{King2025}. By enabling continuous, passive, and subject-aware craving inference from wearable physiology, \textbf{\emph{RETRACE}} could support timely and personalized digital support when physiological patterns indicate elevated craving risk~\cite{Perski2021,Garland2023,Hampton2025}.

\textcolor{purple}{}

\begin{table}[t]
\centering
\small
\setlength{\tabcolsep}{5pt}
\begin{tabular}{lcc|cc}
\toprule
\textbf{Model} 
& \multicolumn{2}{c}{\textbf{HR dynamics Guidance}} 
& \multicolumn{2}{c}{\textbf{Semantic Guidance}} \\
\cmidrule(lr){2-3} \cmidrule(lr){4-5}
& \textbf{BAcc} & \textbf{AUC} 
& \textbf{BAcc} & \textbf{AUC} \\
\midrule
Cross-Attn & 0.635 & 0.635 & 0.500 & 0.599 \\
MoE        & 0.635 & 0.635 & 0.583 & 0.583 \\
FiLM        & 0.625 & 0.448 & 0.562 & 0.410 \\
\textbf{RETRACE} 
& \textbf{0.660} & \textbf{0.679} 
& \textbf{0.686} & 0.564 \\
\bottomrule
\end{tabular}
\caption{
In-the-wild evaluation results under different guidance types. Balanced accuracy (BAcc) is reported as the primary metric due to class imbalance, with AUC provided for reference.
}
\vspace{-10 mm}  
\label{tab:in_the_wild}
\end{table}

\section{Inference-time Efficiency and Runtime Analysis}
\label{runtime-val}

To assess whether \textbf{\emph{RETRACE}} can support real-time wearable and mobile health deployment, we benchmarked inference latency across ten heterogeneous computing platforms. Local inference is important for practical mHealth systems because it can reduce dependence on cloud computation, improve availability in low-connectivity settings, and better protect user privacy~\cite{yurur2014context,lane2010survey,nahum2016just}.

%\textcolor{purple}{To assess the feasibility of deploying \textbf{\emph{RETRACE}} for real-time JITAIs \cite{nahum2016just} on wearable or mobile edge devices, we conducted a runtime analysis across \textit{ten} heterogeneous computing platforms. A practical constraint for mHealth systems is that inference should be performed locally whenever possible, circumventing the need for heavy cloud computation \cite{yurur2014context}. Local execution not only preserves user privacy but also ensures system availability in low-connectivity environments \cite{lane2010survey,nahum2016just}.}

We evaluated both guidance variants: HR dynamics and semantic guidance. For the semantic variant, language representations are pre-computed during onboarding and frozen before deployment. Therefore, inference does not require running a large language model or text encoder on-device. For both variants, runtime computation consists only of processing the physiological feature window, combining it with the static subject-level guidance representation, and producing the craving prediction.

%\textcolor{purple}{We benchmarked both the HR dynamics and semantic auxiliary guidance variants. For the semantic branch, the participant-level language representations are pre-computed and frozen prior to deployment. Consequently, inference for both variants does not require executing a large language model or text encoder on-device. Runtime computation is strictly limited to fusing physiological features with the static auxiliary guidance, followed by the prediction head.}

Because wearable sensing operates on incoming windows sequentially, we measured forward-pass latency with batch size $B=1$, reflecting real-time processing of a single physiological window~\cite{huang2025mmedge,sen2025low,chen2024overload}. Table~\ref{tab:runtime} reports three deployment-relevant metrics: mean latency, 99th percentile (P99) latency to capture worst-case runtime variation, and Resident Set Size (RSS) to measure peak memory footprint.
%\textcolor{purple}{Real-time wearable deployment is inherently sequential, requiring the immediate processing of sensor windows upon capture. We therefore benchmarked forward-pass inference latency at a batch size of $B=1$, reflecting the unbatched, live processing of a single physiological window during deployment \cite{huang2025mmedge,sen2025low,chen2024overload}. }
%\textcolor{purple}{Table \ref{tab:runtime} summarizes the results across three distinct device classes: \textit{Server/Workstation} hardware, \textit{Mobile/Consumer} electronics, and \textit{Edge/IoT} platforms. To rigorously capture real-world operational constraints, we report three key metrics: mean forward-pass latency, 99th percentile (P99) latency to account for worst-case system interruptions and OS overhead, and Resident Set Size (RSS) to quantify the peak memory footprint during execution. }
Despite the HR dynamics guidance model containing more parameters (899,828) than the semantic-guidance model (469,236), both remain exceptionally lightweight. On the Apple M4 CPU, forward inference completes in 0.132 ms for semantic guidance and 0.154 ms for HR dynamics guidance. While the Apple MPS backend is slower at $B=1$—an expected outcome for small, unbatched models where kernel launch and device transfer overheads dominate—larger CUDA devices like the RTX 4090 and NVIDIA GB10 maintain highly stable, sub-millisecond forward latency.

Crucially, results on consumer and edge devices further support practical deployment. On a Google Pixel 6, inference completes in 14.730 ms for semantic guidance and 16.708 ms for HR dynamics guidance, with RSS below 250 MB for both variants. On a Raspberry Pi, inference takes 1.616 ms and 2.654 ms for semantic and HR dynamics guidance, respectively. Even on the slower Jetson Nano, the model remains responsive, with mean latency of 14.531 ms for the HR dynamics branch. These results indicate that \textbf{\emph{RETRACE}} can run locally on common mobile and edge devices within the timing constraints of wearable sensing pipelines~\cite{gao2020edgedrnn,chen2019deep,li2018edge}.

%\textcolor{purple}{Crucially, the Google Pixel 6 results confirm the viability of running \textbf{\emph{RETRACE}} natively on standard consumer smartphones. Forward inference completes in 14.730 ms for the semantic branch and 16.708 ms for the HR dynamics branch. Furthermore, the memory footprint (RSS) remains under 250 MB for both variants, ensuring the model can operate comfortably in the background without draining resources or disrupting the user's primary applications.}

%\textcolor{purple}{The edge-device results also fall well within the timing constraints of wearable sensing pipelines \cite{gao2020edgedrnn,chen2019deep,li2018edge}. On the Raspberry Pi, forward inference takes 1.616 ms with semantic guidance and 2.654 ms with HR dynamics guidance. While the Jetson Nano is comparatively slower, particularly for the HR dynamics branch, it still achieves a highly responsive mean forward inference time of 14.531 ms.}

\begin{table}[h!]
\vspace{-3mm}
\centering

\begin{minipage}{0.48\textwidth}
\centering
\textbf{(a) Semantic guidance branch}\\
\resizebox{\textwidth}{!}{%
\begin{tabular}{llccc}
\toprule
\textbf{Device Class} & \textbf{Hardware Specification} & \multicolumn{3}{c}{\textbf{Inference}}  \\
\cmidrule(lr){3-5}
 & & \textbf{Mean} & \textbf{P99} & \textbf{RSS}  \\
 & & (ms) & (ms) & (MB) \\
\midrule
\textit{Server/Workstation} & AMD Ryzen 9 7950X (CPU) & 0.294 & 0.369 & 405.7  \\
 & NVIDIA RTX 4090 (GPU) & 0.315 & 0.330 & 787.4  \\
 & Intel CPU (20 logical cores) & 1.721 & 2.356 & 436.9 \\
\textit{Mobile/Consumer} & Apple M4 (CPU) & 0.132 & 0.204 & 206.7 \\
 & Apple Silicon (MPS) & 1.226 & 1.456 & 365.6 \\
 & Google Pixel 6 & 14.730 & 21.588 & 191.5  \\
\textit{Edge/IoT} & NVIDIA GB10 platform (CPU) & 5.741 & 18.901 & 507.3 \\
 & NVIDIA GB10 (CUDA) & 0.329 & 0.342 & 987.4  \\
 & Raspberry Pi (aarch64 CPU) & 1.616 & 5.648 & 209.8\\
 & NVIDIA Jetson Nano (CPU) & 9.618 & 16.254 & 268.9 \\
\bottomrule
\end{tabular}%
}
\end{minipage}
\hfill
\begin{minipage}{0.48\textwidth}
\centering
\textbf{(b) HR dynamics guidance branch}\\
\resizebox{\textwidth}{!}{%
\begin{tabular}{llccc}
\toprule
\textbf{Device Class} & \textbf{Hardware Specification} & \multicolumn{3}{c}{\textbf{Inference}}  \\
\cmidrule(lr){3-5} 
 & & \textbf{Mean} & \textbf{P99} & \textbf{RSS}  \\
 & & (ms) & (ms) & (MB)  \\
\midrule
\textit{Server/Workstation} & AMD Ryzen 9 7950X (CPU) & 0.339 & 0.345 & 533.1  \\
 & NVIDIA RTX 4090 (GPU) & 0.337 & 0.352 & 971.7  \\
 & Intel CPU (20 logical cores) & 2.181 & 2.752 & 499.3 \\
\textit{Mobile/Consumer} & Apple M4 (CPU) & 0.154 & 0.203 & 307.1  \\
 & Apple Silicon (MPS) & 1.378 & 1.604 & 468.1 \\
 & Google Pixel 6 & 16.708  & 27.515  & 241.5  \\
\textit{Edge/IoT} & NVIDIA GB10 platform (CPU) & 0.981 & 1.440 & 628.7 \\
 & NVIDIA GB10 (CUDA) & 0.376 & 0.386 & 1241.9  \\
 & Raspberry Pi (aarch64 CPU) & 2.654 & 2.964 & 272.8  \\
 & NVIDIA Jetson Nano (CPU) & 14.531 & 31.019 & 277.9 \\
\bottomrule
\end{tabular}%
}
\end{minipage}

\caption{Training and forward-pass runtime across hardware platforms for batch size $B=1$. Mean and P99 latency are reported in milliseconds. RSS memory denotes resident set size during execution. The Google Pixel 6 row is left blank because evaluation is ongoing.}
\vspace{-11mm}
\label{tab:runtime}
\end{table}

\section{Discussion and limitation}
\label{Discussion}

%The findings of this study suggest that robust, subject-independent physiological craving detection can be achieved by conditioning interpretation on trait-level information. Specifically, we show that resilience does not act as direct predictive evidence, but instead guides how physiological signals should be interpreted. This insight points to a broader modeling principle for human sensing in the presence of inter-individual differences. In this section, we discuss the practical implications of this perspective for real-world deployment, passive personalization, and multimodal design, and then outline key limitations that define the scope and generalizability of our findings.

This section discusses the main implications of \textbf{\emph{RETRACE}} for wearable craving sensing and outlines the limitations that define the scope of the current study.

\subsection{Key Implications}

\begin{itemize}

\item \textbf{Trait-conditioned interpretation without per-user training.}  
\textbf{\emph{RETRACE}} enables passive personalization by using resilience-related context to guide physiological inference, without per-user retraining or labeled craving data. This supports practical wearable deployment where user-specific craving labels are difficult to collect.

%\item \textbf{Person-aware inference does not require per-user training.}
%Rather than adapting or retraining models for each individual, \textbf{\emph{RETRACE}} enables passive personalization by using a fixed, shared classifier whose interpretation of physiological signals is conditioned on resilience-related autobiographical context. In general, per-user fine-tuning relies on collecting user-specific physiological data together with reliable labels or other forms of task-relevant supervision, which poses substantial challenges in real-world wearable deployments due to data collection burden and the need for sustained user interaction \cite{akbari2021meta,vijayan2021review}. These challenges are further amplified for craving detection and other applications involving sensitive or clinical populations, where additional ethical concerns, privacy risks, and participant burden make per-user adaptation particularly impractical. By requiring neither additional physiological data nor craving labels from the target user—and by eliminating user-specific fine-tuning—\textbf{\emph{RETRACE}} offers a scalable solution for personalized modulation of physiological signal interpretation, well suited for real-world deployment.

\item \textbf{Stress as a physiological reference for craving.}
Stress and craving share overlapping physiological responses, but stress signals are stronger and more consistent, whereas craving signals are weaker and more heterogeneous. \textbf{\emph{RETRACE}} uses a frozen stress-pretrained encoder as a physiological reference, while learning craving as a separate resilience-conditioned task. This allows the model to interpret craving-related deviations within broader stress-related physiology without treating stress itself as craving.

\item \textbf{Trade-offs between physiological and semantic guidance.}
The two guidance options offer different deployment advantages. 
HR-dynamics guidance is closely tied to post-stress autonomic recovery and may be preferable in clinical or supervised settings where standardized stress recovery protocols are feasible. Semantic guidance from autobiographical recall is easier to collect, can be reused over time, and may be more suitable for remote or in-the-wild deployment. The strong performance of both variants suggests that the benefit comes from trait-conditioned interpretation rather than from a single guidance modality.

%HR dynamics is closely tied to autonomic dynamics but requires structured stress–recovery protocols under comparable conditions, making it harder to obtain in practice. In contrast, autobiographical language can be collected more easily in unconstrained settings and reused over time as a compact representation of subject-level context. The consistent performance across both suggests that the benefit arises from trait-conditioned interpretation rather than any specific modality. These differences suggest that the choice of guidance should be scenario-dependent: HR dynamics may be preferable in controlled or clinical settings where standardized protocols are feasible, while semantic guidance is more suitable for in-the-wild deployment due to its low collection burden and reusability.

\item \textbf{Ethical and privacy considerations.}  
Autobiographical narratives may contain sensitive personal information, so deployment requires careful privacy protection and user control. In \textbf{\emph{RETRACE}}, raw narratives are not needed at inference time; only compact subject-level embeddings are used. This can reduce exposure of sensitive information, especially if embeddings are computed locally, stored securely, and made deletable by the user. However, embeddings may still encode private information, so misuse, stigmatization, and loss of user agency remain important concerns.

%Trait-level guidance is derived from sensitive autobiographical narratives, requiring careful handling to avoid misuse, stigmatization, and loss of user agency. In \textbf{\emph{RETRACE}}, only compact embeddings are used for inference, enabling on-device processing without storing or transmitting raw personal data, thereby reducing exposure of sensitive information while supporting personalized modeling.
\end{itemize}

This study also has important limitations:
\begin{itemize}
\item \textbf{Dataset and generalization.}  
Our evaluation uses one controlled laboratory dataset and a 14-day in-the-wild feasibility study with one participant. While collecting reliable craving annotations from clinical populations is challenging and comparable sample sizes are common in controlled IMWUT/CHI physiological studies~\cite{gullapalli2019body,gullapalli2021opitrack,ezer2024somaesthetic,zhao2023affective,carreiro2020wearable}, broader validation is needed to establish clinical deployability. Future work should evaluate \textbf{\emph{RETRACE}} on larger, more diverse, longitudinal real-world datasets to test generalization across populations, environments, and daily-life craving patterns.

\item \textbf{Proxy-based representation of resilience.}  
Both HR dynamics and autobiographical language are proxies for resilience-related individual differences, not direct measures of psychological resilience. HR dynamics depends on structured stress recovery measurement, while autobiographical recall may not fully capture the complexity of resilience-related traits. Future work should compare these proxies with validated resilience scales, clinical outcomes, and more structured elicitation methods, such as prompt-guided narratives or multimodal assessments.

%\item \textbf{Resilience Representation}
%While our framework leverages autobiographical language as a scalable proxy for resilience-related individual differences, the current elicitation relies on brief, open-ended memory recall and therefore captures only a coarse representation of a broader psychological construct. 
%An alternative could be coarse resilience categorization assessments based on interviews or self-reports \cite{o2021can}, which are subjective and typically require supervision. Moreover, our “\emph{Binary resilience embedding}”  evaluation in Section \ref{Ablation-Eval} shows that the coarse resilience categorizations do not capture the nuanced, person-specific contextual information necessary to effectively guide physiological interpretation for craving detection under subject-independent generalization. That suggests that relying solely on survey- or interview-based resilience labels—rather than semantically grounded autobiographical language representations—in \textbf{\emph{RETRACE}} would likely be insufficient for achieving robust inter-individual craving detection.
%However, future work could explore more targeted and structured elicitation strategies—such as prompt-guided narratives designed to probe coping strategies, stress appraisal, and recovery experiences—which may yield more precise and interpretable person-level context while remaining practical for deployment.

\end{itemize}

%\section{Data Availability and Ethics}

%To support reproducibility, model implementation details and code are provided in Appendix~\ref{appendix:reproduce}. The extracted physiological features will be made publicly available upon acceptance. This study involves a sensitive clinical population. Due to IRB restrictions, raw data and any identifiable information cannot be shared. In particular, autobiographical memory recall transcriptions are excluded from public release due to their sensitive nature.

\section{Conclusion}
\label{Conclusion}

This work establishes wearable craving detection in OUD as a problem of subject-aware physiological interpretation, rather than only short-window state classification. Our findings show that craving is physiologically measurable but difficult to isolate because it is weak, heterogeneous, and intertwined with stress-related autonomic activity. They also show that resilience is not a momentary wearable signal, but a subject-level factor that helps explain why similar physiological patterns may carry different meanings across individuals. By using resilience-related guidance to condition physiological interpretation, \textbf{\emph{RETRACE}} enables subject-aware craving estimation without requiring per-user craving labels or retraining. More broadly, this work shows that wearable health systems for heterogeneous and clinically sensitive populations can benefit from treating personal context not only as additional input, but as a mechanism for interpreting ambiguous physiological evidence. While further validation in larger, more diverse, and longer-term real-world OUD cohorts is needed, \textbf{\emph{RETRACE}} provides a practical step toward scalable, subject-aware wearable sensing to support OUD recovery.

%This work reframes physiological craving detection as a trait-conditioned inference problem, showing that inter-individual heterogeneity in OUD cannot be resolved by invariant domain-generalization modeling alone. By separating state-level autonomic evidence from trait-level resilience modulation, \textbf{\emph{RETRACE}} achieves robust subject-independent generalization without per-user retraining. Beyond introducing the first dataset jointly capturing stress, craving, and resilience-related context, this work establishes a deployable modeling paradigm for passive personalized, privacy-preserving craving detection. These findings lay the methodological foundation for future large-scale and in-the-wild validation of resilience-guided physiological inference in opioid and substance use disorder populations.
 
\bibliographystyle{ACM-Reference-Format}
\bibliography{sample-base}

@article{waite2006controlling,
  title={Controlling the false discovery rate and increasing statistical power in ecological studies},
  author={Waite, Thomas A and Campbell, Lesley G},
  journal={Ecoscience},
  volume={13},
  number={4},
  pages={439--442},
  year={2006},
  publisher={Taylor \& Francis}
}

@article{weinstein1997imagery,
  title={Imagery of craving in opiate addicts undergoing detoxification},
  author={Weinstein, Aviv and Wilson, Sue and Bailey, Jayne and Myles, Judy and Nutt, David},
  journal={Drug and Alcohol Dependence},
  volume={48},
  number={1},
  pages={25--31},
  year={1997},
  publisher={Elsevier}
}

@article{fredrickson2001role,
  title={The role of positive emotions in positive psychology: The broaden-and-build theory of positive emotions.},
  author={Fredrickson, Barbara L},
  journal={American psychologist},
  volume={56},
  number={3},
  pages={218},
  year={2001},
  publisher={American Psychological Association}
}

@article{tugade2004resilient,
  title={Resilient individuals use positive emotions to bounce back from negative emotional experiences.},
  author={Tugade, Michele M and Fredrickson, Barbara L},
  journal={Journal of personality and social psychology},
  volume={86},
  number={2},
  pages={320},
  year={2004},
  publisher={American Psychological Association}
}

@inproceedings{meegahapola2026stress,
  title={Stress Mindset Matters: Rethinking Mental Stress Detection with Multimodal Wearable Sensors},
  author={Meegahapola, Lakmal and Constantinides, Marios and Radivojevic, Zoran and Li, Hongwei and Eggleston, Michael and Quercia, Daniele},
  booktitle={Proceedings of the 2026 CHI Conference on Human Factors in Computing Systems},
  pages={1--26},
  year={2026}
}

@article{adler2016incremental,
  title={The incremental validity of narrative identity in predicting well-being: A review of the field and recommendations for the future},
  author={Adler, Jonathan M and Lodi-Smith, Jennifer and Philippe, Frederick L and Houle, Iliane},
  journal={Personality and Social Psychology Review},
  volume={20},
  number={2},
  pages={142--175},
  year={2016},
  publisher={Sage Publications Sage CA: Los Angeles, CA}
}

@article{mcadams2001psychology,
  title={The psychology of life stories},
  author={McAdams, Dan P},
  journal={Review of general psychology},
  volume={5},
  number={2},
  pages={100--122},
  year={2001},
  publisher={SAGE Publications Sage CA: Los Angeles, CA}
}

@article{conway2000construction,
  title={The construction of autobiographical memories in the self-memory system.},
  author={Conway, Martin A and Pleydell-Pearce, Christopher W},
  journal={Psychological review},
  volume={107},
  number={2},
  pages={261},
  year={2000},
  publisher={American Psychological Association}
}

@article{anderson2018ethical,
  title={Ethical issues in research involving participants with opioid use disorder},
  author={Anderson, Emily and McNair, Lindsay},
  journal={Therapeutic innovation \& regulatory science},
  volume={52},
  number={3},
  pages={280--284},
  year={2018},
  publisher={SAGE Publications Sage CA: Los Angeles, CA}
}

@article{wolke2025systematic,
  title={A systematic review of conceptualizations and statistical methods in longitudinal studies of resilience},
  author={Wolke, Dieter and Zhou, Yanlin and Liu, Yiwen and Eves, Robert and Mendon{\c{c}}a, Marina and Twilhaar, E Sabrina},
  journal={Nature Mental Health},
  volume={3},
  number={9},
  pages={1088--1099},
  year={2025},
  publisher={Nature Publishing Group US New York}
}

@article{tai2023conceptualizing,
  title={Conceptualizing psychological resilience through resting-state functional MRI in a mentally healthy population: a systematic review},
  author={Tai, Alan PL and Leung, Mei-Kei and Geng, Xiujuan and Lau, Way KW},
  journal={Frontiers in behavioral neuroscience},
  volume={17},
  pages={1175064},
  year={2023},
  publisher={Frontiers Media SA}
}

@article{enkema2020disrupting,
  title={Disrupting the path to craving: Acting without awareness mediates the link between negative affect and craving.},
  author={Enkema, Matthew C and Hallgren, Kevin A and Neilson, Elizabeth C and Bowen, Sarah and Bird, Elizabeth R and Larimer, Mary E},
  journal={Psychology of Addictive Behaviors},
  volume={34},
  number={5},
  pages={620},
  year={2020},
  publisher={American Psychological Association}
}

@article{vafaie2022association,
  title={Association of drug cues and craving with drug use and relapse: a systematic review and meta-analysis},
  author={Vafaie, Nilofar and Kober, Hedy},
  journal={JAMA psychiatry},
  volume={79},
  number={7},
  pages={641--650},
  year={2022}
}

@article{hoffman2019opioid,
  title={Opioid use disorder and treatment: challenges and opportunities},
  author={Hoffman, Kim A and Ponce Terashima, Javier and McCarty, Dennis},
  journal={BMC health services research},
  volume={19},
  number={1},
  pages={884},
  year={2019},
  publisher={Springer}
}

@article{wang2025global,
  title={Global, regional, and national trends and burden of opioid use disorder in individuals aged 15 years and above: 1990 to 2021 and projections to 2040},
  author={Wang, Shuailei and He, Yumiao and Huang, Yuguang},
  journal={Epidemiology and Psychiatric Sciences},
  volume={34},
  pages={e32},
  year={2025},
  publisher={Cambridge University Press}
}

@article{strang2020opioid,
  title={Opioid use disorder},
  author={Strang, John and Volkow, Nora D and Degenhardt, Louisa and Hickman, Matthew and Johnson, Kimberly and Koob, George F and Marshall, Brandon DL and Tyndall, Mark and Walsh, Sharon L},
  journal={Nature reviews Disease primers},
  volume={6},
  number={1},
  pages={3},
  year={2020},
  publisher={Nature Publishing Group UK London}
}

@article{draghmeh2025review,
  title={A Review of Relapse in Opioid Use Disorder},
  author={Draghmeh, Khaled and Gold, Mark S and Fuehrlein, Brian},
  journal={Current Addiction Reports},
  volume={12},
  number={1},
  pages={58},
  year={2025},
  publisher={Springer}
}

@article{perski2022technology,
  title={Technology-mediated just-in-time adaptive interventions (JITAIs) to reduce harmful substance use: a systematic review},
  author={Perski, Olga and H{\'e}bert, Emily T and Naughton, Felix and Hekler, Eric B and Brown, Jamie and Businelle, Michael S},
  journal={Addiction},
  volume={117},
  number={5},
  pages={1220--1241},
  year={2022},
  publisher={Wiley Online Library}
}

@article{garland2023zoom,
  title={Zoom-based mindfulness-oriented recovery enhancement plus just-in-time mindfulness practice triggered by wearable sensors for opioid craving and chronic pain},
  author={Garland, Eric L and Gullapalli, Bhanu T and Prince, Kort C and Hanley, Adam W and Sanyer, Mathias and Tuomenoksa, Mark and Rahman, Tauhidur},
  journal={Mindfulness},
  volume={14},
  number={6},
  pages={1329--1345},
  year={2023},
  publisher={Springer}
}

@article{sayette2016role,
  title={The role of craving in substance use disorders: theoretical and methodological issues},
  author={Sayette, Michael A},
  journal={Annual review of clinical psychology},
  volume={12},
  number={1},
  pages={407--433},
  year={2016},
  publisher={Annual Reviews}
}

@article{goldstein2013differential,
  title={Differential responses of components of the autonomic nervous system},
  author={Goldstein, David S},
  journal={Handbook of clinical neurology},
  volume={117},
  pages={13--22},
  year={2013},
  publisher={Elsevier}
}

@article{allen1989patterns,
  title={Patterns of autonomic response during laboratory stressors},
  author={Allen, Michael T and Crowell, Michael D},
  journal={Psychophysiology},
  volume={26},
  number={5},
  pages={603--614},
  year={1989},
  publisher={Wiley Online Library}
}

@article{nahum2016just,
  title={Just-in-time adaptive interventions (JITAIs) in mobile health: key components and design principles for ongoing health behavior support},
  author={Nahum-Shani, Inbal and Smith, Shawna N and Spring, Bonnie J and Collins, Linda M and Witkiewitz, Katie and Tewari, Ambuj and Murphy, Susan A},
  journal={Annals of behavioral medicine},
  pages={1--17},
  year={2016},
  publisher={Springer}
}

@article{chouhan2026supervised,
  title={Supervised information gain-based feature selection for multimodal physiological signals in stress prediction},
  author={Chouhan, Shefali Arora},
  journal={Scientific Reports},
  year={2026},
  publisher={Nature Publishing Group}
}

@inproceedings{li2018edge,
  title={Edge intelligence: On-demand deep learning model co-inference with device-edge synergy},
  author={Li, En and Zhou, Zhi and Chen, Xu},
  booktitle={Proceedings of the 2018 workshop on mobile edge communications},
  pages={31--36},
  year={2018}
}

@article{chen2019deep,
  title={Deep learning with edge computing: A review},
  author={Chen, Jiasi and Ran, Xukan},
  journal={Proceedings of the IEEE},
  volume={107},
  number={8},
  pages={1655--1674},
  year={2019},
  publisher={IEEE}
}

@article{Perski2021,
  title={Technology-mediated just-in-time adaptive interventions ({JITAIs}) to reduce harmful substance use: a systematic review},
  author={Perski, Olga and H{\'e}bert, Emily T. and Naughton, Felix and Hekler, Eric B. and Brown, Jamie and Businelle, Michael S.},
  journal={Addiction},
  volume={117},
  number={5},
  pages={1220--1241},
  year={2022},
  publisher={Wiley Online Library}
}

@article{Garland2023,
  title={Zoom-Based Mindfulness-Oriented Recovery Enhancement Plus Just-in-Time Mindfulness Practice Triggered by Wearable Sensors for Opioid Craving and Chronic Pain},
  author={Garland, Eric L. and Gullapalli, Bhanu T. and Prince, Kort C. and Hanley, Adam W. and Sanyer, Mathias and Tuomenoksa, Mark and Rahman, Tauhidur},
  journal={Mindfulness},
  volume={14},
  pages={1329--1345},
  year={2023},
  publisher={Springer}
}

@article{Hampton2025,
  title={Digital detection of craving and stress for individuals in recovery from substance use disorder: A qualitative study},
  author={Hampton, Jazmin and Eugene, Reynalde and Kapadia, Nirzari and Caggiano, Emily and Geagea, Amanda and Watson, C. James and Carreiro, Stephanie},
  journal={Drug and Alcohol Dependence Reports},
  volume={15},
  pages={100336},
  year={2025},
  publisher={Elsevier}
}

@inproceedings{King2025,
  title={Predicting Craving-Related Emotions among Opioid Use Disorder Patients: Preliminary Results},
  author={King, Zachary and Setiadi, Zoe and Hamdan, Liana and Ahmed, Hajar and Lamichhane, Bishal and Sabharwal, Ashutosh and Salas, Ramiro and Moukaddam, Nidal and Sano, Akane},
  booktitle={2025 IEEE 21st International Conference on Body Sensor Networks (BSN)},
  year={2025},
  organization={IEEE}
}

@article{gao2020edgedrnn,
  title={EdgeDRNN: Recurrent neural network accelerator for edge inference},
  author={Gao, Chang and Rios-Navarro, Antonio and Chen, Xi and Liu, Shih-Chii and Delbruck, Tobi},
  journal={IEEE Journal on Emerging and Selected Topics in Circuits and Systems},
  volume={10},
  number={4},
  pages={419--432},
  year={2020},
  publisher={IEEE}
}

@article{chirokoff2023craving,
  title={Craving dynamics and related cerebral substrates predict timing of use in alcohol, tobacco, and cannabis use disorders},
  author={Chirokoff, Valentine and Dupuy, Maud and Abdallah, Majd and Fatseas, Melina and Serre, Fuschia and Auriacombe, Marc and Misdrahi, David and Berthoz, Sylvie and Swendsen, Joel and Sullivan, Edith V and others},
  journal={Addiction neuroscience},
  volume={9},
  pages={100138},
  year={2023},
  publisher={Elsevier}
}

@article{grunevski2024predictive,
  title={Predictive Relationship Between Different Timescales of Opioid Craving and Opioid Use},
  author={Grunevski, Sergej and Kong, Julia and Konova, Anna},
  journal={Drug and Alcohol Dependence},
  volume={260},
  pages={109974},
  year={2024},
  publisher={Elsevier}
}

@article{baillet2024craving,
  title={Craving changes in first 14 days of addiction treatment: an outcome predictor of 5 years substance use status?},
  author={Baillet, Emmanuelle and Auriacombe, Marc and Romao, Cassandre and Garnier, H{\'e}l{\`e}ne and Gauld, Christophe and Vacher, Chlo{\'e} and Swendsen, Jo{\"e}l and Fatseas, M{\'e}lina and Serre, Fuschia},
  journal={Translational Psychiatry},
  volume={14},
  number={1},
  pages={497},
  year={2024},
  publisher={Nature Publishing Group UK London}
}

@inproceedings{chen2024overload,
  title={Overload: Latency attacks on object detection for edge devices},
  author={Chen, Erh-Chung and Chen, Pin-Yu and Chung, I and Lee, Che-Rung and others},
  booktitle={Proceedings of the IEEE/CVF Conference on Computer Vision and Pattern Recognition},
  pages={24716--24725},
  year={2024}
}

@article{lane2010survey,
  title={A survey of mobile phone sensing},
  author={Lane, Nicholas D and Miluzzo, Emiliano and Lu, Hong and Peebles, Daniel and Choudhury, Tanzeem and Campbell, Andrew T},
  journal={IEEE Communications magazine},
  volume={48},
  number={9},
  pages={140--150},
  year={2010},
  publisher={IEEE}
}

@article{sen2025low,
  title={Low-Latency Neural Inference on an Edge Device for Real-Time Handwriting Recognition from EEG Signals},
  author={Sen, Ovishake and Soni, Raghav and Virmani, Darpan and Parekh, Akshar and Lehman, Patrick and Jena, Sarthak and Katikhaneni, Adithi and Khalifa, Adam and Chatterjee, Baibhab},
  journal={arXiv preprint arXiv:2510.19832},
  year={2025}
}

@article{huang2025mmedge,
  title={MMEdge: Accelerating On-device Multimodal Inference via Pipelined Sensing and Encoding},
  author={Huang, Runxi and Yu, Mingxuan and Tsoi, Mingyu and Ouyang, Xiaomin},
  journal={arXiv preprint arXiv:2510.25327},
  year={2025}
}

@article{yurur2014context,
  title={Context-awareness for mobile sensing: A survey and future directions},
  author={Y{\"u}r{\"u}r, {\"O}zg{\"u}r and Liu, Chi Harold and Sheng, Zhengguo and Leung, Victor CM and Moreno, Wilfrido and Leung, Kin K},
  journal={IEEE Communications Surveys \& Tutorials},
  volume={18},
  number={1},
  pages={68--93},
  year={2014},
  publisher={IEEE}
}

@article{golden1978stroop,
  title={Stroop color and word test},
  author={Golden, Charles and Freshwater, Shawna M and Golden, Zarabeth},
  year={1978}
}

@article{qu2025nucleus,
  title={Nucleus accumbens dopamine release reflects bayesian inference during instrumental learning},
  author={Q{\"u}, Albert J and Tai, Lung-Hao and Hall, Christopher D and Tu, Emilie M and Eckstein, Maria K and Mishchanchuk, Karyna and Lin, Wan Chen and Chase, Juliana B and MacAskill, Andrew F and Collins, Anne GE and others},
  journal={PLOS Computational Biology},
  volume={21},
  number={7},
  pages={e1013226},
  year={2025},
  publisher={Public Library of Science San Francisco, CA USA}
}

@article{rathinam2021resilience,
  title={Resilience among abstinent individuals with substance use disorder},
  author={Rathinam, Bharath and Ezhumalai, Sinu},
  journal={Indian journal of psychiatric social work},
  volume={12},
  number={2},
  pages={96},
  year={2021}
}

@article{schultz2016dopamine,
  title={Dopamine reward prediction-error signalling: a two-component response},
  author={Schultz, Wolfram},
  journal={Nature reviews neuroscience},
  volume={17},
  number={3},
  pages={183--195},
  year={2016},
  publisher={Nature Publishing Group UK London}
}

@article{rahmati2021investigating,
  title={Investigating the moderating role of self-compassion in the relationship between resilience to stress and drug craving in drug-dependent men},
  author={Rahmati, Zahra and KHODABAKHSHI, KOOLAEE ANAHITA and Jahangiri, Mohammd Mehdi and others},
  year={2021},
  publisher={RESEARCH ON ADDICTION}
}

@article{zhao2023affective,
  title={Affective touch as immediate and passive wearable intervention},
  author={Zhao, Yiran and Tao, Yujie and Le, Grace and Maki, Rui and Adams, Alexander and Lopes, Pedro and Choudhury, Tanzeem},
  journal={Proceedings of the ACM on Interactive, Mobile, Wearable and Ubiquitous Technologies},
  volume={6},
  number={4},
  pages={1--23},
  year={2023},
  publisher={ACM New York, NY, USA}
}

@article{katsumi2020suppress,
  title={Suppress to feel and remember less: Neural correlates of explicit and implicit emotional suppression on perception and memory},
  author={Katsumi, Yuta and Dolcos, Sanda},
  journal={Neuropsychologia},
  volume={145},
  pages={106683},
  year={2020},
  publisher={Elsevier}
}

@article{engen2018memory,
  title={Memory control: A fundamental mechanism of emotion regulation},
  author={Engen, Haakon G and Anderson, Michael C},
  journal={Trends in Cognitive Sciences},
  volume={22},
  number={11},
  pages={982--995},
  year={2018},
  publisher={Elsevier}
}

@article{pascuzzi2017emotion,
  title={Emotion regulation, autobiographical memories and life narratives},
  author={Pascuzzi, Debora and Smorti, Andrea},
  journal={New Ideas in Psychology},
  volume={45},
  pages={28--37},
  year={2017},
  publisher={Elsevier}
}

@article{colombo2021moderating,
  title={The moderating role of emotion regulation in the recall of negative autobiographical memories},
  author={Colombo, Desir{\'e}e and Serino, Silvia and Suso-Ribera, Carlos and Fern{\'a}ndez-{\'A}lvarez, Javier and Cipresso, Pietro and Garc{\'\i}a-Palacios, Azucena and Riva, Giuseppe and Botella, Cristina},
  journal={International Journal of Environmental Research and Public Health},
  volume={18},
  number={13},
  pages={7122},
  year={2021},
  publisher={MDPI}
}

@article{van2013neuroimaging,
  title={Neuroimaging resilience to stress: a review},
  author={van der Werff, Steven JA and van den Berg, Susan M and Pannekoek, Justine Nienke and Elzinga, Bernet M and Van Der Wee, Nic JA},
  journal={Frontiers in behavioral neuroscience},
  volume={7},
  pages={39},
  year={2013},
  publisher={Frontiers Media SA}
}

@article{troy2023psychological,
  title={Psychological resilience: an affect-regulation framework},
  author={Troy, Allison S and Willroth, Emily C and Shallcross, Amanda J and Giuliani, Nicole R and Gross, James J and Mauss, Iris B},
  journal={Annual review of psychology},
  volume={74},
  number={1},
  pages={547--576},
  year={2023},
  publisher={Annual Reviews}
}

@article{troy2011resilience,
  title={Resilience in the face of stress: Emotion regulation as a protective factor},
  author={Troy, Allison S and Mauss, Iris B},
  journal={Resilience and mental health: Challenges across the lifespan},
  volume={1},
  number={2},
  pages={30--44},
  year={2011}
}

@article{compas2017coping,
  title={Coping, emotion regulation, and psychopathology in childhood and adolescence: A meta-analysis and narrative review.},
  author={Compas, Bruce E and Jaser, Sarah S and Bettis, Alexandra H and Watson, Kelly H and Gruhn, Meredith A and Dunbar, Jennifer P and Williams, Ellen and Thigpen, Jennifer C},
  journal={Psychological bulletin},
  volume={143},
  number={9},
  pages={939},
  year={2017},
  publisher={American Psychological Association}
}

@article{aldao2010emotion,
  title={Emotion-regulation strategies across psychopathology: A meta-analytic review},
  author={Aldao, Amelia and Nolen-Hoeksema, Susan and Schweizer, Susanne},
  journal={Clinical psychology review},
  volume={30},
  number={2},
  pages={217--237},
  year={2010},
  publisher={Elsevier}
}

@article{schetter2011resilience,
  title={Resilience in the context of chronic stress and health in adults},
  author={Schetter, Christine Dunkel and Dolbier, Christyn},
  journal={Social and personality psychology compass},
  volume={5},
  number={9},
  pages={634--652},
  year={2011},
  publisher={Wiley Online Library}
}

@article{pehlivanouglu2005computer,
  title={Computer adapted Stroop colour-word conflict test as a laboratory stress model},
  author={Pehlivano{\u{g}}lu, Bilge and Durmazlar, Nezih and Balkanc{\i}, Dicle},
  journal={Journal of Clinical Practice and Research},
  volume={27},
  number={2},
  pages={58},
  year={2005}
}

@article{karthikeyan2014analysis,
  title={Analysis of stroop color word test-based human stress detection using electrocardiography and heart rate variability signals},
  author={Karthikeyan, P and Murugappan, M and Yaacob, S},
  journal={Arabian Journal for Science and Engineering},
  volume={39},
  number={3},
  pages={1835--1847},
  year={2014},
  publisher={Springer}
}

@article{breese2005stress,
  title={Stress enhancement of craving during sobriety: a risk for relapse},
  author={Breese, George R and Chu, Kathleen and Dayas, Christopher V and Funk, Douglas and Knapp, Darin J and Koob, George F and L{\^e}, Dzung Anh and O’Dell, Laura E and Overstreet, David H and Roberts, Amanda J and others},
  journal={Alcoholism: Clinical and Experimental Research},
  volume={29},
  number={2},
  pages={185--195},
  year={2005},
  publisher={Wiley Online Library}
}

@article{grusser2006relationship,
  title={The relationship of stress, coping, effect expectancies and craving},
  author={Gr{\"u}sser, Sabine Miriam and M{\"o}rsen, Chantal Patricia and W{\"o}lfling, Klaus and Flor, Herta},
  journal={European Addiction Research},
  volume={13},
  number={1},
  pages={31--38},
  year={2006},
  publisher={S. Karger AG Basel, Switzerland}
}

@article{preston2011stress,
  title={Stress in the daily lives of cocaine and heroin users: relationship to mood, craving, relapse triggers, and cocaine use},
  author={Preston, Kenzie L and Epstein, David H},
  journal={Psychopharmacology},
  volume={218},
  number={1},
  pages={29--37},
  year={2011},
  publisher={Springer}
}

@article{mcewen2020revisiting,
  title={Revisiting the stress concept: implications for affective disorders},
  author={McEwen, Bruce S and Akil, Huda},
  journal={Journal of Neuroscience},
  volume={40},
  number={1},
  pages={12--21},
  year={2020},
  publisher={Society for Neuroscience}
}

@inproceedings{seikavandi2025mumtaffect,
  title={MuMTAffect: A Multimodal Multitask Affective Framework for Personality and Emotion Recognition from Physiological Signals},
  author={Seikavandi, Meisam Jamshidi and Narcizo, Fabricio Batista and Vucurevich, Ted and Dittberner, Andrew Burke and Burelli, Paolo},
  booktitle={Proceedings of the 3rd International Workshop on Multimodal and Responsible Affective Computing},
  pages={100--108},
  year={2025}
}

@article{singh2024visiophysioenet,
  title={VisioPhysioENet: Multimodal Engagement Detection using Visual and Physiological Signals},
  author={Singh, Alakhsimar and Verma, Nischay and Goyal, Kanav and Singh, Amritpal and Kumar, Puneet and Li, Xiaobai},
  journal={arXiv preprint arXiv:2409.16126},
  year={2024}
}

@article{wu2025comprehensive,
  title={A comprehensive review of multimodal emotion recognition: Techniques, challenges, and future directions},
  author={Wu, You and Mi, Qingwei and Gao, Tianhan},
  journal={Biomimetics},
  volume={10},
  number={7},
  pages={418},
  year={2025},
  publisher={MDPI}
}

@article{li2025multimodal,
  title={Multimodal physiological signals from wearable sensors for affective computing: A systematic review},
  author={Li, Fang and Zhang, Dan},
  journal={Intelligent Sports and Health},
  volume={1},
  number={4},
  pages={210--222},
  year={2025},
  publisher={Elsevier}
}

@article{mcnaboe2025optimizing,
  title={Optimizing Sensor Locations for Electrodermal Activity Monitoring Using a Wearable Belt System},
  author={McNaboe, Riley Q and Kong, Youngsun and Henderson, Wendy A and Cong, Xiaomei and Li, Aolan and Seo, Min-Hee and Chen, Ming-Hui and Feng, Bin and Posada-Quintero, Hugo F},
  journal={Journal of sensor and actuator networks},
  volume={14},
  number={2},
  pages={31},
  year={2025},
  publisher={MDPI}
}

@article{king2019micro,
  title={Micro-stress EMA: A passive sensing framework for detecting in-the-wild stress in pregnant mothers},
  author={King, Zachary D and Moskowitz, Judith and Egilmez, Begum and Zhang, Shibo and Zhang, Lida and Bass, Michael and Rogers, John and Ghaffari, Roozbeh and Wakschlag, Laurie and Alshurafa, Nabil},
  journal={Proceedings of the ACM on interactive, mobile, wearable and ubiquitous technologies},
  volume={3},
  number={3},
  pages={1--22},
  year={2019},
  publisher={ACM New York, NY, USA}
}

@inproceedings{caine2016local,
  title={Local standards for sample size at CHI},
  author={Caine, Kelly},
  booktitle={Proceedings of the 2016 CHI conference on human factors in computing systems},
  pages={981--992},
  year={2016}
}

@inproceedings{ezer2024somaesthetic,
  title={Somaesthetic Meditation Wearable: Exploring the Effect of Targeted Warmth Technology on Meditators' Experiences},
  author={Ezer, Talia Sofia and Giron, Jonathan and Erel, Hadas and Zuckerman, Oren},
  booktitle={Proceedings of the 2024 CHI Conference on Human Factors in Computing Systems},
  pages={1--14},
  year={2024}
}

@article{dehghani2019quantitative,
  title={A quantitative comparison of overlapping and non-overlapping sliding windows for human activity recognition using inertial sensors},
  author={Dehghani, Akbar and Sarbishei, Omid and Glatard, Tristan and Shihab, Emad},
  journal={Sensors},
  volume={19},
  number={22},
  pages={5026},
  year={2019},
  publisher={MDPI}
}

@inproceedings{mccarthy2016validation,
  title={Validation of the Empatica E4 wristband},
  author={McCarthy, Cameron and Pradhan, Nikhilesh and Redpath, Calum and Adler, Andy},
  booktitle={2016 IEEE EMBS international student conference (ISC)},
  pages={1--4},
  year={2016},
  organization={IEEE}
}

@article{benjamini1995controlling,
  title={Controlling the false discovery rate: a practical and powerful approach to multiple testing},
  author={Benjamini, Yoav and Hochberg, Yosef},
  journal={Journal of the Royal statistical society: series B (Methodological)},
  volume={57},
  number={1},
  pages={289--300},
  year={1995},
  publisher={Wiley Online Library}
}

@article{dastranj2024forecasting,
  title={Forecasting Mortality Rates: Unveiling Patterns with a PCA-GEE Approach},
  author={Dastranj, Reza and Kolar, Martin},
  journal={arXiv preprint arXiv:2407.01547},
  year={2024}
}

@article{liang1986longitudinal,
  title={Longitudinal data analysis using generalized linear models},
  author={Liang, Kung-Yee and Zeger, Scott L},
  journal={Biometrika},
  volume={73},
  number={1},
  pages={13--22},
  year={1986},
  publisher={Oxford University Press}
}

@article{chatzaki2025overview,
  title={An overview of stress analysis based on physiological signals: Systematic review of open datasets and current trends},
  author={Chatzaki, Chariklia and Tsiknakis, Manolis},
  journal={Sensors},
  volume={25},
  number={23},
  pages={7108},
  year={2025},
  publisher={MDPI}
}

@article{iqbal2022stress,
  title={Stress monitoring using wearable sensors: a pilot study and stress-predict dataset},
  author={Iqbal, Talha and Simpkin, Andrew J and Roshan, Davood and Glynn, Nicola and Killilea, John and Walsh, Jane and Molloy, Gerard and Ganly, Sandra and Ryman, Hannah and Coen, Eileen and others},
  journal={Sensors},
  volume={22},
  number={21},
  pages={8135},
  year={2022},
  publisher={Multidisciplinary Digital Publishing Institute}
}

@article{hui2025adjusted,
  title={Adjusted predictions for generalized estimating equations},
  author={Hui, Francis KC and Muller, Samuel and Welsh, Alan H},
  journal={Biometrics},
  volume={81},
  number={3},
  pages={ujaf090},
  year={2025},
  publisher={Oxford University Press}
}

@article{fox2008enhanced,
  title={Enhanced sensitivity to stress and drug/alcohol craving in abstinent cocaine-dependent individuals compared to social drinkers},
  author={Fox, Helen C and Hong, Kwang-Ik A and Siedlarz, Kristen and Sinha, Rajita},
  journal={Neuropsychopharmacology},
  volume={33},
  number={4},
  pages={796--805},
  year={2008},
  publisher={Nature Publishing Group}
}

@article{sinha2009modeling,
  title={Modeling stress and drug craving in the laboratory: implications for addiction treatment development},
  author={Sinha, Rajita},
  journal={Addiction biology},
  volume={14},
  number={1},
  pages={84--98},
  year={2009},
  publisher={Wiley Online Library}
}

@article{sinha2000psychological,
  title={Psychological stress, drug-related cues and cocaine craving},
  author={Sinha, Rajita and Fuse, Tiffany and Aubin, L-R and O'Malley, S Stephanie},
  journal={Psychopharmacology},
  volume={152},
  number={2},
  pages={140--148},
  year={2000},
  publisher={Springer}
}

@article{den2022conceptualizing,
  title={Conceptualizing and measuring psychological resilience: What can we learn from physics?},
  author={den Hartigh, Ruud and Hill, Yannick},
  journal={New Ideas in Psychology},
  volume={66},
  pages={100934},
  year={2022},
  publisher={PERGAMON-ELSEVIER SCIENCE LTD}
}

@article{windle2011resilience,
  title={What is resilience? A review and concept analysis},
  author={Windle, Gill},
  journal={Reviews in clinical gerontology},
  volume={21},
  number={2},
  pages={152--169},
  year={2011},
  publisher={Cambridge university press}
}

@article{moon2024comparative,
  title={Comparative feasibility study of physiological signals from wristband-type wearable sensors to assess occupants' thermal comfort},
  author={Moon, Sungwoo and Kim, Sun Sook and Choi, Byungjoo},
  journal={Energy and Buildings},
  volume={308},
  pages={114032},
  year={2024},
  publisher={Elsevier}
}

@article{wang2023wearable,
  title={Wearable sensors for activity monitoring and motion control: A review},
  author={Wang, Xiaoming and Yu, Hongliu and Kold, S{\o}ren and Rahbek, Ole and Bai, Shaoping},
  journal={Biomimetic Intelligence and Robotics},
  volume={3},
  number={1},
  pages={100089},
  year={2023},
  publisher={Elsevier}
}

@article{sinha2024stress,
  title={Stress and substance use disorders: risk, relapse, and treatment outcomes},
  author={Sinha, Rajita and others},
  journal={The Journal of clinical investigation},
  volume={134},
  number={16},
  year={2024},
  publisher={American Society for Clinical Investigation}
}

@article{koob2013addiction,
  title={Addiction is a reward deficit and stress surfeit disorder},
  author={Koob, George F},
  journal={Frontiers in psychiatry},
  volume={4},
  pages={72},
  year={2013},
  publisher={Frontiers Media SA}
}

@article{koob2008addiction,
  title={Addiction and the brain antireward system},
  author={Koob, George F and Le Moal, Michel},
  journal={Annu. Rev. Psychol.},
  volume={59},
  number={1},
  pages={29--53},
  year={2008},
  publisher={Annual Reviews}
}

@article{koob2007stress,
  title={Stress, dysregulation of drug reward pathways, and the transition to drug dependence},
  author={Koob, George and Kreek, Mary Jeanne},
  journal={American journal of psychiatry},
  volume={164},
  number={8},
  pages={1149--1159},
  year={2007},
  publisher={American Psychiatric Association}
}

@article{koob2014addiction,
  title={Addiction as a stress surfeit disorder},
  author={Koob, George F and Buck, Cara L and Cohen, Ami and Edwards, Scott and Park, Paula E and Schlosburg, Joel E and Schmeichel, Brooke and Vendruscolo, Leandro F and Wade, Carrie L and Whitfield Jr, Timothy W and others},
  journal={Neuropharmacology},
  volume={76},
  pages={370--382},
  year={2014},
  publisher={Elsevier}
}

@article{foll2021flirt,
  title={FLIRT: A feature generation toolkit for wearable data},
  author={F{\"o}ll, Simon and Maritsch, Martin and Spinola, Federica and Mishra, Varun and Barata, Filipe and Kowatsch, Tobias and Fleisch, Elgar and Wortmann, Felix},
  journal={Computer Methods and Programs in Biomedicine},
  volume={212},
  pages={106461},
  year={2021},
  publisher={Elsevier}
}

@article{rogers2024addiction,
  title={Addiction neurobiologists should study resilience},
  author={Rogers, Alexandra and Leslie, Frances},
  journal={Addiction Neuroscience},
  volume={11},
  pages={100152},
  year={2024},
  publisher={Elsevier}
}

@article{qin2022domain,
  title={Domain generalization for activity recognition via adaptive feature fusion},
  author={Qin, Xin and Wang, Jindong and Chen, Yiqiang and Lu, Wang and Jiang, Xinlong},
  journal={ACM Transactions on Intelligent Systems and Technology},
  volume={14},
  number={1},
  pages={1--21},
  year={2022},
  publisher={ACM New York, NY}
}

@article{apicella2024toward,
  title={Toward cross-subject and cross-session generalization in EEG-based emotion recognition: Systematic review, taxonomy, and methods},
  author={Apicella, Andrea and Arpaia, Pasquale and D’Errico, Giovanni and Marocco, Davide and Mastrati, Giovanna and Moccaldi, Nicola and Prevete, Roberto},
  journal={Neurocomputing},
  volume={604},
  pages={128354},
  year={2024},
  publisher={Elsevier}
}

@article{ungar2020resilience,
  title={Resilience and mental health: How multisystemic processes contribute to positive outcomes},
  author={Ungar, Michael and Theron, Linda},
  journal={The Lancet Psychiatry},
  volume={7},
  number={5},
  pages={441--448},
  year={2020},
  publisher={Elsevier}
}

@article{yamashita2021resilience,
  title={Resilience and related factors as predictors of relapse risk in patients with substance use disorder: a cross-sectional study},
  author={Yamashita, Ayako and Yoshioka, Shin-ichi and Yajima, Yuki},
  journal={Substance abuse treatment, prevention, and policy},
  volume={16},
  number={1},
  pages={40},
  year={2021},
  publisher={Springer}
}

@article{yang2020resilience,
  title={How resilience promotes mental health of patients with DSM-5 substance use disorder? The mediation roles of positive affect, self-esteem, and perceived social support},
  author={Yang, Chunyu and Zhou, You and Xia, Mengfan},
  journal={Frontiers in Psychiatry},
  volume={11},
  pages={588968},
  year={2020},
  publisher={Frontiers Media SA}
}

@article{cha1994partial,
  title={Partial least squares},
  author={Cha, Jaesung},
  journal={Adv. Methods Mark. Res},
  volume={407},
  pages={52--78},
  year={1994}
}

@incollection{crowell2013psychosocial,
  title={Psychosocial stress, emotion regulation, and resilience in adolescence},
  author={Crowell, Sheila E and Skidmore, Chloe R and Rau, Holly K and Williams, Paula G},
  booktitle={Handbook of adolescent health psychology},
  pages={129--141},
  year={2013},
  publisher={Springer}
}

@article{garcia2013integrative,
  title={An integrative study of autobiographical memory for positive and negative experiences},
  author={Garc{\'\i}a-Bajos, Elvira and Migueles, Malen},
  journal={The Spanish Journal of Psychology},
  volume={16},
  pages={E102},
  year={2013},
  publisher={Cambridge University Press}
}

@article{jeyadevan2024time,
  title={A Time-Limited Scoping Review Investigating the Use of Wearable Biosensors in the Substance Use Field},
  author={Jeyadevan, Ashani and Grigg, Jodie},
  journal={Current Addiction Reports},
  volume={11},
  number={5},
  pages={928--939},
  year={2024},
  publisher={Springer}
}

@String{Computing = "Computing" }

@String{Computer = "{IEEE} Computer" }

@String{Springer = "Springer-Verlag" }

@ArtifactSoftware{R,
    title = {R: A Language and Environment for Statistical Computing},
    author = {{R Core Team}},
    organization = {R Foundation for Statistical Computing},
    address = {Vienna, Austria},
    year = {2019},
    url = {https://www.R-project.org/},
}

@article{gullapalli2019body,
  title={On-body sensing of cocaine craving, euphoria and drug-seeking behavior using cardiac and respiratory signals},
  author={Gullapalli, Bhanu Teja and Natarajan, Annamalai and Angarita, Gustavo A and Malison, Robert T and Ganesan, Deepak and Rahman, Tauhidur},
  journal={Proceedings of the ACM on Interactive, Mobile, Wearable and Ubiquitous Technologies},
  volume={3},
  number={2},
  pages={1--31},
  year={2019},
  publisher={ACM New York, NY, USA}
}

@article{gullapalli2021opitrack,
  title={Opitrack: a wearable-based clinical opioid use tracker with temporal convolutional attention networks},
  author={Gullapalli, Bhanu Teja and Carreiro, Stephanie and Chapman, Brittany P and Ganesan, Deepak and Sjoquist, Jan and Rahman, Tauhidur},
  journal={Proceedings of the ACM on interactive, mobile, wearable and ubiquitous technologies},
  volume={5},
  number={3},
  pages={1--29},
  year={2021},
  publisher={ACM New York, NY, USA}
}

@article{maclean2019stress,
  title={Stress and opioid use disorder: A systematic review},
  author={MacLean, R Ross and Armstrong, Jessica L and Sofuoglu, Mehmet},
  journal={Addictive behaviors},
  volume={98},
  pages={106010},
  year={2019},
  publisher={Elsevier}
}

@article{almazrouei2023method,
  title={A method to induce stress in human subjects in online research environments},
  author={Almazrouei, Mohammed A and Morgan, Ruth M and Dror, Itiel E},
  journal={Behavior research methods},
  volume={55},
  number={5},
  pages={2575--2582},
  year={2023},
  publisher={Springer}
}

@article{dobriban2020permutation,
  title={Permutation methods for factor analysis and PCA},
  author={Dobriban, Edgar},
  year={2020}
}

@article{abdi2010principal,
  title={Principal component analysis},
  author={Abdi, Herv{\'e} and Williams, Lynne J},
  journal={Wiley interdisciplinary reviews: computational statistics},
  volume={2},
  number={4},
  pages={433--459},
  year={2010},
  publisher={Wiley Online Library}
}

@inproceedings{krueger2021out,
  title={Out-of-distribution generalization via risk extrapolation (rex)},
  author={Krueger, David and Caballero, Ethan and Jacobsen, Joern-Henrik and Zhang, Amy and Binas, Jonathan and Zhang, Dinghuai and Le Priol, Remi and Courville, Aaron},
  booktitle={International conference on machine learning},
  pages={5815--5826},
  year={2021},
  organization={PMLR}
}

@article{sagawa2019distributionally,
  title={Distributionally robust neural networks for group shifts: On the importance of regularization for worst-case generalization},
  author={Sagawa, Shiori and Koh, Pang Wei and Hashimoto, Tatsunori B and Liang, Percy},
  journal={arXiv preprint arXiv:1911.08731},
  year={2019}
}

@article{sinha1999stress,
  title={Stress-induced craving and stress response in cocaine dependent individuals},
  author={Sinha, Rajita and Catapano, Dana and O’Malley, Stephanie},
  journal={Psychopharmacology},
  volume={142},
  number={4},
  pages={343--351},
  year={1999},
  publisher={Springer}
}

@article{adler2012living,
  title={Living into the story: agency and coherence in a longitudinal study of narrative identity development and mental health over the course of psychotherapy.},
  author={Adler, Jonathan M},
  journal={Journal of personality and social psychology},
  volume={102},
  number={2},
  pages={367},
  year={2012},
  publisher={American Psychological Association}
}

@article{fischer2015motivational,
  title={Motivational basis of personality traits: A meta-analysis of value-personality correlations},
  author={Fischer, Ronald and Boer, Diana},
  journal={Journal of personality},
  volume={83},
  number={5},
  pages={491--510},
  year={2015},
  publisher={Wiley Online Library}
}

@article{xiao2025human,
  title={Human Heterogeneity Invariant Stress Sensing},
  author={Xiao, Yi and Sharma, Harshit and Kaur, Sawinder and Bergen-Cico, Dessa and Salekin, Asif},
  journal={Proceedings of the ACM on Interactive, Mobile, Wearable and Ubiquitous Technologies},
  volume={9},
  number={3},
  pages={1--42},
  year={2025},
  publisher={ACM New York, NY, USA}
}

@article{arjovsky2019invariant,
  title={Invariant risk minimization},
  author={Arjovsky, Martin and Bottou, L{\'e}on and Gulrajani, Ishaan and Lopez-Paz, David},
  journal={arXiv preprint arXiv:1907.02893},
  year={2019}
}

@inproceedings{jat2023harnessing,
  title={Harnessing the digital revolution: a comprehensive review of mHealth applications for remote monitoring in transforming healthcare delivery},
  author={Jat, Avnish Singh and Gr{\o}nli, Tor-Morten},
  booktitle={International Conference on Mobile Web and Intelligent Information Systems},
  pages={55--67},
  year={2023},
  organization={Springer}
}

@inproceedings{song2020information,
  title={Information leakage in embedding models},
  author={Song, Congzheng and Raghunathan, Ananth},
  booktitle={Proceedings of the 2020 ACM SIGSAC conference on computer and communications security},
  pages={377--390},
  year={2020}
}

@article{carreiro2024evaluation,
  title={Evaluation of a digital tool for detecting stress and craving in SUD recovery: An observational trial of accuracy and engagement},
  author={Carreiro, Stephanie and Ramanand, Pravitha and Taylor, Melissa and Leach, Rebecca and Stapp, Joshua and Sherestha, Sloke and Smelson, David and Indic, Premananda},
  journal={Drug and Alcohol Dependence},
  volume={261},
  pages={111353},
  year={2024},
  publisher={Elsevier}
}

@article{schmidt2018wearable,
  title={Wearable affect and stress recognition: A review},
  author={Schmidt, Philip and Reiss, Attila and Duerichen, Robert and Van Laerhoven, Kristof},
  journal={arXiv preprint arXiv:1811.08854},
  year={2018}
}

@article{shazeer2017outrageously,
  title={Outrageously large neural networks: The sparsely-gated mixture-of-experts layer},
  author={Shazeer, Noam and Mirhoseini, Azalia and Maziarz, Krzysztof and Davis, Andy and Le, Quoc and Hinton, Geoffrey and Dean, Jeff},
  journal={arXiv preprint arXiv:1701.06538},
  year={2017}
}

@article{vaswani2017attention,
  title={Attention is all you need},
  author={Vaswani, Ashish and Shazeer, Noam and Parmar, Niki and Uszkoreit, Jakob and Jones, Llion and Gomez, Aidan N and Kaiser, {\L}ukasz and Polosukhin, Illia},
  journal={Advances in neural information processing systems},
  volume={30},
  year={2017}
}

@inproceedings{perez2018film,
  title={Film: Visual reasoning with a general conditioning layer},
  author={Perez, Ethan and Strub, Florian and De Vries, Harm and Dumoulin, Vincent and Courville, Aaron},
  booktitle={Proceedings of the AAAI conference on artificial intelligence},
  volume={32},
  number={1},
  year={2018}
}

@article{jenner2025using,
  title={Using large language models for narrative analysis: a novel application of generative AI},
  author={Jenner, Sarah and Raidos, Dimitris and Anderson, Emma and Fleetwood, Stella and Ainsworth, Ben and Fox, Kerry and Kreppner, Jana and Barker, Mary},
  journal={Methods in Psychology},
  volume={12},
  pages={100183},
  year={2025},
  publisher={Elsevier}
}

@inproceedings{barros2025large,
  title={Large language model for qualitative research: A systematic mapping study},
  author={Barros, Cau{\~a} Ferreira and Azevedo, Bruna Borges and Neto, Valdemar Vicente Graciano and Kassab, Mohamad and Kalinowski, Marcos and Do Nascimento, Hugo Alexandre D and Bandeira, Michelle CGSP},
  booktitle={2025 IEEE/ACM International Workshop on Methodological Issues with Empirical Studies in Software Engineering (WSESE)},
  pages={48--55},
  year={2025},
  organization={IEEE}
}

@article{kirschbaum1993trier,
  title={The ‘Trier Social Stress Test’--a tool for investigating psychobiological stress responses in a laboratory setting},
  author={Kirschbaum, Clemens and Pirke, Karl-Martin and Hellhammer, Dirk H},
  journal={Neuropsychobiology},
  volume={28},
  number={1-2},
  pages={76--81},
  year={1993},
  publisher={Karger Publishers}
}

@article{makowski2021neurokit2,
  title={NeuroKit2: A Python toolbox for neurophysiological signal processing},
  author={Makowski, Dominique and Pham, Tam and Lau, Zen J and Brammer, Jan C and Lespinasse, Fran{\c{c}}ois and Pham, Hung and Sch{\"o}lzel, Christopher and Chen, SH Annabel},
  journal={Behavior research methods},
  pages={1--8},
  year={2021},
  publisher={Springer}
}

@article{connolly2018negative,
  title={Negative event recall as a vulnerability for depression: Relationship between momentary stress-reactive rumination and memory for daily life stress},
  author={Connolly, Samantha L and Alloy, Lauren B},
  journal={Clinical Psychological Science},
  volume={6},
  number={1},
  pages={32--47},
  year={2018},
  publisher={Sage Publications Sage CA: Los Angeles, CA}
}

@article{mereish2024vicarious,
  title={Vicarious heterosexism-based stress induces alcohol, nicotine, and cannabis craving and negative affect among sexual minority young adults: An experimental study},
  author={Mereish, Ethan H and Miranda Jr, Robert},
  journal={Neurobiology of Stress},
  volume={32},
  pages={100668},
  year={2024},
  publisher={Elsevier}
}

@article{brouwer2014new,
  title={A new paradigm to induce mental stress: the Sing-a-Song Stress Test (SSST)},
  author={Brouwer, Anne-Marie and Hogervorst, Maarten A},
  journal={Frontiers in neuroscience},
  volume={8},
  pages={224},
  year={2014},
  publisher={Frontiers Media SA}
}

@article{xiao2024reading,
  title={Reading between the heat: Co-teaching body thermal signatures for non-intrusive stress detection},
  author={Xiao, Yi and Sharma, Harshit and Zhang, Zhongyang and Bergen-Cico, Dessa and Rahman, Tauhidur and Salekin, Asif},
  journal={Proceedings of the ACM on Interactive, Mobile, Wearable and Ubiquitous Technologies},
  volume={7},
  number={4},
  pages={1--30},
  year={2024},
  publisher={ACM New York, NY, USA}
}

@article{schaefer2010assessing,
  title={Assessing the effectiveness of a large database of emotion-eliciting films: A new tool for emotion researchers},
  author={Schaefer, Alexandre and Nils, Fr{\'e}d{\'e}ric and Sanchez, Xavier and Philippot, Pierre},
  journal={Cognition and emotion},
  volume={24},
  number={7},
  pages={1153--1172},
  year={2010},
  publisher={Taylor \& Francis}
}

@article{tumanova2019autonomic,
  title={Autonomic nervous system response to speech production in stuttering and normally fluent preschool-age children},
  author={Tumanova, Victoria and Backes, Nicole},
  journal={Journal of Speech, Language, and Hearing Research},
  volume={62},
  number={11},
  pages={4030--4044},
  year={2019},
  publisher={ASHA}
}

@article{tulen1989characterization,
  title={Characterization of stress reactions to the Stroop Color Word Test},
  author={Tulen, JHM and Moleman, P and Van Steenis, HG and Boomsma, F},
  journal={Pharmacology Biochemistry and Behavior},
  volume={32},
  number={1},
  pages={9--15},
  year={1989},
  publisher={Elsevier}
}

@article{carreiro2020wearable,
  title={Wearable sensor-based detection of stress and craving in patients during treatment for substance use disorder: A mixed methods pilot study},
  author={Carreiro, Stephanie and Chintha, Keerthi Kumar and Shrestha, Sloke and Chapman, Brittany and Smelson, David and Indic, Premananda},
  journal={Drug and alcohol dependence},
  volume={209},
  pages={107929},
  year={2020},
  publisher={Elsevier}
}

@article{sharma2022psychophysiological,
  title={Psychophysiological Arousal in Young Children Who Stutter: An Interpretable AI Approach},
  author={Sharma, Harshit and Xiao, Yi and Tumanova, Victoria and Salekin, Asif},
  journal={Proceedings of the ACM on interactive, mobile, wearable and ubiquitous technologies},
  volume={6},
  number={3},
  pages={1--32},
  year={2022},
  publisher={ACM New York, NY, USA}
}

@inproceedings{ollander2016comparison, title={A comparison of wearable and stationary sensors for stress detection}, author={Ollander, Simon and Godin, Christelle and Campagne, Aur{\'e}lie and Charbonnier, Sylvie}, booktitle={2016 IEEE International Conference on Systems, Man, and Cybernetics (SMC)}, pages={004362--004366}, year={2016}, organization={IEEE} }

@article{menghini2019stressing,
  title={Stressing the accuracy: Wrist-worn wearable sensor validation over different conditions},
  author={Menghini, Luca and Gianfranchi, Evelyn and Cellini, Nicola and Patron, Elisabetta and Tagliabue, Mariaelena and Sarlo, Michela},
  journal={Psychophysiology},
  volume={56},
  number={11},
  pages={e13441},
  year={2019},
  publisher={Wiley Online Library}
}

@article{chapman2022impact,
  title={Impact of individual and treatment characteristics on wearable sensor-based digital biomarkers of opioid use},
  author={Chapman, Brittany P and Gullapalli, Bhanu Teja and Rahman, Tauhidur and Smelson, David and Boyer, Edward W and Carreiro, Stephanie},
  journal={NPJ Digital Medicine},
  volume={5},
  number={1},
  pages={123},
  year={2022},
  publisher={Nature Publishing Group UK London}
}

@inproceedings{motti2015users,
  title={Users’ privacy concerns about wearables: impact of form factor, sensors and type of data collected},
  author={Motti, Vivian Genaro and Caine, Kelly},
  booktitle={International conference on financial cryptography and data security},
  pages={231--244},
  year={2015},
  organization={Springer}
}

@article{zhang2025survey,
  title={A survey on security and privacy issues in wearable health monitoring devices},
  author={Zhang, Bonan and Chen, Chao and Lee, Ickjai and Lee, Kyungmi and Ong, Kok-Leong},
  journal={Computers \& Security},
  pages={104453},
  year={2025},
  publisher={Elsevier}
}

@article{wang2023smoking,
  title={Smoking behavior detection algorithm based on YOLOv8-MNC},
  author={Wang, Zhong and Lei, Lanfang and Shi, Peibei},
  journal={Frontiers in Computational Neuroscience},
  volume={17},
  pages={1243779},
  year={2023},
  publisher={Frontiers Media SA}
}

@article{krishnan2011partial,
  title={Partial Least Squares (PLS) methods for neuroimaging: a tutorial and review},
  author={Krishnan, Anjali and Williams, Lynne J and McIntosh, Anthony Randal and Abdi, Herv{\'e}},
  journal={Neuroimage},
  volume={56},
  number={2},
  pages={455--475},
  year={2011},
  publisher={Elsevier}
}

@article{abdi2010partial,
  title={Partial least squares regression and projection on latent structure regression (PLS Regression)},
  author={Abdi, Herv{\'e}},
  journal={Wiley interdisciplinary reviews: computational statistics},
  volume={2},
  number={1},
  pages={97--106},
  year={2010},
  publisher={Wiley Online Library}
}

@article{vos2023generalizable,
  title={Generalizable machine learning for stress monitoring from wearable devices: A systematic literature review},
  author={Vos, Gideon and Trinh, Kelly and Sarnyai, Zoltan and Azghadi, Mostafa Rahimi},
  journal={International Journal of Medical Informatics},
  volume={173},
  pages={105026},
  year={2023},
  publisher={Elsevier}
}

\appendix

\newpage

\section{Methodological Transparency and Reproducibility Appendix}
\label{appendix:reproduce}

This appendix documents implementation details, data usage, and evaluation procedures for the current version of the proposed framework. Due to IRB restrictions involving a vulnerable population with opioid use disorder (OUD), raw physiological recordings and autobiographical narratives cannot be publicly released. To support methodological transparency, we release the full analysis and modeling code used for statistical analysis, model training, and evaluation in an anonymized repository.
\href{https://anonymous.4open.science/r/Craving_model-16DC/README.md}{Anonymized GitHub link}.

\subsection{Current Main Model Implementation}
\label{appendix:mainmodel}

\paragraph{Overview.}
The main model operates on a 303-dimensional physiological feature vector together with a subject-level guidance representation. The guidance input can take two forms: (i) HR-dynamics-based embeddings derived from post-task heart-rate dynamics, or (ii) semantic embeddings derived from autobiographical memory narratives.Details on how these two guidance representations are extracted are provided in Appendix~\ref{app:guidance}.

In the HR-dynamics configuration, the subject representation is formed by concatenating a 16-dimensional post calm task embedding and a 16-dimensional post stress task embedding into a 32-dimensional vector. In the semantic configuration, the representation is formed by concatenating 1536-dimensional good-memory and bad-memory embeddings into a 3072-dimensional vector. 

The model follows a dual-path architecture consisting of a frozen stress branch and a trainable physiology branch. Subject-level guidance is applied through feature-level input gating and representation-level fusion. Table~\ref{tab:main_model_arch} summarizes the architecture of the HR-dynamics configuration.

\paragraph{Architecture Details.}
Each trainable module is implemented using standard feedforward blocks consisting of a linear transformation followed by GELU activation, layer normalization, and dropout where applicable. The frozen stress encoder shares the same MLP structure but is pretrained separately and remains fixed during craving training.

\begin{table}[t]
\centering
\caption{Architecture details of the best HR-AUC main model.}
\label{tab:main_model_arch}
\begin{tabular}{lccc}
\toprule
\textbf{Component} & \textbf{Layers} & \textbf{Input $\rightarrow$ Output} & \textbf{Notes} \\
\midrule

Stress Encoder & 4 & 303 $\rightarrow$ 384 & Frozen, dropout = 0.2 \\

Physiology Branch & 2 & 303 $\rightarrow$ 384  & Dropout = 0.193 \\

Gate Projection & 2 & 32 $\rightarrow$ 303 & Sigmoid output (feature gating) \\

Fusion Projection & 1 & 32 $\rightarrow$ 384 & Subject-conditioned \\

Fusion Gate & 2 & 3*384 $\rightarrow$  1 & Sample-wise scalar weight \\

Classifier & 3 & 384   $\rightarrow$ 2 & Binary logits \\

Aux Stress Head & 1 & 384 $\rightarrow$ 2 & Training only \\

\bottomrule
\end{tabular}
\end{table}

\paragraph{Subject-Guided Conditioning.}
The 64-dimensional HR-AUC embedding is first processed through a shared projection block. From this representation, two pathways are derived. The gate pathway produces feature-wise multiplicative gates that are applied element-wise to the 303-dimensional physiological input before it is passed to the trainable physiology branch. The fusion pathway produces a subject-conditioned representation used during branch fusion.

\paragraph{Fusion Mechanism.}
Fusion is implemented using a sample-dependent gating mechanism. The projected stress representation, the trainable physiology representation, and the subject-conditioned fusion representation are concatenated into a 1152-dimensional vector. This vector is processed by a two-layer gating network to produce a scalar fusion weight, which is used to interpolate between the frozen stress branch and the trainable physiology branch.

\paragraph{Classifier Head.}
The fused representation is passed through a three-layer MLP classifier, consisting of two hidden layers followed by a final linear layer that outputs logits for binary craving classification.

\paragraph{Training Configuration.}
The model is trained for 24 epochs with batch size 30 using AdamW. The selected hyperparameters are:
\begin{itemize}
    \item Learning rate: 0.00131
    \item Weight decay: $1.12 \times 10^{-6}$
    \item Gradient clipping: 0.424
    \item Dropout (trainable branch): 0.193
\end{itemize}
No learning-rate scheduler is used.

\paragraph{Normalization and Loss.}
Subject-wise z-score normalization is applied to all physiological features. The model is trained with cross-entropy loss using label smoothing of 0.0175.

\paragraph{Regularization.}
The following regularization terms are applied during training:
\begin{itemize}
    \item Input-gate sparsity weight: 0.00102
    \item Auxiliary stress loss weight: 0.000596
    \item Orthogonality regularization weight: 0.000541
\end{itemize}

\section{Physiological Feature Dimensionality and Quality Control}
\label{appendix:feature}

\paragraph{Feature Extraction.}
For each 30-second physiological window, we extracted a common multimodal feature space from acceleration (ACC), blood volume pulse (BVP), electrodermal activity (EDA), heart rate (HR), heart-rate variability (HRV), and skin temperature (TEMP). The initial shared feature space contained 334 dimensions.

The modality-wise feature dimensionalities were as follows: 99 ACC features, 34 BVP-derived features, 105 EDA features, 31 HR features, 35 HRV features, and 30 TEMP features.

Across modalities, the feature set included three broad families of descriptors:
\begin{itemize}
    \item \textbf{Summary statistics:} mean, median, standard deviation, variance, interquartile range, minimum, maximum, percentile summaries, skewness, and kurtosis,
    \item \textbf{Shape descriptors:} root mean square (RMS), mean absolute value, median absolute deviation, peak amplitude, crest factor, impulse factor, waveform length, and zero-crossing counts,
    \item \textbf{Spectral descriptors:} total spectral power, spectral centroid, peak frequency, bandwidth, spectral entropy, and band-limited power features.
\end{itemize}

In addition to these shared descriptors, modality-specific features were included. ACC features captured per-axis statistics (x/y/z), magnitude-based features, inter-axis correlations, signal energy, mean absolute temporal differences, and signal magnitude area. BVP features included waveform statistics, peak-based descriptors, and signal-quality indicators derived from PPG signals, from which HR and HRV features were computed. EDA features included tonic and phasic decomposition outputs and skin conductance response (SCR) event descriptors such as onset counts, peak counts, amplitudes, rise times, and recovery times. HR features included temporal difference measures, and TEMP features primarily consisted of statistical, shape, and spectral summaries.

\paragraph{Quality Control Procedure.}
  A global quality-control procedure was applied uniformly across all participants and conditions prior to model training. Features were excluded based on two criteria:
  \begin{itemize}
      \item \textbf{Missingness:} features with a missing-value rate greater than 20\% across windows,
      \item \textbf{Low variation:} features with fewer than two unique non-null values,

  \end{itemize}

Feature filtering was performed without reference to stress labels, craving labels, subject identity, or downstream model outcomes.

The removed features were concentrated in measurements known to be unreliable under short-window analysis, including low-frequency HRV components (e.g., ultra-low- and very-low-frequency bands), long-horizon HRV statistics (e.g., SDANN and SDNNI), and a small number of sparse or weakly varying spectral and zero-crossing descriptors across ACC, HR, EDA, and TEMP modalities.

\paragraph{Final Feature Space.}
After quality control and final feature harmonization, the predictive feature space used in all main experiments consisted of 303 physiological dimensions per window. This representation was held fixed across all models and experimental conditions.

\section{Additional Details for Physiological Modulation Analysis}
\label{app:rq1_methods}

\subsection{Physiological Systems Organization Strategy}
\label{feature:system}
For interpretability and higher-level analysis, extracted features in Section \ref{sec:feature} were further organized into four physiological systems based on their sensing modality and physiological function signals \cite{li2025multimodal,mcnaboe2025optimizing,wang2023wearable,moon2024comparative}. \textit{Cardiovascular} features capture heart-rate and heart-rate variability dynamics derived from photoplethysmography (PPG); \textit{Electrodermal} features characterize tonic and phasic skin conductance activity reflecting sympathetic arousal; \textit{Movement} features summarize body motion and acceleration patterns from the wrist-worn accelerometer; and \textit{Thermoregulation} features capture changes in peripheral skin temperature. This system-level organization is used to assess whether physiological modulation generalizes beyond individual features and recurs at the level of coordinated physiological subsystems.

\subsection{Task-Level Feature Aggregation}

Window-level physiological features were first globally standardized and then aggregated to the level of each experimental task block defined in Section~\ref{datacollectionprotocol}. For each participant and task, we averaged all valid windows within the task interval. Task-level aggregation was used because stress and craving annotations were collected at the task level rather than at individual physiological window timestamps.

To isolate task-evoked physiological modulation and reduce between-subject baseline differences, all task-level features were baseline-corrected using the neutral \emph{Calm video viewing} task as a subject-specific reference~\cite{hui2025adjusted,iqbal2022stress,chatzaki2025overview}. This correction emphasizes physiological changes relative to each participant's own relaxed baseline.

\subsection{GEE Model Specification}

Task-level associations were estimated using population-average generalized estimating equations (GEE) with a Gaussian family and an exchangeable working correlation structure~\cite{liang1986longitudinal}. GEE was used because each participant contributed repeated task observations, and the method estimates population-average effects while providing robust standard errors even when the within-subject correlation structure is imperfectly specified.

Each GEE model included the task condition of interest, group membership (Control vs.\ OUD), and the number of aggregated windows as covariates. Group membership controlled for between-group differences, while the number of windows controlled for variability in physiological evidence availability across tasks. All GEE models were fitted on the pooled cohort including both Control and OUD participants.

\subsection{Feature-Level and System-Level Analyses}

Extracted physiological features were organized a priori into four systems based on sensing modality and physiological function: \textbf{\emph{Cardiovascular, Electrodermal, Movement, and Thermoregulation}} (Section~\ref{feature:system}). Analyses were conducted at both the feature and system levels.

At the feature level, each baseline-corrected task-level feature was tested individually using the GEE framework described above. Multiple comparisons were controlled using the Benjamini--Hochberg false discovery rate (FDR) procedure~\cite{benjamini1995controlling}.

At the system level, feature-level effects were summarized within each physiological system using two complementary representations. First, we computed mean-based composites by averaging standardized features within each system. Second, we derived PCA-based representations using the first principal component (PC1) within each system~\cite{dastranj2024forecasting}. Both representations were evaluated using the same GEE specification.

System-level effects were considered robust when feature-level changes within the same physiological modality were directionally consistent and supported by permutation-based significance testing. System-level aggregation is used here to assess whether task-related modulation is consistently observed across related features from the same sensing modality, rather than to assign a single physical meaning to the aggregated representation.

\section{Subject-Level Guidance Embeddings}
\label{app:guidance}

The model supports two types of subject-level guidance embeddings: (i) HR-dynamics-based embeddings derived from post-task physiological signals, and (ii) semantic embeddings derived from autobiographical narratives. Both representations are computed once per subject and used as fixed inputs during model training and inference.

\paragraph{HR-Dynamics Embedding.}
The HR-dynamics embedding is constructed from short heart-rate segments collected after calm and stress tasks. For each subject, heart-rate signals are first normalized by subtracting a baseline value computed during a calm reference period (Calm video viewing). Two 15s segments are then extracted: a post-calm segment and a post-stress segment.

Each segment is summarized using 16 statistical and temporal descriptors capturing both magnitude and dynamic properties of heart-rate changes:
(1) mean, (2) standard deviation, (3) maximum, (4) minimum,
(5) signed area under the curve (AUC), (6) absolute AUC,
(7) positive AUC, (8) negative AUC,
(9) difference between final and initial values, (10) slope,
(11) time to peak, (12) time to trough,
(13) peak value, (14) trough value,
(15) number of zero crossings, and (16) curvature (mean absolute second difference).

Concatenating the post-calm and post-stress representations yields a 32-dimensional HR-dynamics embedding per subject.
\paragraph{Semantic Embedding.}
The semantic embedding is derived from autobiographical memory narratives collected during the study. Each subject provides two narratives: one corresponding to a positive (good) memory and one corresponding to a negative (bad) memory.

Each narrative is independently encoded using a pretrained text embedding model. In the current implementation, embeddings are generated using an OpenAI text embedding model (\texttt{text-embedding-3-small}), producing a 1536-dimensional vector for each narrative. For robustness analysis, embeddings generated using a higher-capacity model (\texttt{text-embedding-3-large}) were also evaluated, but the main results use the smaller model.

The final subject-level semantic representation is constructed by concatenating the good-memory and bad-memory embeddings, resulting in a 3072-dimensional vector.

No fine-tuning or task-specific adaptation is applied to the embedding model. The semantic embeddings are computed once per subject and reused across all physiological windows associated with that subject, ensuring strict separation between subject-level context and window-level inputs.

\paragraph{Summary.}
Both HR-dynamics and semantic embeddings follow the same interface: a fixed subject-level vector that encodes individual differences. They differ only in modality and dimensionality, with HR-dynamics capturing physiological recovery patterns and semantic embeddings capturing narrative-level representations of autobiographical memory.

\section{Additional Details for Autobiographical Recall as a Resilience Proxy}
\label{app:autobio_proxy_methods}

\subsection{Neurophysiological Context: Resilience and Autobiographical Memory Recall}
\label{sec:rq2_context}

We next introduce a brief neurophysiological context to motivate the use of autobiographical memory recall as a marker of resilience-related individual differences.

\paragraph{Resilience and Positive Autobiographical memories}

In opioid use disorder (OUD), chronic drug exposure disrupts both the brain’s reward and stress systems, leading to accumulated allostatic load and a pronounced reward deficit \cite{koob2007stress,koob2008addiction,koob2013addiction}. The \emph{reward deficit and stress surfeit} framework proposes that addiction is characterized by two co-occurring processes: reduced sensitivity to non-drug rewards and heightened stress reactivity \cite{koob2014addiction}. Together, these changes fundamentally alter how individuals with OUD respond to emotionally salient cues, including autobiographical memories, particularly the positive memory recalls.

In healthy individuals without substance use disorder, recalling a positive personal memory typically reactivates natural reward circuits, producing a pleasant emotional state similar to re-experiencing the event itself. In this case, the expected reward associated with the memory closely matches the experienced emotional response, resulting in a stable and positive affective experience. In contrast, among individuals with OUD— like the patients in our study who require MOUD—this process can become disrupted. The reward deficit and stress surfeit models established by Koob and colleagues provide a neurobiological framework for understanding this phenomenon. \cite{koob2014addiction,koob2008addiction}

Because baseline hedonic tone is blunted and stress systems are sensitized, attempts to recall positive, non-drug-related memories may fail to elicit the expected pleasurable response. Instead, positive memory recall can evoke ambivalent affect, anxiety, and physiological stress responses rather than relief or pleasure. This counterintuitive phenomenon reflects altered reward processing following chronic opioid exposure. It has been hypothesized that key reward-related regions, including the nucleus accumbens, function as comparators that evaluate the difference between expected and experienced reward \cite{qu2025nucleus,schultz2016dopamine}. When recalling a positive memory, individuals with OUD may experience a mismatch between the anticipated pleasure and the diminished emotional response, resulting in a negative prediction error. Rather than reinforcing positive affect, this discrepancy may emphasize loss or inability to feel pleasure, thereby triggering stress and anxiety that manifest as heightened physiological arousal. These contrasting processes can be summarized as follows:

Among healthy individuals without OUD or other SUD:
\begin{itemize}
    \item Positive memory recall task reactivates the reward circuits and produces a positive hedonic state
    \item This results in pleasant memory reactivation whereby their expectation matches their experience and results in a matched comparison between the expected and experienced positive memory recall. 
\end{itemize}
Among individuals with OUD in a reward deficit state:
\begin{itemize}
    \item Positive memory recall task attempt to reactivate reward circuits results in a failure to achieve expected hedonic state
    \item This results in a discrepancy between the expected positive hedonic state (negative prediction error) and the outcome.
    \item This negative prediction error may be interpreted as loss or threat and produce anxiety and physiological stress response.
\end{itemize}

\begin{figure}[htbp]
    \centering
    \includegraphics[width=0.95\textwidth]{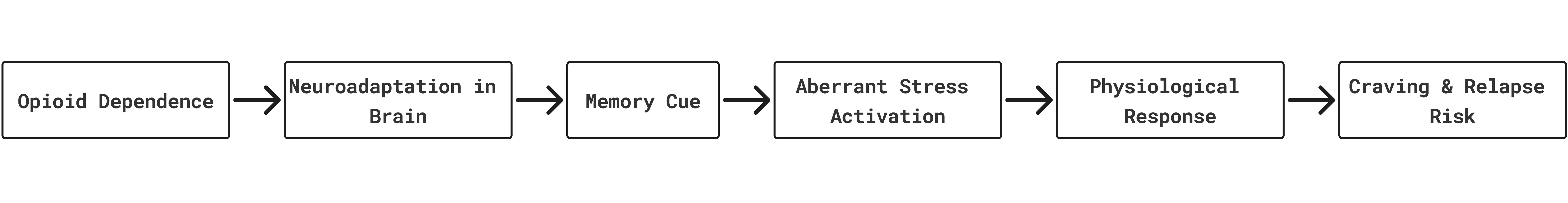}
    \caption{Neurobiological flow of stress–reward dysregulation in opioid use disorder (OUD). 
    Chronic opioid dependence alters reward and stress circuits, transforming positive memory recall into aberrant stress activation, which increases craving and relapse vulnerability.}
    \label{fig:OUD_flow}
\end{figure}

Figure~\ref{fig:OUD_flow} summarizes this pathway. 
Chronic opioid dependence drives neuroadaptations in the mesolimbic reward circuit (ventral tegmental area, nucleus accumbens), the extended amygdala, and the HPA axis. 
When a memory cue is encountered, the blunted reward circuit fails to produce the anticipated pleasure, while the sensitized stress system produces exaggerated physiological arousal—creating a feedback loop that links memory-evoked stress to craving and relapse risk.

Together, this framework provides a mechanistic basis for our hypothesis that the affective and semantic characteristics of positive autobiographical memory recall reflect resilience-related differences in stress reactivity and craving vulnerability.

\paragraph{Resilience and Negative Autobiographical memories} 

Negative autobiographical memories provide a natural context for examining resilience because they directly engage stress and emotion regulation processes.
Prior work shows that emotion regulation plays a central role in how negative autobiographical memories are recalled and experienced \cite{colombo2021moderating,pascuzzi2017emotion}. 

Individuals who use adaptive regulation strategies, such as cognitive reappraisal, tend to recall negative personal events with greater detail and reduced negative bias \cite{colombo2021moderating}. In addition, the ability to control or suppress the retrieval of intrusive memories is considered a core component of emotion regulation and has been shown to reduce the emotional impact of unwanted recollections \cite{engen2018memory,katsumi2020suppress}.

Resilience is commonly conceptualized as a set of adaptive processes that enable individuals to maintain functioning in the face of adversity \cite{troy2023psychological}. Emotion regulation is a key component of these processes \cite{schetter2011resilience}, as it governs how emotional and physiological responses to stressors and adverse events are modulated \cite{aldao2010emotion,troy2011resilience,compas2017coping}. Neurophysiological evidence further suggests that brain systems involved in arousal regulation and emotional reappraisal mediate the relationship between trait resilience and post-stress outcomes \cite{van2013neuroimaging}.

Taken together, these findings suggest that resilience shapes how negative experiences are encoded, recalled, and physiologically regulated. Individual differences in regulatory capacity influence not only the emotional tone and structure of negative autobiographical narratives, but also the magnitude and recovery of stress responses associated with them. Accordingly, negative autobiographical memory recall offers a meaningful lens through which resilience-related differences in stress regulation can be examined.

\subsection{Language Embedding Construction}

Autobiographical Language was derived from participants' descriptions of recent positive and negative personal experiences. Context-neutral speech was derived from responses to neutral everyday prompts collected during the study protocol. The context-neutral condition was used as a control for general linguistic properties such as speaking style, verbosity, and transcription characteristics without autobiographical or emotional content.

Each transcript was encoded into a high-dimensional semantic embedding using a pretrained language model. For each participant, embeddings from positive and negative autobiographical recall were aggregated to obtain a subject-level autobiographical representation. Context-neutral speech embeddings were processed separately and used only as a comparison condition.

\subsection{PLS-Based Representational Alignment}

To test whether language representations encode resilience-related recovery information, we used Partial Least Squares (PLS) to align language embeddings with HR-AUC. Unlike PCA, which identifies directions of maximal variance in the embedding space, PLS identifies latent dimensions that maximize covariance between language embeddings and an external variable of interest. Here, the external variable is HR-AUC, our physiology-grounded proxy for post-stress recovery.

For each language condition, embeddings were projected onto a single PLS component to obtain a compact latent representation. We then assessed the monotonic association between the PLS-derived language representation and HR-AUC using Spearman rank correlation. Statistical significance was evaluated using permutation testing with 5{,}000 permutations, where HR-AUC values were randomly permuted across participants to construct a nonparametric null distribution.

\subsection{Interpretation and Limitations}

A strong association between autobiographical language and HR-AUC suggests that autobiographical recall captures subject-level information related to physiological recovery. However, this analysis does not imply that language directly measures psychological resilience. Instead, autobiographical language is treated as a resilience-related descriptor that aligns with a physiology-grounded recovery proxy.

Finally, autobiographical narratives may contain sensitive personal information. In deployment, the raw transcript should not be required at inference time; instead, a compact subject-level embedding can be computed once, stored securely, and reused as guidance for wearable craving inference.

\section{Partial Least Squares (PLS) Implementation Details}
\label{appendix:pls}
Partial Least Squares (PLS) is a multivariate dimensionality reduction technique designed to relate two sets of variables measured on the same observations by extracting latent components that maximize their shared covariance. Given a predictor matrix $\mathbf{X} \in \mathbb{R}^{N \times D}$ (language embeddings) and a target vector $\mathbf{y} \in \mathbb{R}^{N}$ (heart-rate recovery AUC), PLS identifies latent variables defined as linear projections of $\mathbf{X}$ that are optimally aligned with $\mathbf{y}$ in terms of covariance. In this work, we employ a single-component PLS model, yielding a one-dimensional latent representation. Formally, PLS learns a weight vector $\mathbf{w} \in \mathbb{R}^{D}$ such that the projected scores $\mathbf{z} = \mathbf{X}\mathbf{w}$ maximize $\mathrm{Cov}(\mathbf{z}, \mathbf{y})$ under standard normalization constraints. This projection corresponds to a weighted combination of all embedding dimensions rather than the selection of individual dimensions and differs from variance-based methods such as principal component analysis (PCA), which optimize $\mathrm{Var}(\mathbf{X}\mathbf{w})$ independently of the outcome variable. Prior to PLS estimation, language embeddings were standardized to zero mean and unit variance across participants, and heart-rate recovery AUC values were used as provided by the physiological preprocessing pipeline described in Section~\ref{sec:resilience_def}. PLS was implemented using a standard Partial Least Squares regression formulation, with the number of components fixed to one to obtain a compact shared latent axis. Associations between the resulting PLS-derived latent scores and physiological recovery were evaluated using Spearman rank correlation, with statistical significance assessed via permutation testing (5,000 permutations), in which HR AUC values were randomly permuted across participants to construct a nonparametric null distribution.

\section{Qualitative Visualization: Narrative Organization and Appraisal}
\label{app:rq2_visualization}

\begin{figure}[htbp]
  \centering
  \begin{subfigure}[t]{0.48\textwidth}
    \centering
    \includegraphics[width=\textwidth]{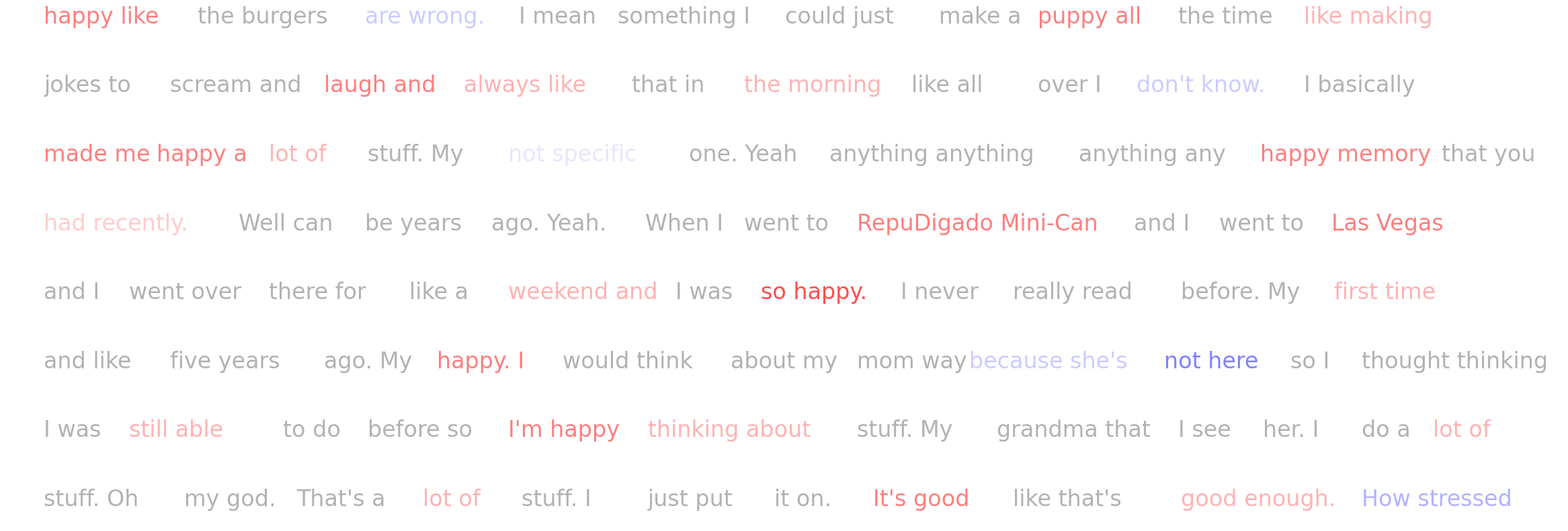}
    \caption{Participant with OUD recalling a positive memory}
    \label{fig:oud_good}
  \end{subfigure}
  \hfill
  \begin{subfigure}[t]{0.48\textwidth}
    \centering
    \includegraphics[width=\textwidth]{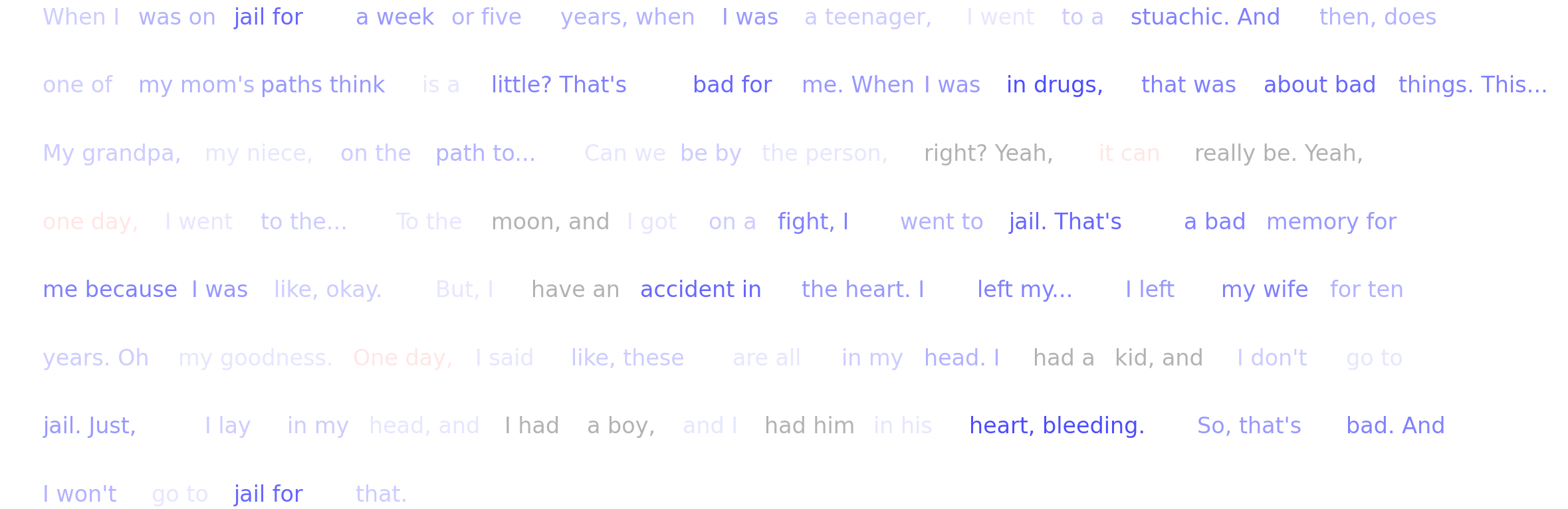}
    \caption{Participant with OUD recalling a negative memory}
    \label{fig:oud_bad}
  \end{subfigure}

  \vspace{0.5em}
  \begin{subfigure}[t]{0.48\textwidth}
    \centering
    \includegraphics[width=\textwidth]{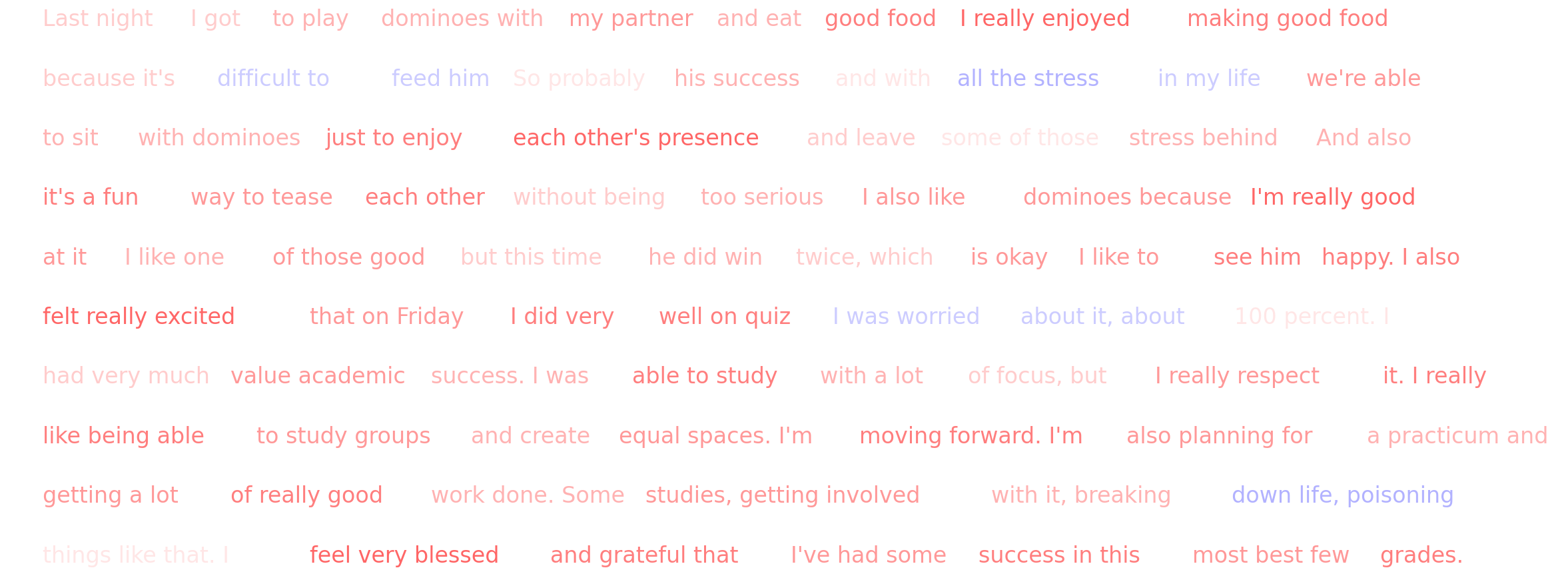}
    \caption{Control participant recalling a positive memory}
    \label{fig:ctrl_good}
  \end{subfigure}
  \hfill
  \begin{subfigure}[t]{0.48\textwidth}
    \centering
    \includegraphics[width=\textwidth]{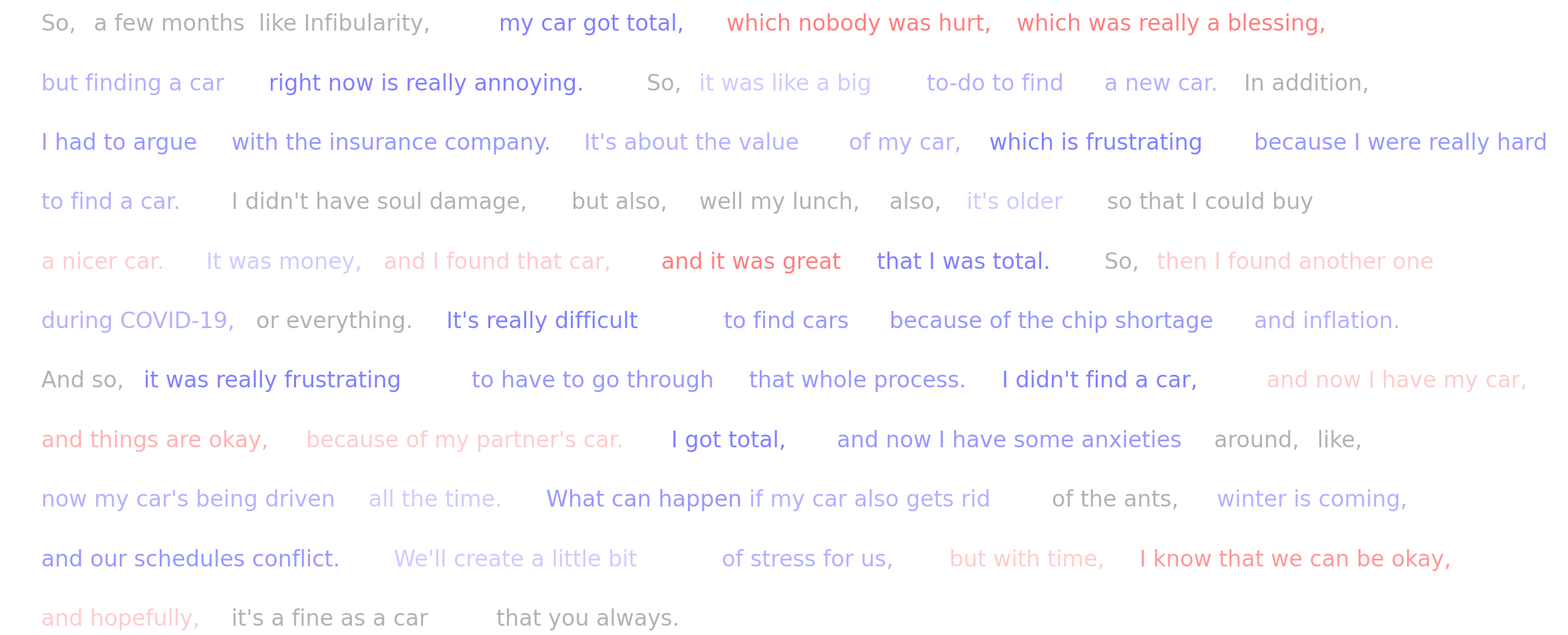}
    \caption{Control participant recalling a negative memory}
    \label{fig:ctrl_bad}
  \end{subfigure}

  \caption{Qualitative visualizations of autobiographical narratives based on LLM-derived, phrase-level resilience-related appraisal. Red indicates relatively positive resilience-related appraisal, blue indicates relatively negative appraisal, and gray indicates neutral language; color intensity reflects the relative strength of appraisal within each narrative.}
  \label{fig:resilience_vis}
\end{figure}

To aid interpretation of the alignment between autobiographical language and physiological resilience, we present qualitative visualizations of representative autobiographical narratives (Figure~\ref{fig:resilience_vis}). Each narrative is rendered using phrase-level appraisal scores generated by a large language model (GPT-4o), prompted to assess resilience-related psychological appraisal within narrative context \cite{barros2025large,jenner2025using}. The model outputs relative phrase-level scores reflecting positive (red), negative (blue), or neutral (gray) appraisal, with color intensity indicating relative affective strength within each narrative; the full prompt specification is provided in the Appendix~\ref{app:llm_prompt}. These visualizations are intended as interpretive tools rather than quantitative measurements and do not represent direct estimates of psychological resilience or momentary affect.

Across narratives, systematic differences in affective distribution and organization are evident. Positive autobiographical memories from control participants tend to exhibit localized and thematically coherent clusters of positively appraised language, whereas negative memories show denser and more sustained negatively appraised phrasing organized around salient events. In contrast, autobiographical narratives from participants with opioid use disorder (OUD) frequently display mixed or negatively appraised segments even during positive memory recall, accompanied by abrupt affective shifts and reduced thematic integration. These patterns suggest difficulties in maintaining coherent positive appraisal despite ostensibly positive content, consistent with prior narrative and clinical accounts of disrupted emotion regulation and meaning integration in addiction-related populations \cite{adler2012living,fischer2015motivational}. 

Taken together, these qualitative observations indicate that autobiographical narratives encode structured information about how individuals appraise, organize, and regulate personal experience. Differences in affective coherence, mixed valence, and narrative organization reflect systematic resilience-related appraisal processes rather than simple differences in memory valence, providing qualitative support for the modeling approach introduced in RQ3.

\section{Resilience Grouping}
\label{app:resilience_group}

To analyze subject-specific differences in physiological interpretation, participants are grouped into high- and low-resilience groups based on their post-stress heart-rate recovery. Specifically, resilience is quantified using the HR-AUC measure derived from post-stress heart-rate dynamics, computed as the deviation from baseline following stress exposure. Lower HR-AUC values indicate faster recovery and higher resilience, while higher HR-AUC values indicate slower recovery and lower resilience. Participants are split into two groups using a median split over subject-level HR-AUC values. Subjects with HR-AUC values below the median are assigned to the high-resilience group, and those above the median are assigned to the low-resilience group. 

\section{LLM Prompt for Resilience-Related Narrative Appraisal}
\label{app:llm_prompt}

To generate the qualitative narrative visualizations presented in Appendix~\ref{app:rq2_visualization}, we employ a large language model (GPT-4o) as an interpretive annotation tool to assess \emph{resilience-related psychological appraisal} within autobiographical narratives. This procedure is designed to support qualitative inspection of narrative organization, mixed valence, and appraisal coherence, rather than to produce quantitative estimates of psychological resilience or momentary affective states.

The language model is prompted to analyze how individuals appraise and organize personal experiences in narrative language. Specifically, the model is first asked to form an overall impression of the narrative’s psychological tone along a resilience--vulnerability dimension (e.g., resilient versus vulnerable). This overall assessment serves only as contextual background to anchor relative phrase-level appraisal within the narrative and is not treated as a measurement of resilience.

The model then segments each narrative into short, continuous phrases (1--5 words each), ensuring full coverage of the text in sequential order. For each phrase, the model assigns a relative appraisal score reflecting how positively, negatively, or neutrally the phrase contributes to the narrative’s overall psychological appraisal. Importantly, the prompt explicitly allows local deviations and mixed appraisal, such that negatively appraised segments may appear within an overall positive narrative and vice versa. All phrase-level scores are interpreted relative to the narrative context and are not intended to be comparable across individuals or narratives.

The full prompt used in our experiments is reproduced below for transparency and reproducibility.

\vspace{0.5em}

\begin{quote}
\small
\textbf{Prompt:}

You are a psychologist and linguistics expert who analyzes how individuals appraise and organize personal experiences in narrative language.

Your task is to assess resilience-related psychological appraisal within an autobiographical narrative. This appraisal reflects how positively, negatively, or neutrally experiences are interpreted and integrated, rather than momentary emotional intensity.

First, read the entire text carefully to form an overall impression of the narrative’s psychological tone along a resilience--vulnerability dimension, ranging from $-1$ (highly vulnerable, hopeless, anxious) to $+1$ (highly resilient, hopeful, calm, confident). This overall score is used only as contextual background.

Next, segment the text into short, continuous phrases (1--5 words each), ensuring that every word in the text is covered exactly once and in order.

For each phrase:
\begin{itemize}
    \item Consider the phrase’s psychological meaning within the broader narrative context.
    \item Assign a relative appraisal score that reflects how positively, negatively, or neutrally the phrase contributes to the narrative’s overall psychological appraisal.
    \item Allow local deviations and mixed appraisal, even when the overall narrative tone is positive or negative.
    \item Output a score between $-1$ and $+1$, where $0$ represents neutral or descriptive language.
\end{itemize}

Return ONLY valid JSON in the following format:
\begin{verbatim}
{
  "overall_score": <float between -1 and 1>,
  "phrases": [
    {"phrase": "<text>", "score": <float between -1 and 1>}
  ]
}
\end{verbatim}
\end{quote}

\vspace{0.5em}

 These annotations are used exclusively for qualitative visualization and interpretive analysis and are not used as predictive features or targets in any modeling stage.

\end{document}